\documentclass[10pt,journal,compsoc]{IEEEtran}

\ifCLASSOPTIONcompsoc
  \usepackage[nocompress]{cite}
\else
  \usepackage{cite}
\fi

\usepackage{graphicx}
\usepackage{amsmath,amssymb} 
\usepackage{color}

\usepackage{multirow}

\usepackage{amsmath}

\usepackage{wrapfig}
\usepackage{float}
\usepackage{breqn}

\usepackage{subfigure}
\usepackage{booktabs}
\usepackage{makecell}

\usepackage[misc]{ifsym}

\begin{document}
%



\title{T-Net: Deep Stacked Scale-Iteration Network for Image Dehazing}


\author{Lirong Zheng,
        Yanshan Li,
        Kaihao Zhang,
        Wenhan Luo
        \IEEEcompsocitemizethanks{
\IEEEcompsocthanksitem Lirong Zheng and Yanshan Li (Corresponding author) are with the ATR National Key Lab. of Defense Technology and Guangdong Key Laboratory of Intelligent Information Processing, Shenzhen University, Shenzhen, 518060, China. E-mail: \{zhenglirong2019@email.szu.edu.cn, lys@szu.edu.cn\} \protect\\
\IEEEcompsocthanksitem Kaihao Zhang is with the College of Engineering and Computer Science, Australian National University, Canberra, ACT, Australia. E-mail: \{kaihao.zhang@anu.edu.au\} \protect\\
\IEEEcompsocthanksitem Wenhan Luo is with Tencent, Shenzhen, China. E-mail: \{whluo.china@gmail.com\}  \protect\\
}

\thanks{Manuscript received April 19, 2005; revised August 26, 2015.}
}

%
%

\markboth{Journal of \LaTeX\ Class Files,~Vol.~14, No.~8, August~2015}%
{Shell \MakeLowercase{\textit{et al.}}: Bare Advanced Demo of IEEEtran.cls for IEEE Computer Society Journals}
%



\IEEEtitleabstractindextext{%
\begin{abstract}
Hazy images reduce the visibility of the image content, and haze will lead to failure in handling subsequent computer vision tasks. In this paper, we address the problem of image dehazing by proposing a dehazing network named T-Net, which consists of a backbone network based on the U-Net architecture and a dual attention module. And it can achieve multi-scale feature fusion by using skip connections with a new fusion strategy. Furthermore, by repeatedly unfolding the plain T-Net, Stack T-Net is proposed to take advantage of the dependence of deep features across stages via a recursive strategy. In order to reduce network parameters, the intra-stage recursive computation of ResNet is adopted in our Stack T-Net. And we take both the stage-wise result and the original hazy image as input to each T-Net and finally output the prediction of clean image. Experimental results on both synthetic and real-world images demonstrate that our plain T-Net and the advanced Stack T-Net perform favorably against the state-of-the-art dehazing algorithms, and show that our Stack T-Net could further improve the dehazing effect, demonstrating the effectiveness of the recursive strategy. 
\end{abstract}

}

\maketitle

\IEEEdisplaynontitleabstractindextext

%
\IEEEpeerreviewmaketitle

\section{Introduction}
Haze is a common atmospheric phenomenon that not only has a serious adverse effect on human visual perception, but also has a serious impact on the performance of modern computer vision systems for various visual tasks, such as image classification, object detection and video surveillance. Since hazy conditions like fog, aerosols, sands and mists can scatter and adsorb light, images taken in the hazy environment suffer from obscured visibility, reduced contrast, color cast and many other degradations. Therefore, it is crucial to develop effective solutions to reduce the impact of image distortion caused by environmental conditions through dehazing. 

\begin{figure}[tbp]
    \centering
    \renewcommand{\tabcolsep}{1pt}
    \renewcommand{\arraystretch}{1}
    \begin{center}
    \begin{tabular}{ccc}
    \includegraphics[width=0.32\linewidth]{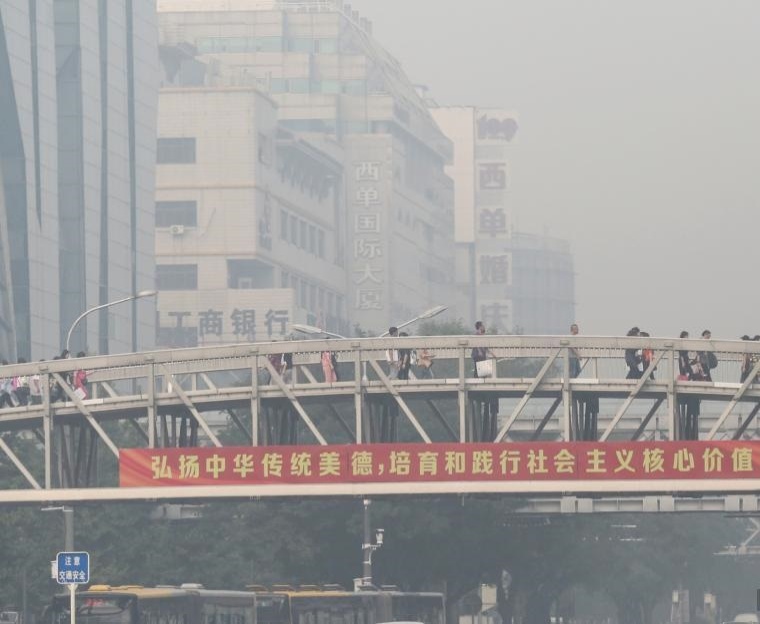} &
    \includegraphics[width=0.32\linewidth]{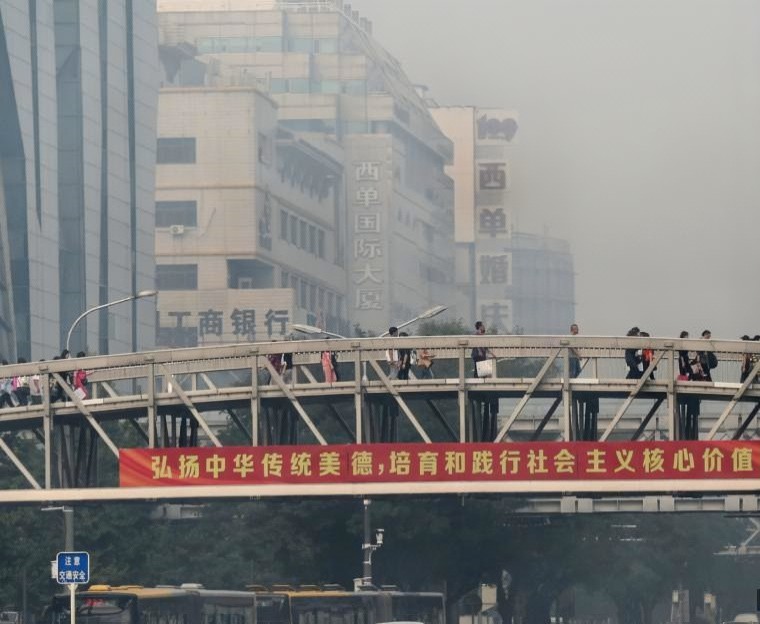} &
    \includegraphics[width=0.32\linewidth]{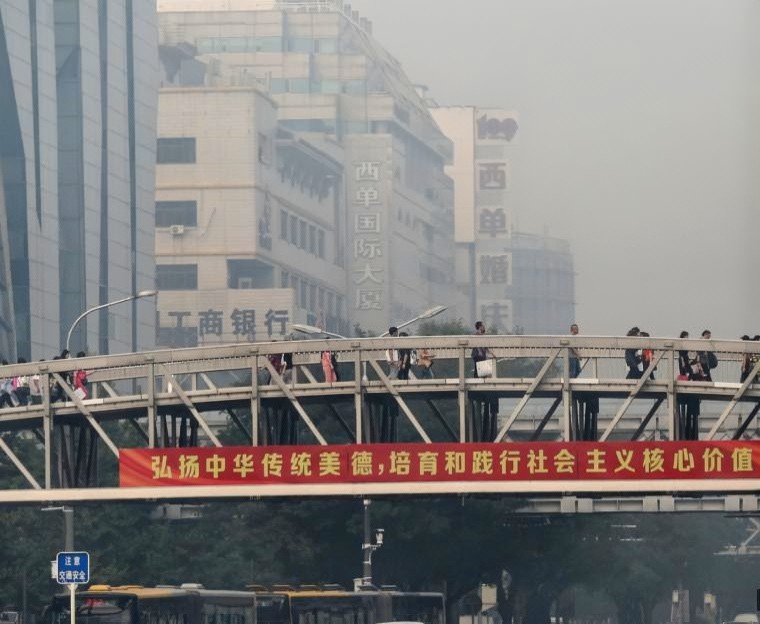} \\

    \includegraphics[width=0.32\linewidth]{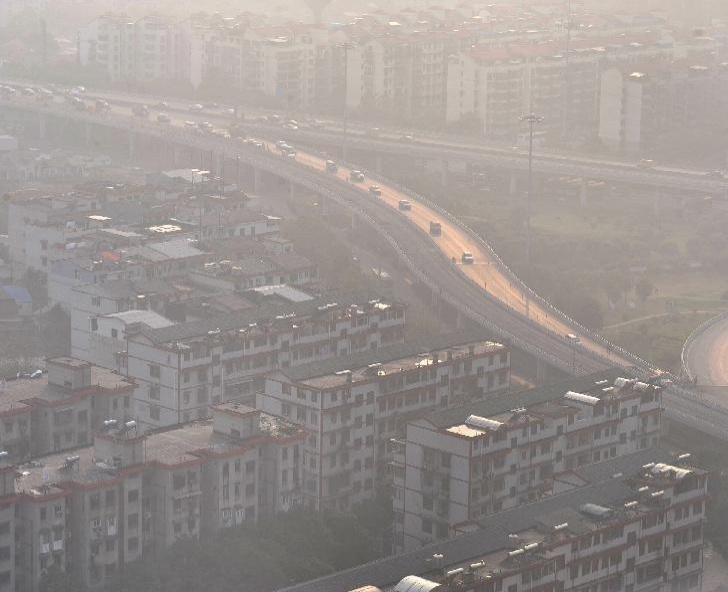} &
    \includegraphics[width=0.32\linewidth]{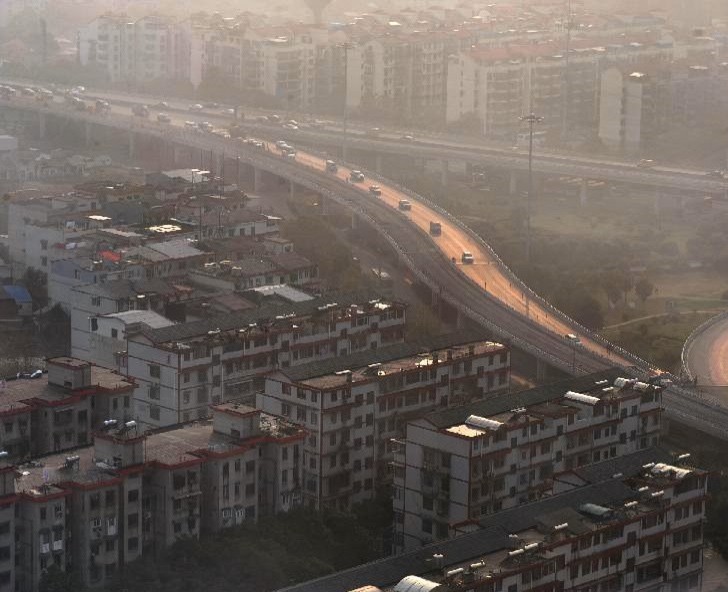} &
    \includegraphics[width=0.32\linewidth]{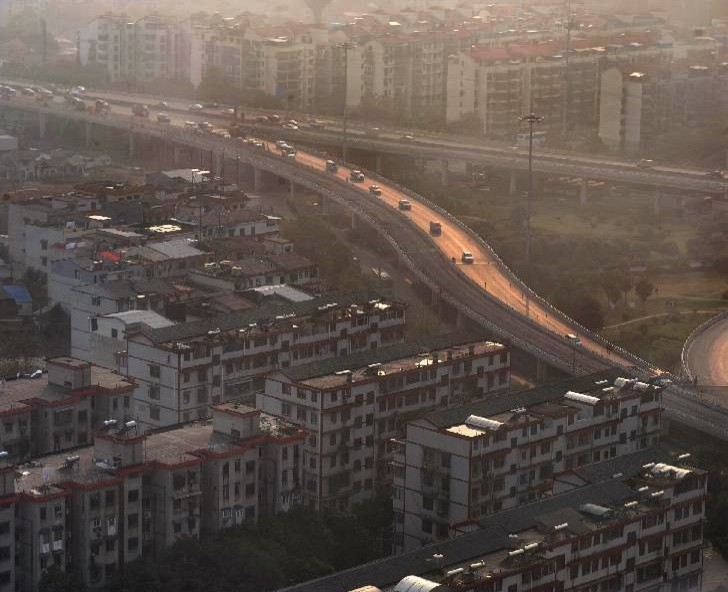} \\
        
    \includegraphics[width=0.32\linewidth]{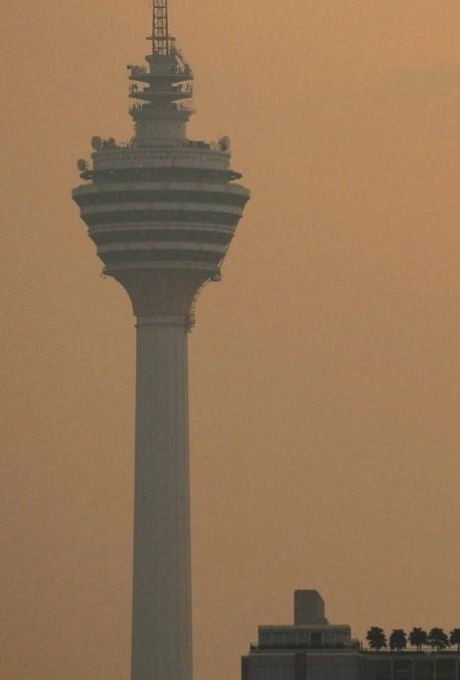} &
    \includegraphics[width=0.32\linewidth]{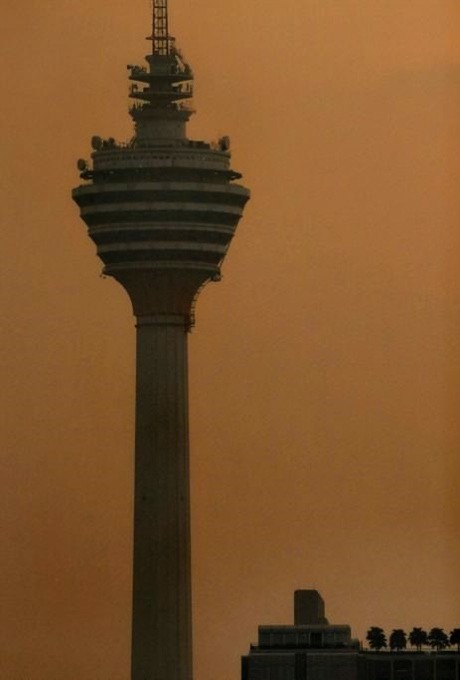} &
    \includegraphics[width=0.32\linewidth]{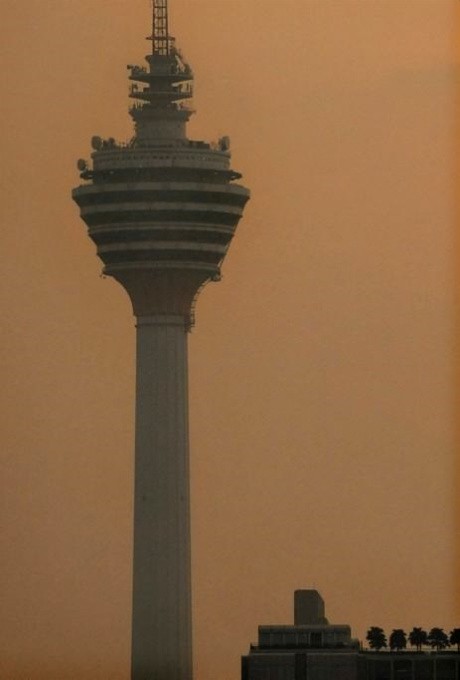} \\
    
    (a) Hazy image &
    (b) T-Net &
    (c) Stack T-Net\\
    \end{tabular}
    \end{center}
    \caption{Dehazed results on real-world hazy images. We show the results of T-Net and Stack T-Net, which both perform well on these images. Compared with T-Net, the results of Stack T-Net are cleaner and brighter and exhibit finer details.}
    \label{fig: intro}
\end{figure}

In the past ten years, single image dehazing has attracted widespread attention in the computer vision community. Its goal is to restore clean scenes from hazy images. According to the atmosphere scattering model~\cite{1977Optics,2002Vision,2016Haze}, the hazy process could be approximated as,
\begin{equation}
    I(x)=J(x)t(x)+A(x)(1-t(x)),
\end{equation}
where $I(x)$ and $J(x)$ separately denote a hazy image and its clean scene, $A(x)$ represents the global atmospheric light, $t(x)$ describes the transmission map and $x$ is the pixel location. And the transmission map $t(x)$ is related to the depth $d(x)$ of the image, represented as $t(x)=e^{-\beta{d(x)}}$, where $\beta$ is the atmospheric scattering coefficient and controls the scale of the haze. Many methods based on this physical model restore clean scene $J(x)$ from a hazy image $I(x)$ by estimating the global atmospheric light and the transmission map of the hazy image.

Early prior-based methods attempt to estimate the transmission map by using the difference in statistical properties of hazy and clear images, such as contrast prior~\cite{4587643}, dark channel prior (DCP)~\cite{5567108}, color-lines~\cite{2651362}, color attenuation prior~\cite{7128396}, haze-lines~\cite{Berman_2016_CVPR}, and color ellipsoidal prior~\cite{8101508}. However, these hypothetical priors are not suitable for all of the real-world images, and it is easy to obtain inaccurate approximations of the transmission map, which will in turn cause the quality degradation of the restored image. To solve this problem, driven by the unprecedented success in recent years of deep learning in low level vision tasks~\cite{2016Image,2016Deeply,Ledig_2017_CVPR,2017Image,2016Beyond}, Convolution Neural Networks (CNNs) have been adopted for the task of image dehazing.

Earlier deep learning based methods, including the DehazeNet~\cite{2016DehazeNet}, the multi-scale CNN (MSCNN)~\cite{2016Single}, and the residue learning technique~\cite{8451663}, use deep networks to estimate the transmission map and exploit the conventional DCP's~\cite{5567108} bright pixels method to estimate the atmospheric light. However, since the estimation of the atmospheric light is coarse, the results of these methods are not sufficiently satisfactory. Thereby some methods try to use deep networks to estimate the transmission map and atmospheric light, simultaneously or respectively. For example, Li et al.~\cite{8237773} propose a network named AOD-Net to estimate a new variable which combines the transmission map with the atmospheric light. And Zhang et al.~\cite{8578435} use a two-stream network to estimate these two variables respectively.

In addition to the methods based on the atmospheric scattering model, there are some works that use end-to-end deep neural networks to directly estimate clean images instead of performing explicit estimation of the transmission map and atmospheric light. Considering that the estimation of the transmission map and atmospheric light sometimes deviates from the real hazy image, directly estimating the clean image is able to avoid sub-optimal restoration. Inspired by the successes of the generative adversarial networks (GANs) in synthesizing realistic images, many methods use GANs as the framework to transform hazy images to clean images. For instance, Qu et al.~\cite{8953692} propose a GAN based on pix2pix to realize image-to-image translation for dehazing. 

Image restoration methods based on deep learning are driven by data, requiring large-scale synthetic or real-world datasets for training. These datasets are created by using the atmosphere scattering model to simulate the process of image degradation. Although proper guidance by the transmission map and atmospheric light is beneficial for effective dehazing under a wide range of haze density and for dealing with color distortions caused by haze, the algorithms based on the estimation of the transmission map and atmospheric light also rely on the same physical model. These algorithms may suffer from inherent performance loss on real-world images due to the over-fitting of the algorithms on synthetic datasets. In comparison, the methods of direct mapping exhibit higher robustness on real-world hazy images.

For this reason, we design an end-to-end network named T-Net for single image dehazing, which directly predicts clean images without using the atmospheric scattering model. Our T-Net is a symmetrical T-shaped architecture which is composed of two components, a backbone module and a dual attention module. The backbone module is a network based on the U-Net~\cite{2015UNet} architecture, in which residual dense block, upsampling block and downsampling block are used as basic blocks. In order to obtain distinctive features conducive to dehazing, a dual attention module~\cite{Fu_2019_CVPR} is embedded in the backbone module, which can also enhance the robustness of the network. Meanwhile, in view of the effectiveness of multi-scale features for image restoration~\cite{2016Single,7327182,Yang_2017_CVPR,Zhang_2018_CVPR,8575288,Liu_2019_ICCV,9237759}, we use skip connections with a new fusion strategy~\cite{NIPS2017_3f5ee243} in T-Net to realize adaptive multi-scale feature fusion. Moreover, inspired by the recurrent structure of~\cite{Yang_2017_CVPR} and~\cite{Ren_2019_CVPR}, we propose Stack T-Net by repeatedly unfolding our T-Net to further improve the performance of dehazing. Each stage of the Stack T-Net uses the output of the previous stage and the original hazy image as input and outputs clean image. Fig.~\ref{fig: intro} shows the dehazed results of our proposed T-Net and Stack T-Net on exemplar real-world images.

To summarize, our work has three-fold contributions as follows.

1) T-Net, a novel network with a dual attention module based on the U-Net architecture, is proposed to realize efficient information exchange across the multi-scale features from different levels to directly predict clean images.

2) Stack T-Net, which is created by repeatedly unfolding T-Net, can further improve the performance of dehazing by taking both the stage-wise result and the original rainy image as input to each stage.

3) Extensive experiments show that our plain T-Net and Stack T-Net perform favorably against the state-of-the-art methods on both synthetic and real-world hazy images. Meanwhile, an ablation study is conducted to demonstrate the effects of different modules in the proposed network.

The rest of the paper is organized as follows. Section~\ref{sec: rel} discusses the related work of single image dehazing. Section~\ref{sec: tnet} describes the structure of the plain network T-Net, including a backbone module and a dual attention module. Section~\ref{sec: stack} explains the details of our overall dehazing network Stack T-Net and defines the loss function for training. Section~\ref{sec: exp} introduces the datasets used in the experiments, and presents the results of the ablation study and the quantitative and qualitative comparisons between our approach and the state-of-the-art methods on synthetic and real-world images. Section~\ref{sec: concl} concludes the paper and discusses the direction of the future work. 


\section{Related Work}
\label{sec: rel}
In this section, we briefly discuss the single image dehazing methods, which could be roughly divided into two categories, prior-based methods and learning-based methods as mentioned above.

\subsection{Prior based methods}
Single image dehazing is a highly ill-posed problem in computer vision, and it is difficult to dehaze images without additional information. To address this challenge, different priors or assumptions obtained through observations and statistics on a large amount of real data have been used in many dehazing methods. Except the earliest methods which are based on image enhancement algorithms~\cite{557356,663733,736004,2791513}, most prior-based dehazing methods are proposed based on the atmosphere scattering model, obtaining clean images by estimating the transmission map and the atmospheric light to invert Eq.(1). Representative works following this route include~\cite{1360671,4587643,5567108,2651362,7128396,Berman_2016_CVPR,8101508}, etc.

Fattal et al.~\cite{1360671} propose a dehazing technique by estimating the albedo under the assumption that the transmission map and surface shading are locally uncorrelated. Tan et al.~\cite{4587643} propose a local contrast-maximization method based on Markov Random Field (MRF), due to the observation that the contrast of clear images is higher than the contrast of the corresponding hazy images. Observing that the intensity of at least one color channel in local regions of natural haze-free images is close to zero and that the pixel intensity increases as haze increases, He et al.~\cite{5567108} propose the dark channel prior (DCP) as the approximation of haze distribution to estimate the transmission map. Meanwhile, this work also proposes a popular way to estimate the atmosphere light by averaging the top 0.1$\%$ of brightness pixels in hazy images. Many subsequent works make improvement on the DCP to refine the estimation of the transmission map by using different edge-preserving smoothing filters~\cite{8030116,6957555,7279117,6319316}. Discovering that haze causes the color-lines~\cite{1315267} to deviate from the origin, Fattal et al.~\cite{2651362} propose a new method to recover the transmission map, where color-lines are proposed by observing that pixels of small image patches typically exhibit an one-dimensional distribution. The color attenuation prior adopted in the linear model of~\cite{7128396} is based on the assumption that, as haze increases, the brightness of images increases but saturation decreases. With the assumption that several hundreds of distinct colors can well represent colors of a haze-free image, Berman et al.~\cite{Berman_2016_CVPR} observe that these colors in hazy images form clusters along lines in the RGB space and these lines pass through the coordinate value corresponding to the atmospheric light, where these lines are named haze-lines. Based on this observation, they proposed a method to estimate the transmission map by calculating the haze-lines with the predicted atmospheric light which is computed with the brightness pixels method of DCP. Biu and Kim~\cite{8101508} construct color ellipsoids by statistically fitting haze pixel clusters in the RGB space and then calculate a prior vector through color ellipsoid geometry to obtain the transmission map. In spite of different degrees of success which have been obtained, the prior based methods are limited by the hypothetical priors themselves. There is a certain gap between these priors and the reality, which can cause that the performance of these dehazing methods is not sufficiently satisfactory.

\subsection{Deep learning based methods}
Recently, deep learning achieves significant success in low-level vision tasks such as image super-resolution \cite{ledig2017photo,johnson2016perceptual,niu2021blind}, deblurring \cite{zhang2020deblurring,zhang2018adversarial}, deraining \cite{zhang2020beyond,zhang2021beyond}, desnowing \cite{liu2018desnownet,zhang2021deep}, which also include dehazing \cite{2016DehazeNet,2016Single}.
%
At present, there are two kinds of main ideas about the dehazing methods based on deep learning. One is to estimate the transmission map and atmospheric light according to the atmospheric scattering model and the other uses deep learning network to directly predict clean images.

The deep learning dehazing methods according to the physical model use the same strategy as the prior based methods to restore clean images, but generally estimate the transmission map and atmospheric light by using specially designed CNNs instead of priors so as to avoid the limitation on performance caused by the gap between hand-crafted priors and the reality. Cai et al.~\cite{2016DehazeNet} propose a dehazing model, DehazeNet, to estimate the transmission map. Ren et al.~\cite{2016Single} design a Multi-Scale CNN (MSCNN) to estimate the transmission map with a coarse-to-fine strategy. Zhang et al.~\cite{8578435} employ a densely connected pyramid dehazing network (DPCN) which is a two-stream network to predict the transmission map and the atmospheric light, respectively. Li et al.~\cite{8237773} create a reformulation of the atmosphere scattering model by using a new variable to integrate the transmission map and the atmospheric light and design an end-to-end neural network named AOD-Net to estimate this variable. Zhang et al.~\cite{8753731} introduce Famed-net to estimate the same variable of~\cite{8237773}. Dudhane et al.~\cite{8802288} propose PYF-Net, which consists of a YNet for the estimation of the transmission map in the RGB and YCbCr space and a FNet to fuse two transmission maps. Recently, there are some works that combine the traditional method DCP with deep learning. For example, Golts et al.~\cite{8897130} design a new loss based on the DCP, which is used for the training of an unsupervised deep network to estimate the transmission map. Chen et al.~\cite{8954465} exploit a Patch Map Selection Net to adaptively set the patch size of dark channel corresponding to each pixel.  

The deep learning dehazing methods of direct mapping regard dehazing as an image-to-image translation problem. In recent years, generative adversarial networks (GANs) have shown great success in image generation and translation~\cite{Isola_2017_CVPR}~\cite{Zhu_2017_ICCV}~\cite{Ledig_2017_CVPR}, and are subsequently used in the field of dehazing. Qu et al.~\cite{8953692} propose a pix2pix GAN with two enhancing blocks to predict the clean images and enhance the detail and color. Raj et al.~\cite{9198400} design a condition GAN based on the U-Net architecture for dehazing. Engin et al.~\cite{8575279} design an end-to-end dehazing network based on CycleGAN with no need for paired hazy and corresponding clean images for training. Du et al.~\cite{Du2018PerceptuallyOG} design a GAN with an adaptive loss to facilitate end-to-end perceptual optimization and propose a new post processing method for halo artifacts removal using guide filters. Shao et al.~\cite{Shao_2020_CVPR} apply an end-to-end network which is made up of two sub-networks, a bidirectional translation network based on CycleGAN to bridge the gap between the synthetic and real domains and a dehazing network to restore clean images from the hazy images before and after translation. In addition to GAN, some works are proposed based on other architectures. Ren et al.~\cite{Ren_2018_CVPR} introduce an end-to-end network based on the encoder-decoder architecture, which adopts a novel fusion-based strategy that derives three inputs by using three pre-processing methods. Liu et al.~\cite{Liu_2019_ICCV} propose a grid network GridDehazeNet for single image dehazing, which consists of three modules, pre-processing, backbone, and post-processing. Li et al.~\cite{Li_2019_ICCV} introduce a level-aware progressive network (LAP-Net), of which each stage learns different levels of haze with different supervision and the final output is yielded with an adaptive integration strategy. Dong et al.~\cite{Dong_2020_CVPR} propose a Multi-Scale Boosted Dehazing Network (MSBDN) based on the U-Net architecture with two principles, boosting and error feedback to realize dense feature fusion.

\section{T-Net}
\label{sec: tnet}
\begin{figure}[t]
    \centering
    \renewcommand{\tabcolsep}{1pt} 
    \renewcommand{\arraystretch}{1} 
    \includegraphics[width=8.5CM]{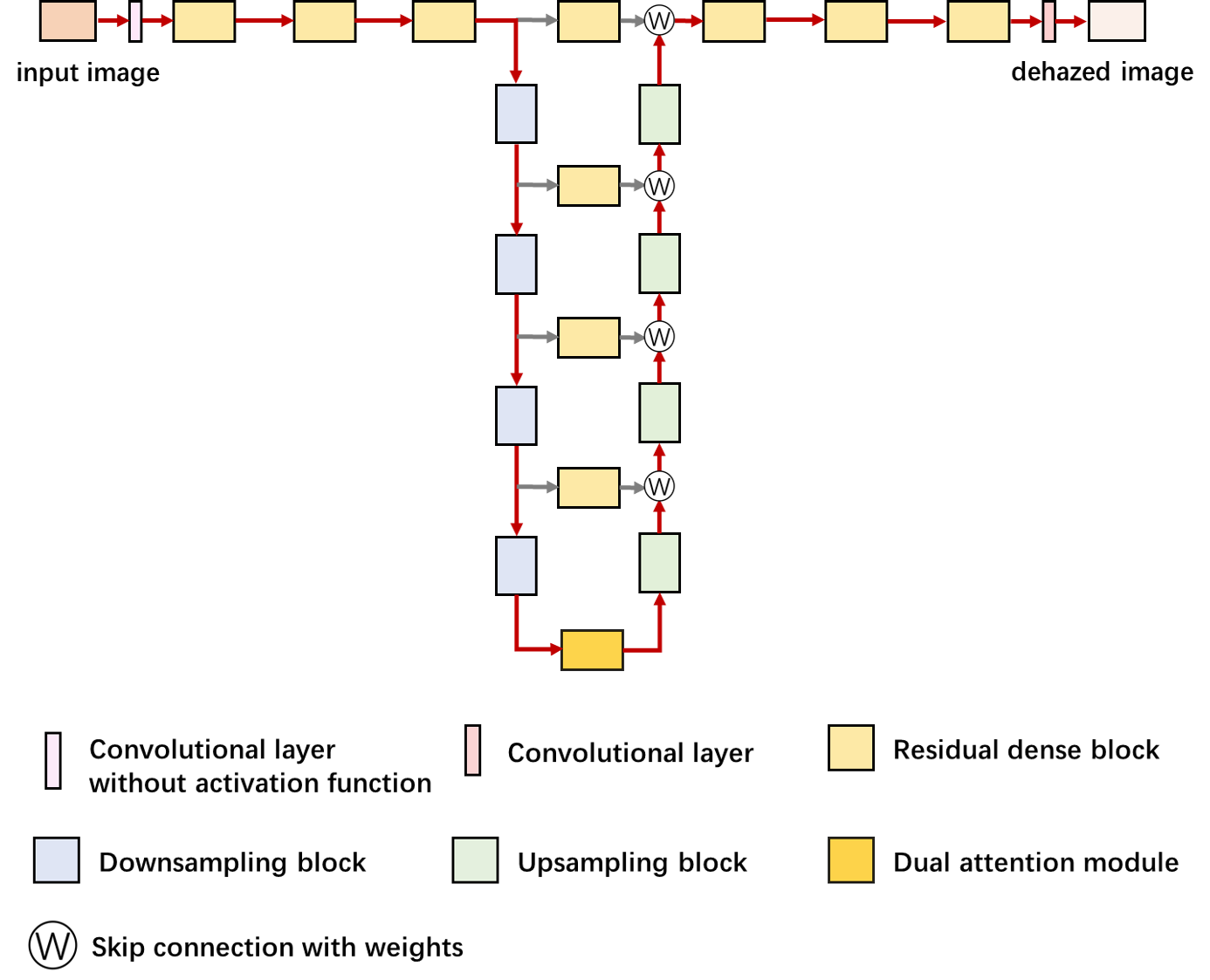}
    \caption{The architecture of T-Net. It is an end-to-end network to directly obtain the dehazed result, consisting of a backbone module and a dual attention module (dark yellow rectangular block). The red path is the trunk road of the network, and the gray paths are skip connections with weights.}
    \label{fig: T-Net}
\end{figure}

In this section, we describe the proposed T-Net which is a symmetrical T-shaped network consisting of two sub-modules, a backbone module based on the U-Net architecture and a dual attention module. Fig.~\ref{fig: T-Net} illustrates the architecture of T-Net, and the details are given in the following.

\subsection{Backbone module}
\begin{figure}[ht]
    \centering
    \renewcommand{\tabcolsep}{1pt} 
	\renewcommand{\arraystretch}{1} 
    \includegraphics[width=8.5cm]{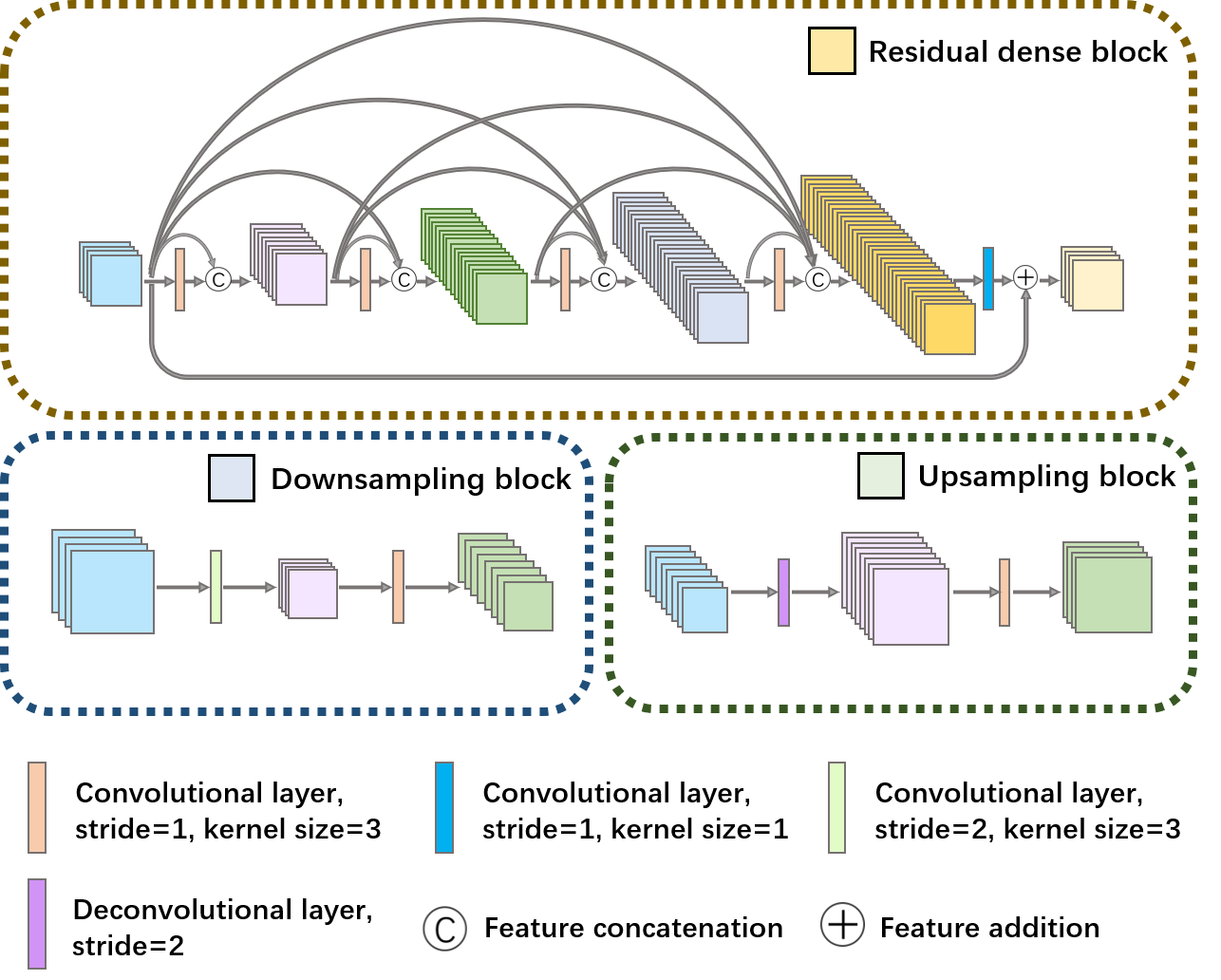}
    \caption{The detailed structure of three kinds of blocks in the backbone module of the T-Net. The yellow, blue and green doted boxes separately represent the RDB block, the downsampling block and the upsampling block, where we illustrate the variation of the feature size and the number of channels in each layer of each kind of block.}
    \label{fig: Blocks}
\end{figure}

The backbone module of T-Net is based on the U-Net architecture, and we adopt a new fusion strategy to make use of the skip connections. As shown in Fig.~\ref{fig: T-Net}, the network mainly includes three kinds of basic blocks, residual dense block (RDB)~\cite{Zhang1_2018_CVPR}, upsampling block and downsampling block. And the detailed structure of these three kinds of blocks is shown in Fig.~\ref{fig: Blocks}.

Extensive research has demonstrated that the use of multi-scale features is beneficial to various tasks of image understanding. The high-level features are of the downsampled spatial resolution but compress more semantic contextual information that is necessary for scene understanding of images. In contrast, the low-level features are of higher resolution to help localize objects but contain less semantic contextual information. Thereby the feature fusion of different levels can simultaneously preserve spatial information from low-level features and exploit the semantic contextual information from high-level features for image understanding.

Actually, image restoration includes the process of image understanding to determine what should be kept on the images and what should be removed. In order to make full use of the information from different levels of features, multi-scale feature fusion has been applied to many image restoration tasks such as image derainning~\cite{Yang_2017_CVPR,Zhang_2018_CVPR}, image debluring~\cite{Kupyn_2019_ICCV,Nah_2017_CVPR}  and image dehazing~\cite{8575288,Liu_2019_ICCV,Dong_2020_CVPR}, which significantly improves the performance of the algorithms. The U-Net~\cite{2015UNet} architecture is originally proposed for semantic segmentation, consisting of a contracting path to capture contextual cues and a symmetric expanding path for precise localization as well as multiple lateral connections between these contracting paths and their symmetric expanding paths for multi-scale feature fusion. We design our T-Net based on this architecture and make several improvements for the need of dehazing. 

As shown in Fig.~\ref{fig: T-Net}, the backbone is a symmetric network, which mainly includes three pairs of RDB blocks and four pairs of upsampling and downsampling blocks in the trunk road (the red path in Fig.~\ref{fig: T-Net}), and three RDB blocks in the lateral connections. In addition, there is a convolutional layer without activation function at the beginning of the network, which generates 16 linear feature maps as the learned input from hazy image. And there is another convolutional layer at the end, which is symmetrical to the beginning and used to generate high-quality dehazed images. 

The RDB~\cite{Zhang1_2018_CVPR} block which keeps the number of feature maps unchanged can extract abundant local features via densely connected convolutional layers, so we choose it as the basic block for feature generation. As shown in Fig.~\ref{fig: Blocks}, we use five convolutional layers in each RDB and set the growth rate to 16. Every convolutional layer takes the concatenation of the output features of all convolutional layers before it as input, and the final output of each RDB block is the combination of the output of the last layer and the input of this RDB block through channel-wise addition, where the first four convolutional layers are used to extract features and the last layer (kernel size = 1, 1x1 convolution) is used to keep the number of channels of the output feature the same as that of the input feature. In order to reduce information loss, we use convolutional layers to realize upsampling and downsampling, and the detailed structure is shown in Fig.~\ref{fig: Blocks}. The upsampling block and the downsampling block both consist of two convolutional layers of which the first layer adjusts the size of feature maps by setting different kernel sizes of convolutional layers and the other layer uses 1x1 convolution to change the number of channels. In each upsampling block, the number of feature maps decreases by half as the size of feature maps increases by two times, which is the reverse in each downsampling block.

We use a new fusion strategy to realize skip connections, for the consideration that features from different scales may not be equally important. Motivated by~\cite{NIPS2017_3f5ee243}, we set two trainable fusion parameters for each skip connection, where every fusion parameter is a $n$-dimensional vector ($n$ is the number of channels of the features before fusion). Moreover, RDB blocks instead of 1x1 convolution are used in the lateral connection to obtain more feature combinations, which improves the possibility of obtaining more effective information for dehazing. The backbone includes four pairs of upsampling and downsampling blocks, so the feature from feature fusion could be expressed as
\begin{equation}
    \begin{split}
        \breve{F}_i^j=\alpha &f_r(F_i^j)+\beta f_u(f_x(F_{i+1}^j)),\\
        &F_{i+1}^j=f_d(F_i^j),\\
        f_x(\cdot)&=
        \begin{cases}
            f_r(\cdot),\quad i < m-1,\\
            f_a(\cdot),\quad i = m-1.
        \end{cases}\\
        i=0,1,&\cdots,m-1;\quad j=0,1,\cdots,n,
    \end{split}
\end{equation}
where $\alpha$ and $\beta$ are the fusion parameters for the two features to be fused, $f_r(\cdot)$, $f_u(\cdot)$, $f_d(\cdot)$ and $f_a(\cdot)$ separately stand for the functions of RDB, upsampling, downsampling and dual attention, $F^j_i$ represents the $j$-th feature channel after the $i$-th downsampling, $\breve{F}_i^j$ represents the fused feature symmetric with $F^j_i$, $m$ is the number of the upsampling and downsampling blocks pairs, and $n$ is the number of channels of the features.

\subsection{Dual attention module}
\begin{figure}[ht]
    \centering
    \renewcommand{\tabcolsep}{1pt} 
	\renewcommand{\arraystretch}{1} 
    \includegraphics[width=8.5cm]{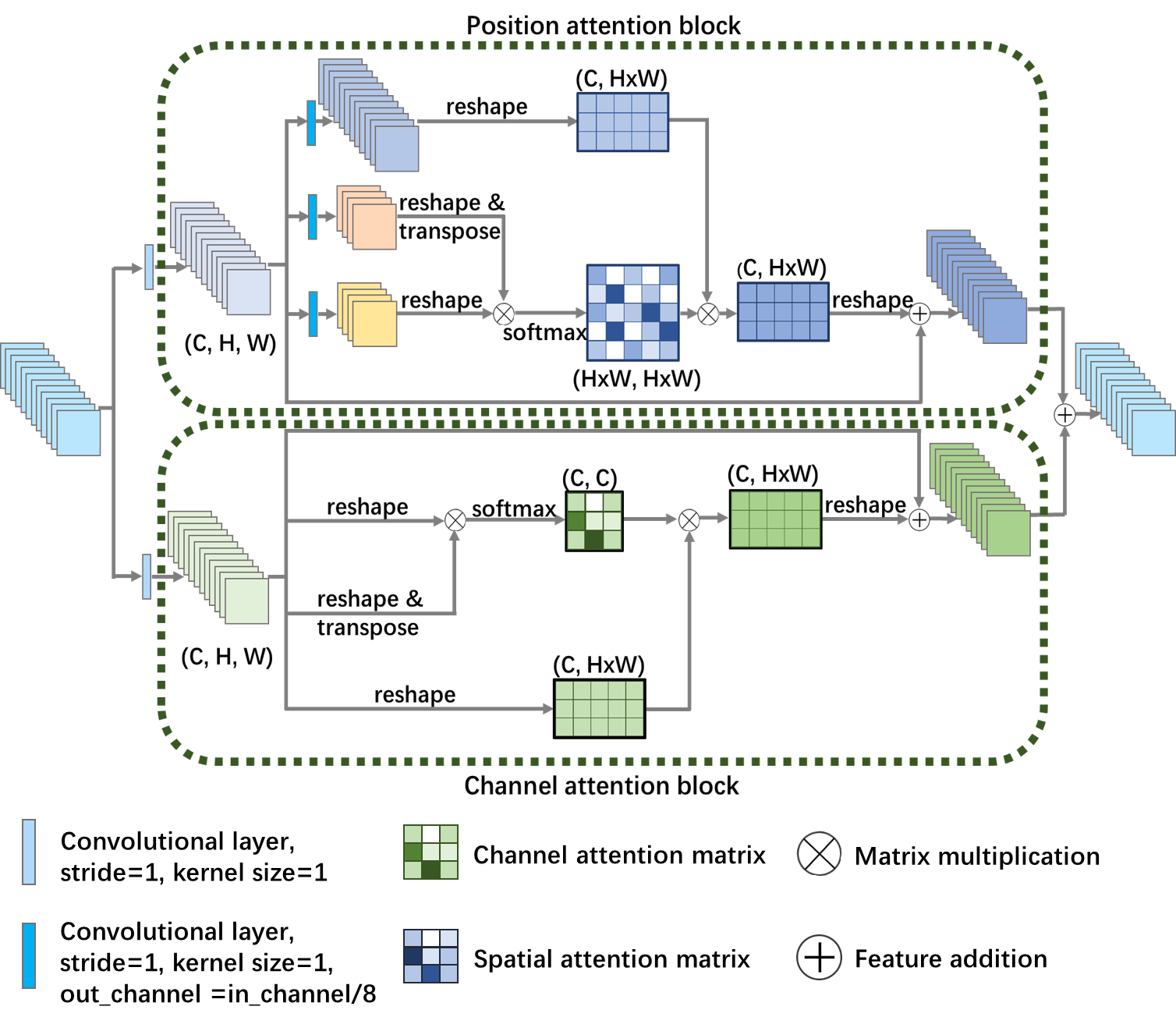}
    \caption{An overview of the dual attention module. The final output of the dual attention module is the sum of the outputs of the position attention block and the channel attention block, and has the same feature size and number of channels as the input.}
    \label{fig: Dual}
\end{figure}

As shown in Fig.~\ref{fig: T-Net}, in addition to the backbone module, we use a dual attention module in the T-Net, which is embedded in the middle of the backbone module and includes two blocks, the position attention block and the channel attention block. Fig.~\ref{fig: Dual} is an overview of the dual attention module.

Discriminant feature representations are the key for scene understanding, which could be obtained by capturing long-range contextual information. The dual attention module is originally proposed in~\cite{Fu_2019_CVPR} for scene segmentation, which could adaptively integrate local features with their global dependencies. The position attention block encodes a wider range of contextual information into local features by a weighted sum of the features at all positions, thus enhancing their representation capability. The channel attention block emphasizes interdependent feature channels by integrating associated features among all channels, to further improve the feature representation of specific semantics. These two attention blocks could be simply regarded as two different dimension feature enhancers, and expand the range of semantic contextual information from two different dimensions.

As shown in Fig.~\ref{fig: Dual}, the position attention block generates feature maps of spatial long-range contextual through three steps. Firstly, generating a spatial attention matrix by a softmaxed product of two new features of which one is obtained by reshaping the input feature to $\mathbb{R}^{C\times (H\times W)}$, and the other is the transposition of first new feature, where $(H\times W)$ is the number of pixels. Secondly, getting an attention increment by performing a matrix multiplication between the attention matrix and the second new feature. Thirdly, obtaining the final representations reflecting long-range contexts through an element-wise sum of the attention increment and the original input feature. The channel attention block captures the channel relationship through similar steps, except for the first step, in which channel attention matrix is calculated in channel dimension instead of position dimension. 

As explained above, it is obvious that these two attention blocks focus on the content similarity of features, that is, the interdependence between features. They can adaptively emphasize the effective information in the features under the guidance of the loss function and obtain discriminant feature representations according to different input images, without being limited by the network structure itself. Thereby, we believe combining the U-Net architecture with the dual attention module can further improve the robustness of the network, and the discriminant features obtained by the dual attention module can guide the parameter learning of the network. As shown in Fig.~\ref{fig: T-Net}, we embed the dual attention module in the middle of the last pair of upsampling and downsampling blocks to extract discriminant feature representations from high-level semantic features.

\section{Stack T-Net}
\label{sec: stack}
\begin{figure*}
    \centering
    \renewcommand{\tabcolsep}{1pt} 
	\renewcommand{\arraystretch}{1} 
    \includegraphics[width=18cm]{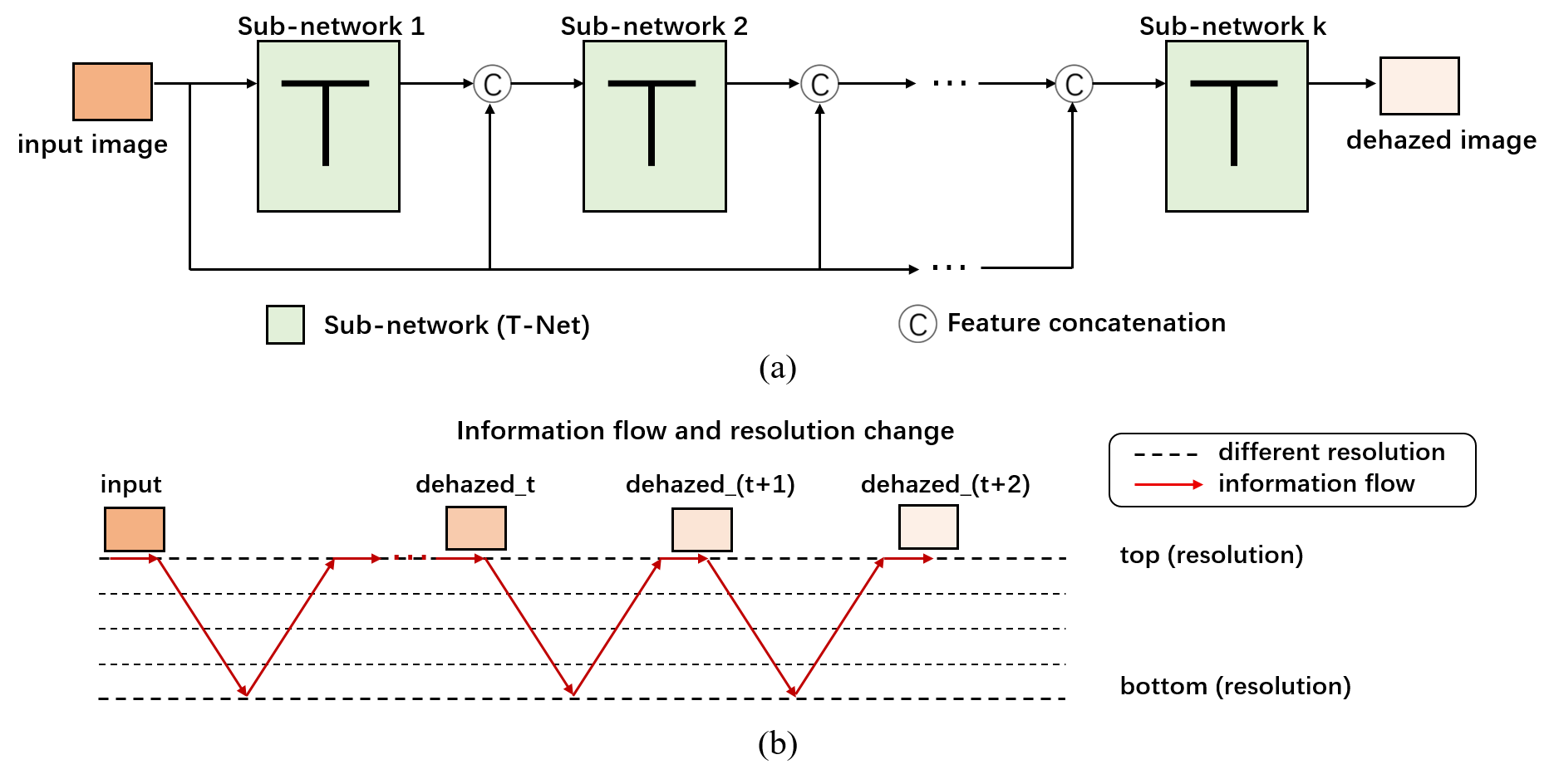}
    \caption{The architecture of Stack T-Net. (a) The detailed structure. (b)The information flow and the feature resolution change among different stages. We use T-Net as the sub-network to build Stack T-Net by repeatedly unfolding it several times. Each dotted line of (b) represents different resolution, and the resolution of each dotted line is 1/4 of it of the above one. Since we use T-Net with four pairs of upsampling and downsampling blocks as sub-network, there are five dotted lines in (b) (containing the original resolution).}
    \label{fig: Stack}
\end{figure*}

In this section, we introduce the overall dehazing network Stack T-Net which is constructed by repeatedly unfolding the plain T-Net, and the loss function used in our method. The first subsection below explains the details of the architecture of Stack T-Net, and Fig.~\ref{fig: Stack} is the illustration a $K$-stage Stack T-Net, where sub-figure (a) exhibits the detailed structure and sub-figure (b) shows the information flow and the feature resolution change among different stages. The second subsection introduces the loss function we used for training the full network.

\subsection{The details of the architecture}
Inspired by~\cite{Yang_2017_CVPR} and~\cite{Ren_2019_CVPR}, we introduce the recurrent structure to our method. In~\cite{Yang_2017_CVPR}, Yang et al. use recurrent network for deraining but just use the deraining result of the previous stage as the input of the next stage, while the performance of deraining is limited by the deraining result of each stage. So we use the strategy of~\cite{Ren_2019_CVPR}, using the concatenation of the output of the previous stage and the original hazy image as input, where the original hazy image can supplement the information loss caused by each dehazing stage. As shown in Fig.~\ref{fig: Stack}, we use T-Net with the same structure in each stage. The inference of Stack T-Net at the $k$-th stage can be formulated as
\begin{equation}
    \begin{split}
        x^k&=f_{in}(x^0,y^{k-1}),\\
        &y^k=f^k_T(x^k),
    \end{split}
\end{equation}
where $x^k$ and $y^k$ separately represent the input and the output of the $k$-th stage, $x^0$ is the original hazy image and $y^0=x^0$, $f_{in}$ stands for the operation of concatenation, $f_T^k$ denotes the mapping of T-Net at the $k$-th stage, and meanwhile $k\in \left\{1,\cdots,K\right\}$. T-Net of each stage plays the role of a dehazing sub-network, which directly learns the mapping from hazy image to dehazy results. Thereby, the input of each stage is a six-channel image which is the concatenation of two three-channel RGB images, and the output is a three-channel RGB image. Meanwhile, in order to be consistent with other stages, the first stage uses the concatenation of two identical original hazy images as input. And we choose the output of the last stage as the final dehazing result. 

However, the network of recurrent structure has considerably large amount of parameters to learn, which take a lot of memory and decrease the efficiency of dehazing. And the more complex the structure of network is, the more likely it leads to over-fitting. Witnessing the success of recursive computation in image restoration~\cite{Kim_2016_CVPR,Li_2018_ECCV,Tai_2017_CVPR}, we apply it to our method. In our practice, we utilize the inter-stage recursive computation instead of repeatedly unfolding the plain network. Since inter-stage recursive computation requires each stage to share the same network parameters, each stage can be expressed as
\begin{equation}
    f_T^k(\cdot)=f_T(\cdot).
\end{equation}
Experimental results verify that this recursive strategy can improve the dehazing effectiveness without designing a deeper and more complex network.

\subsection{Loss function}
The loss function of our method measures the error of the dehazing results at each-stage, and is used to train the proposed network. The MSE loss has smooth function curve, which is convenient for the use of the gradient descent algorithm. But it is more sensitive to outliers and easier to cause gradient explosions than the $L_1$ loss. Therefore, we employ the smooth $L_1$ loss which combines the MSE loss and the $L_1$ loss to enhance the robustness of the network. 

Let $I_c(x)$ denote the intensity of the $c$-th color channel at pixel $x$ in the ground truth, and $J_c(x)^k$ represents the intensity of the same position of the dehazed image in the $k$-th stage. $K$ and $N$ separately denote the number of the stages and the number of pixels. The smooth $L_1$ loss of the $k$-th stage ($L_{SL_1}^{k}$) can be defined as
\begin{equation}
    L_{SL_1}^{k}=\frac{1}{N}\sum_{x=1}^{N}\sum_{c=1}^{3}f_{SL_1}(\left|J_c^k(x)-I_c(x)\right|),
\end{equation}
where 
\begin{equation}
    f_{SL_1}(e)=
    \begin{cases}
        0.5e^2,\quad 0\leq e<1,\\
        e-0.5,\quad e\ge1.
    \end{cases}
\end{equation}
When the error between the dehazed result and the ground truth exceeds the limited value, the $L_1$ norm is used instead of the $L_2$ norm to reduce the influence of outliers on the network parameters.

The perceptual loss~\cite{2016Perceptual} measures image visual similarities between the dehazed image and the ground truth more effectively than the pixel-wise loss, for which we take advantage of the perceptual loss to strengthen the finer details of the dehazed images. By leveraging multi-scale features extracted from VGG16~\cite{simonyan2015deep} pre-trained on ImageNet~\cite{2015ImageNet}, the perceptual loss of the $k$-th dehazed stage can be defined as
\begin{equation}
    L_p^k=\sum_{j=1}^3\frac{1}{C_jH_jW_j}\left\|F_{j}(J^k)-F_j(I)\right\|_2^2,
\end{equation}
where $J^k$ represents the dehazed result of the $k$-th stage, $I$ represents the ground truth, each $F_{j}(\cdot)$ ($j =\left\{1,2,3\right\}$) separately denotes the output feature of relu1-2, relu2-2, relu3-3 layer of VGG-16 and $C_j$, $H_j$ and $W_j$ specify the dimension of these features.

By combining the smooth L1 loss and the perceptual loss, the total loss of all stages is defined as
\begin{equation}
    L=\sum_{k=1}^KL_{SL_1}^k+\lambda \sum_{k=1}^{K}L_p^k,
\end{equation}
that is,
\begin{equation}
    L=L_{SL_1}+\lambda L_P,
\end{equation}
where $\lambda$ is set to $0.04$ for controlling the relative weights on the two loss components.

\section{Experiments}
\label{sec: exp}
In this section, quantitative and qualitative experimental results are shown to demonstrate the effectiveness of the proposed method. We conduct a series of experiments to compare the performance of our method with the state-of-the-art approaches on both synthetic and real-world datasets. Moreover, we carry out an ablation study to demonstrate the effectiveness of each module of our network.

\subsection{Datasets}
In the training phase, we use the datasets of~\cite{Shao_2020_CVPR} as training set. This dataset contains $6000$ synthetic hazy images from the RESIDE dataset~\cite{8451944}, where the RESIDE dataset~\cite{8451944} contains both synthesized and real-world hazy/clean image pairs of indoor and outdoor scenes and is split into five subsets, namely, ITS (Indoor Training Set), OTS (Outdoor Training Set), SOTS (Synthetic Object Testing Set), URHI (Unannotated Real Hazy Images), and RTTS (Real Task-driven Testing Set). Among the $6000$ images, $3000$ are chosen from the ITS and the rest is from the OTS. All the images are randomly cropped to $256\times256$ and randomly flipped for data augmentation, and then the pixel values are normalized to $[-1, 1]$. Since our training set is a subset of ITS and OTS, we use SOTS as the total testing set in the test phase to compare the performance of our proposed method with other methods. Furthermore, real-world hazy images are chosen from URHI to evaluate the generalization of our method in the real-world scenery.

\subsection{Network implementation}
We implement our framework with Pytorch~\cite{Paszke2017AutomaticDI} and exploit the Adam optimizer~\cite{kingma2017adam} with a batch size $14$ to train the network, where the momentum $\beta_1$ and $\beta_2$ adopt the default values of $0.9$ and $0.999$, respectively. We train every model for $2000$ epochs in total. The learning rate is initially set to $0.001$, reduced by half every $20$ epochs, and kept fixed as $0.0001$ from the $80$-th epoch. Our proposed method is evaluated against the following state-of-the-art approaches: DCP~\cite{5567108}, MSCNN~\cite{2016Single}, DehazeNet~\cite{2016DehazeNet}, NLD~\cite{Berman_2016_CVPR}, AOD-Net~\cite{8237773}, GFN~\cite{Ren_2018_CVPR}, DCPDN~\cite{8578435}, EPDN~\cite{8953692}, and DA$\_$dehazing~\cite{Shao_2020_CVPR}. We adopt peak-signal-to-noise ratio (PSNR) and structural similarity measure (SSIM) as quantitative evaluation indexes, where PSNR can measure the pixel to pixel difference and SSIM can quantify the structural difference between a dehazed output image and the corresponding ground truth. The results of all experiments are shown in the following. 

\subsection{Ablation study}

To evaluate the effectiveness of several key modules in our network, we perform ablation studies with the following three strategies.

\textbf{The first ablation study} is conducted to determine the configurations of the backbone module of the proposed T-Net, that is, the number of upsampling and downsampling blocks pairs and the number of RDB blocks pairs in the trunk road. RDB blocks are used as feature generator in our network. Simply put, the more RDB block pairs we use, the deeper the network and the deeper the features we can extract. The position attention block contains a matrix multiplication of $(N, C)\times(C, N)$, where $N$ is the number of the pixels in each channel of the feature. So it is easy to cause out of memory if the input feature size of this block is too large. We set the initial pairs of upsampling and downsampling blocks as 2 to avoid this problem. Table~\ref{tab: tabel1} shows the performance of T-Net with different configurations on the SOTS dataset, where $m$ and $n$ respectively represent the number of upsampling and downsampling block pairs and the number of RDB block pairs in the trunk road. 

\newcolumntype{I}{!{\vrule width 0.8pt}}
\newcommand{\PreserveBackslash}[1]{\let\temp=\\#1\let\\=\temp}
\newcolumntype{C}[1]{>{\PreserveBackslash\centering}p{#1}}
\newcolumntype{R}[1]{>{\PreserveBackslash\raggedleft}p{#1}}
\newcolumntype{L}[1]{>{\PreserveBackslash\raggedright}p{#1}}
\renewcommand\arraystretch{1.25}

\begin{table}[tbp]
\centering
\caption{Comparison on the SOTS dataset for T-Net with different configurations.}
\begin{tabular}{Ic|c|c|cI}
\Xhline{0.8pt}
\multicolumn{2}{Ic|}{Configuration} & \multicolumn{2}{cI}{SOTS}\\
\Xhline{0.8pt}
\multicolumn{1}{IC{1.2cm}|}{$m$} &\multicolumn{1}{C{1.2cm}|}{$n$} & \multicolumn{1}{C{1.5cm}|}{PSNR} & \multicolumn{1}{C{1.5cm}I}{SSIM} \\
\Xhline{0.8pt}
\multirow{2}[3]{*}{2} & 1     &22.90 &0.8954 \\
\cline{2-4}           & 2     &26.86 &0.9451 \\ 
\cline{2-4}           & 3     &27.43 &0.9473 \\ 
\hline
\multirow{2}[3]{*}{3} & 1     &22.97 &0.8949 \\ 
\cline{2-4}           & 2     &27.49 &0.9520 \\ 
\cline{2-4}           & 3     &28.13 &0.9536 \\ 
\hline
\multirow{2}[3]{*}{4} & 1     &22.86 &0.8932\\ 
\cline{2-4}           & 2     &28.30 &0.9535\\ 
\cline{2-4}           & 3     &\textbf{28.55} &\textbf{0.9543}\\ 
\Xhline{0.8pt}
\end{tabular}%
\label{tab: tabel1}
\end{table}%

There is a tendency shown in Table~\ref{tab: tabel1} that the average PSNR and SSIM values increase as $m$ and $n$ increase. The exception happens when $n=1$, i.e., the performance of T-Net does not get improved with the increase of $m$ in this case. The reason is that, the representation ability of features is influenced by the number of RDB block pairs in the trunk road. When $n$ is set to 1, the features extracted by our network contains insufficient deep discriminant information, which makes the fitting ability of neural network badly limited. Moreover, when $n>1$, the performance of adding a pair of upsampling and downsampling blocks is better than adding a pair of RDB blocks (see e.g., (3,2) versus (2,3), (4,2) versus (3,3) in the form of (m,n)), which verifies the effectiveness of multi-scale features. Compared with single-scale features, multi-scale features can provide more discriminant information which is helpful for image understanding. As shown in Table~\ref{tab: tabel1}, the average PSNR and SSIM values are the highest when $m=4, n=3$.Therefore, we use a T-Net with four pairs of upsampling and downsampling blocks and three pairs of RDB blocks in the trunk road as our network in the following experiments.

\begin{figure*}[tbp]
	\scriptsize
	\centering
	\renewcommand{\tabcolsep}{1pt} 
	\renewcommand{\arraystretch}{1}
	\begin{center}
		\begin{tabular}{cccccccccc}
			\includegraphics[width=0.095\linewidth]{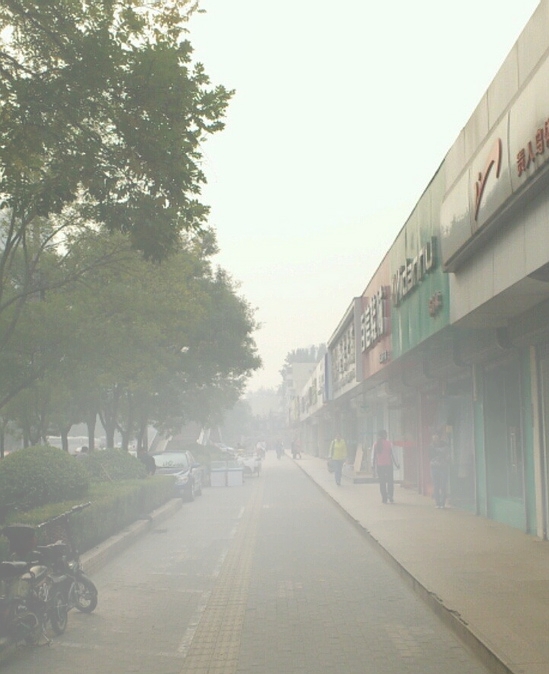} &
			\includegraphics[width=0.095\linewidth]{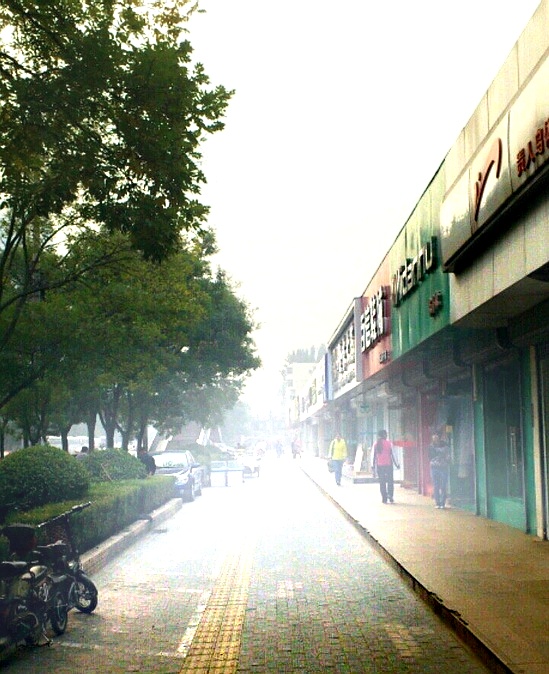} &
			\includegraphics[width=0.095\linewidth]{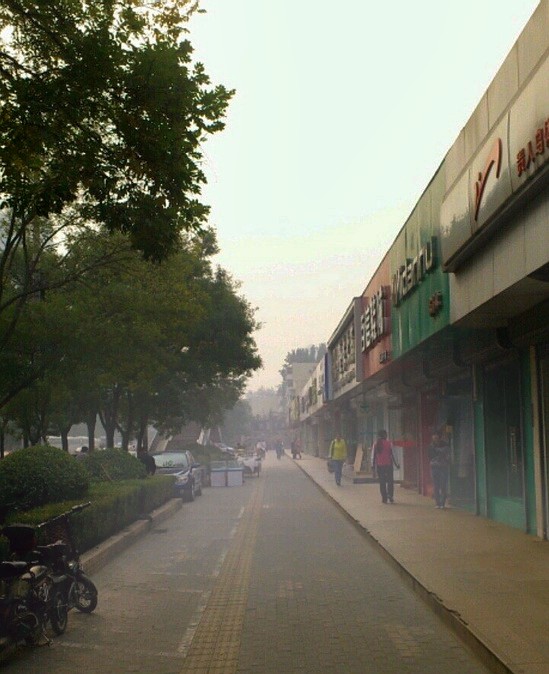} &
			\includegraphics[width=0.095\linewidth]{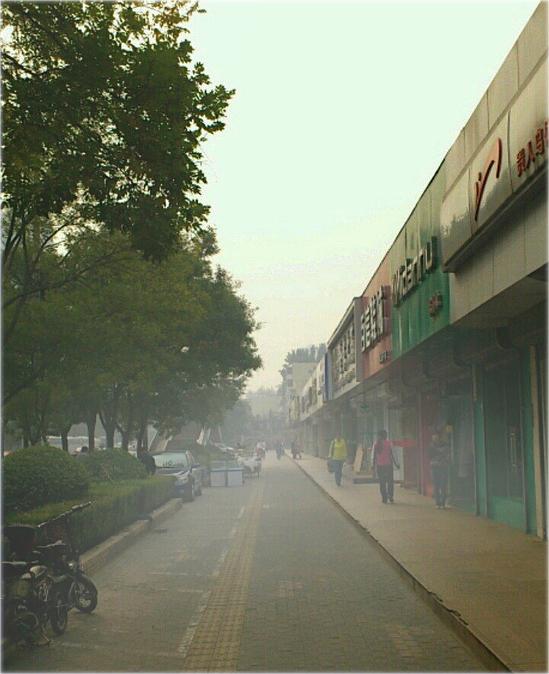} &
			\includegraphics[width=0.095\linewidth]{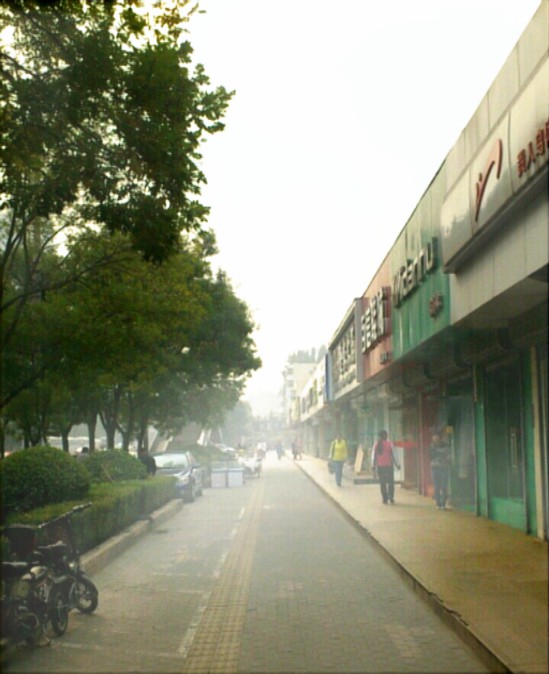} &
			\includegraphics[width=0.095\linewidth]{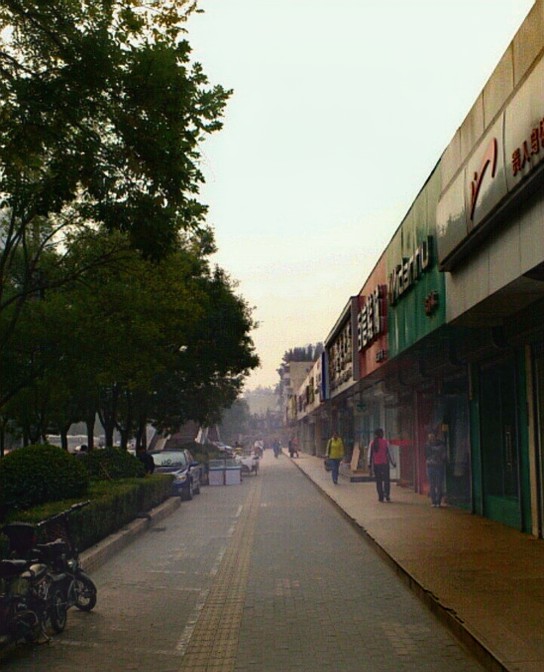} &
			\includegraphics[width=0.095\linewidth]{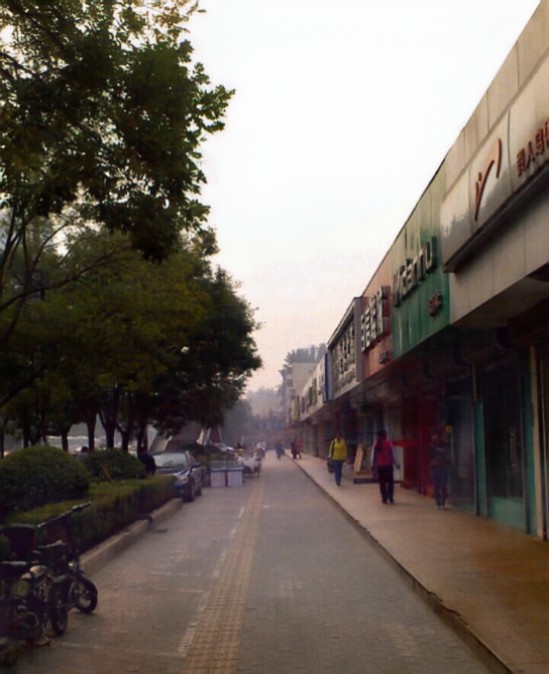} &
			\includegraphics[width=0.095\linewidth]{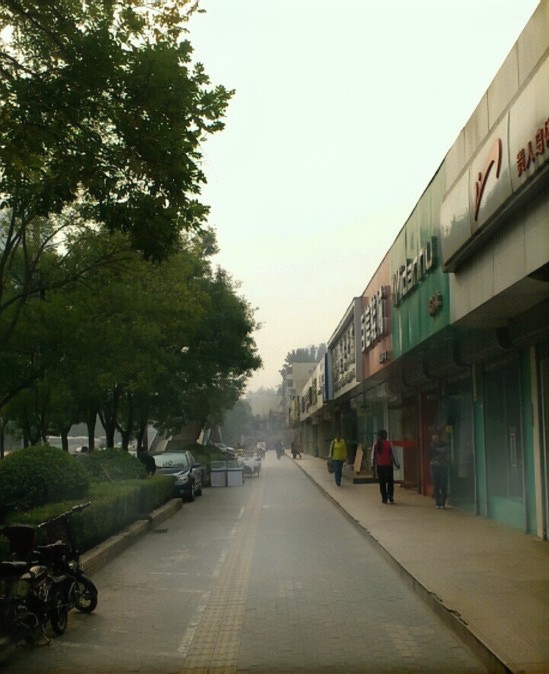} &
			\includegraphics[width=0.095\linewidth]{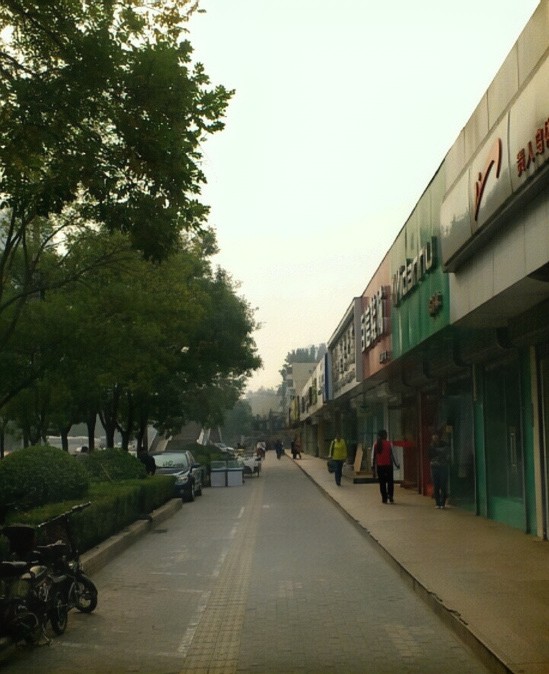} &
			\includegraphics[width=0.095\linewidth]{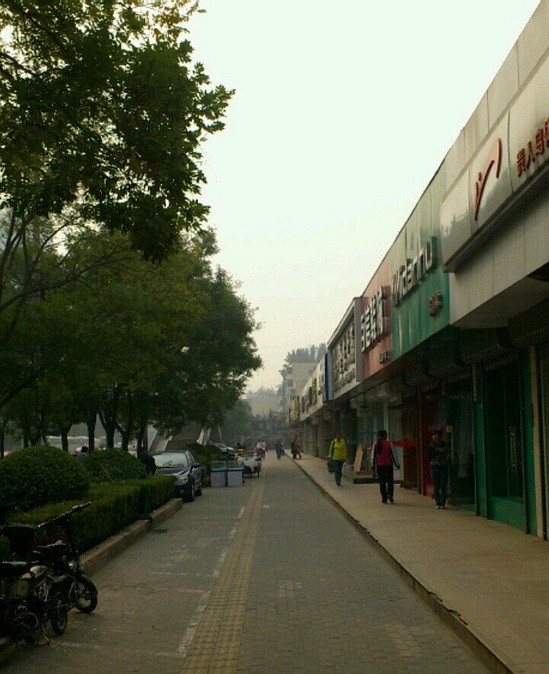} \\

			\includegraphics[width=0.095\linewidth]{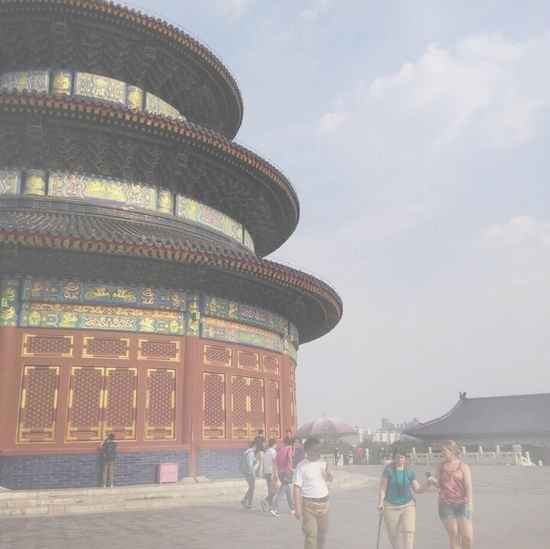} &
			\includegraphics[width=0.095\linewidth]{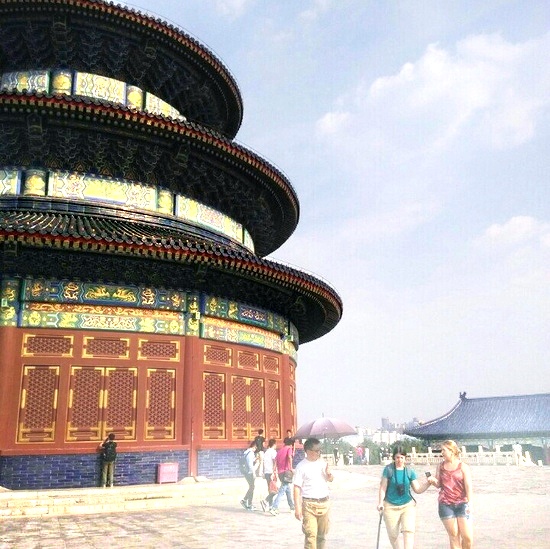} &
			\includegraphics[width=0.095\linewidth]{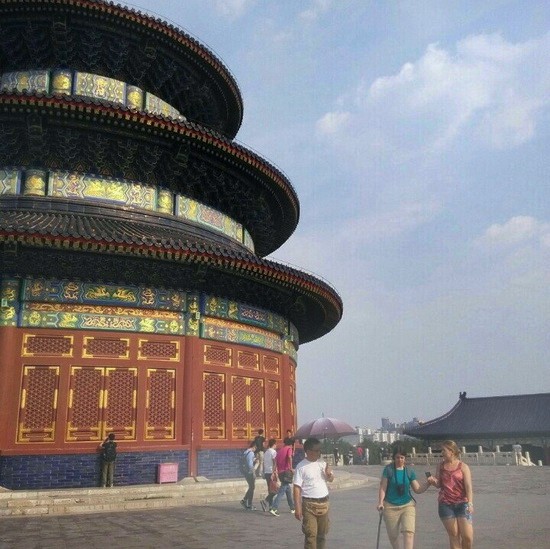} &
			\includegraphics[width=0.095\linewidth]{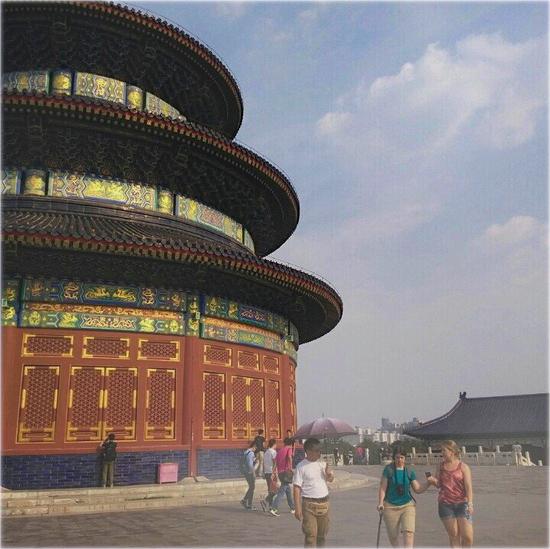} &
			\includegraphics[width=0.095\linewidth]{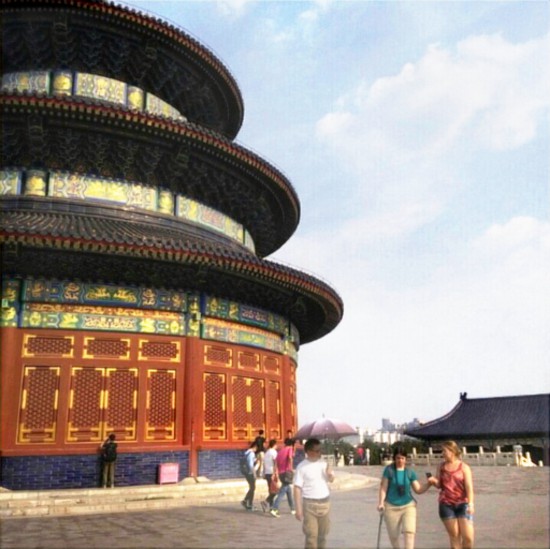} &
			\includegraphics[width=0.095\linewidth]{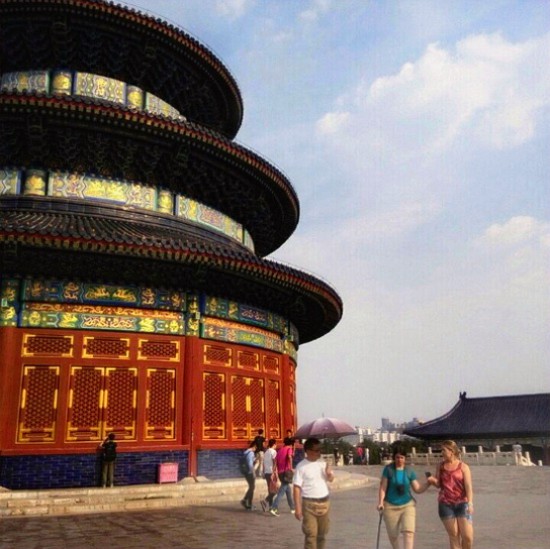} &
			\includegraphics[width=0.095\linewidth]{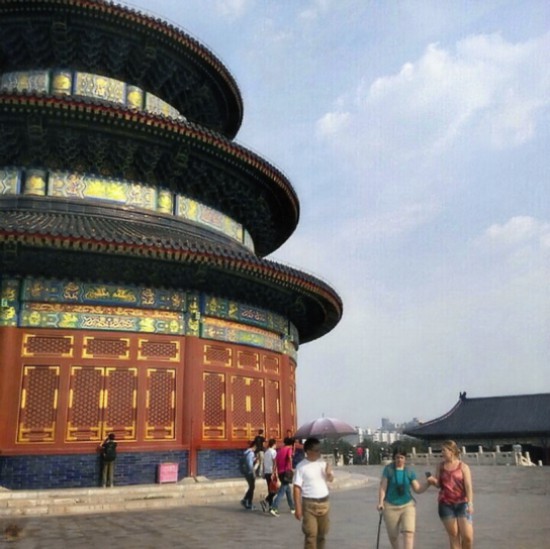} &
			\includegraphics[width=0.095\linewidth]{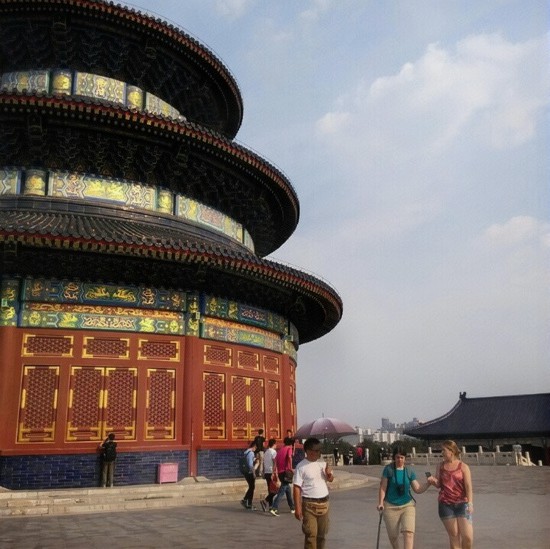} &
			\includegraphics[width=0.095\linewidth]{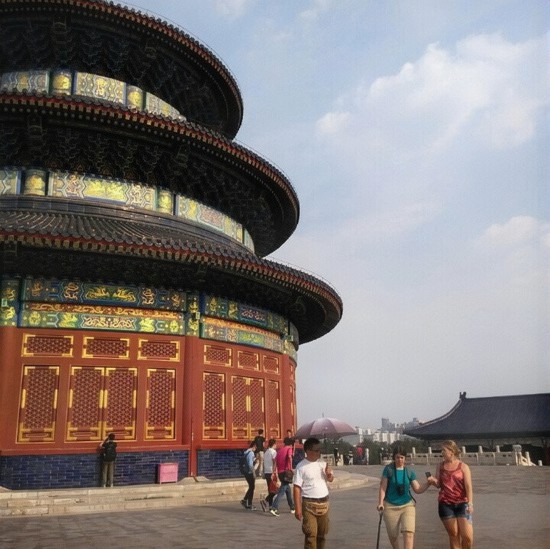} &
			\includegraphics[width=0.095\linewidth]{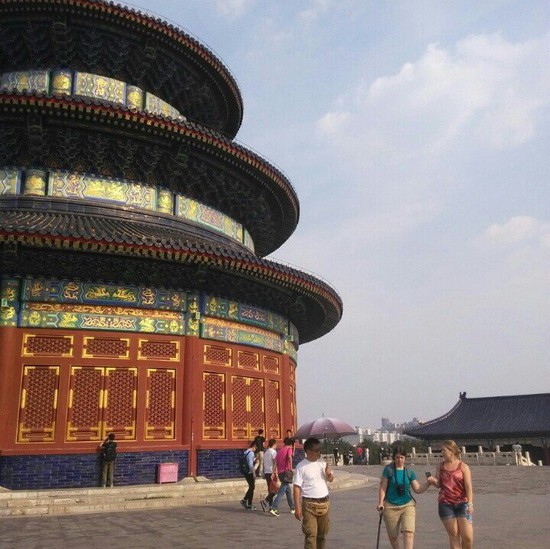} \\
			
			\includegraphics[width=0.095\linewidth]{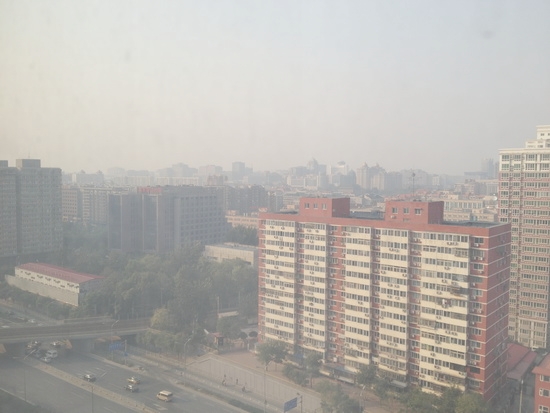} &
			\includegraphics[width=0.095\linewidth]{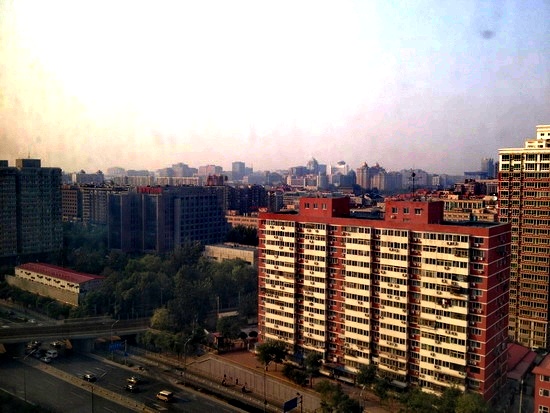} &
			\includegraphics[width=0.095\linewidth]{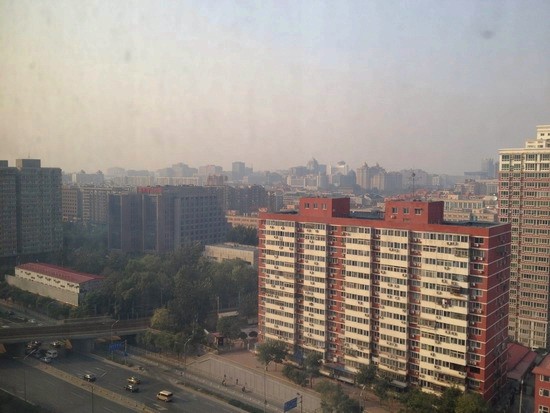} &
			\includegraphics[width=0.095\linewidth]{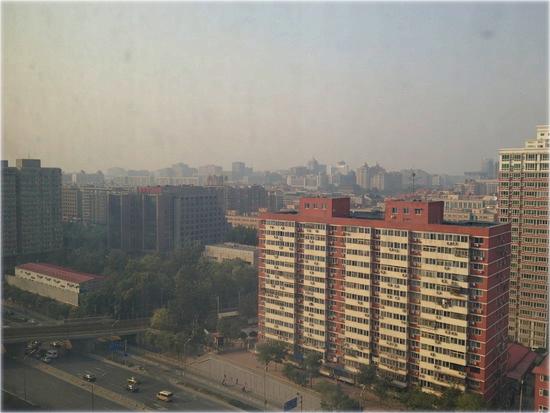} &
			\includegraphics[width=0.095\linewidth]{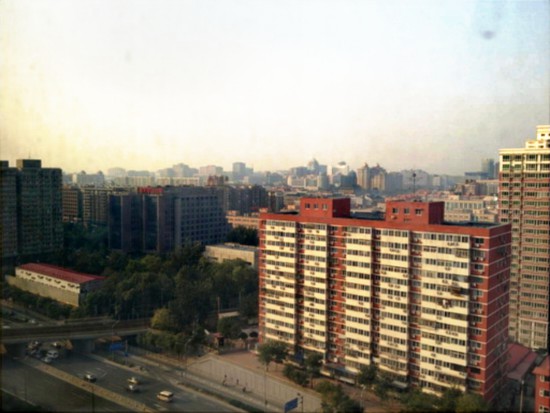} &
			\includegraphics[width=0.095\linewidth]{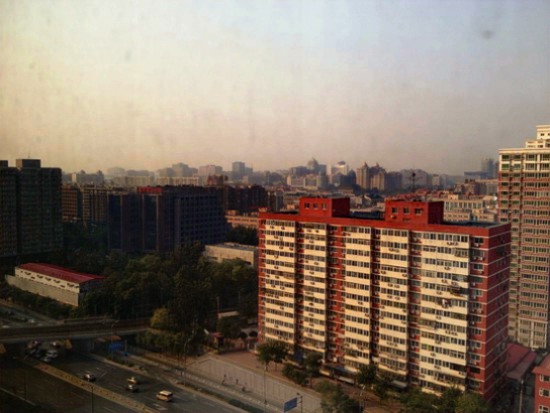} &
			\includegraphics[width=0.095\linewidth]{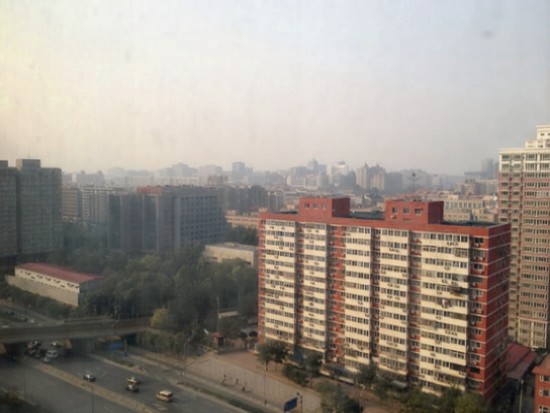} &
			\includegraphics[width=0.095\linewidth]{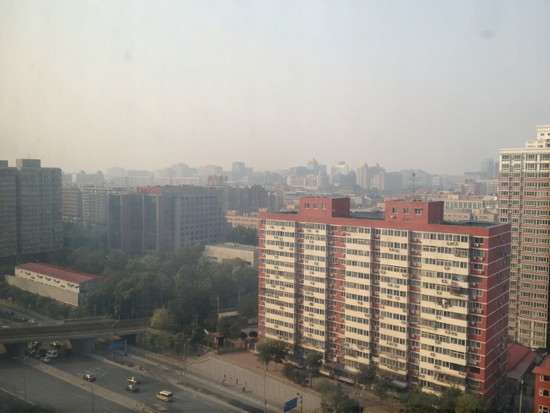} &
			\includegraphics[width=0.095\linewidth]{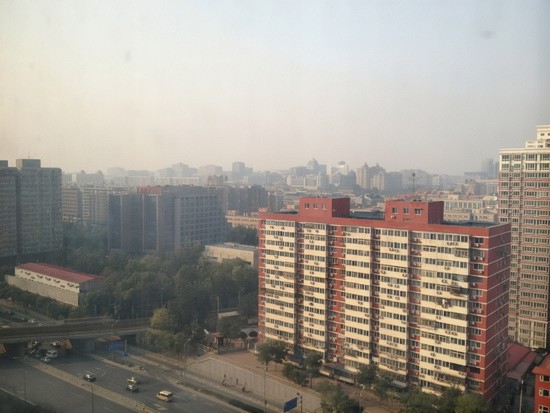} &
			\includegraphics[width=0.095\linewidth]{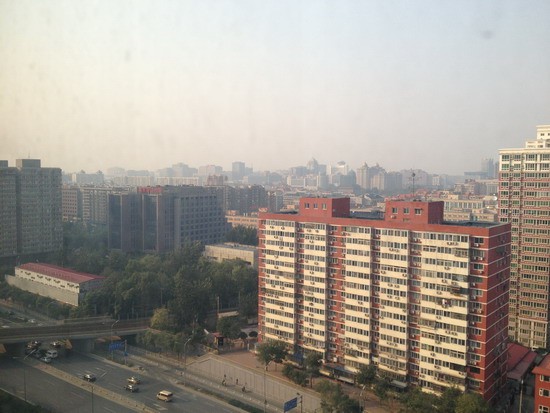} \\
			
			\includegraphics[width=0.095\linewidth]{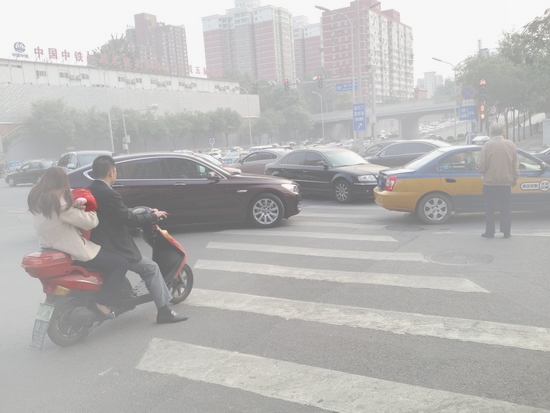} &
			\includegraphics[width=0.095\linewidth]{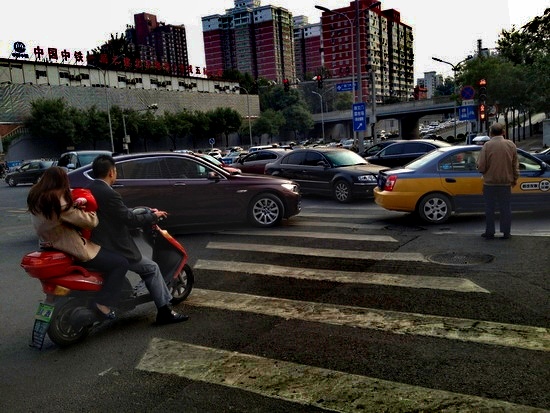} &
			\includegraphics[width=0.095\linewidth]{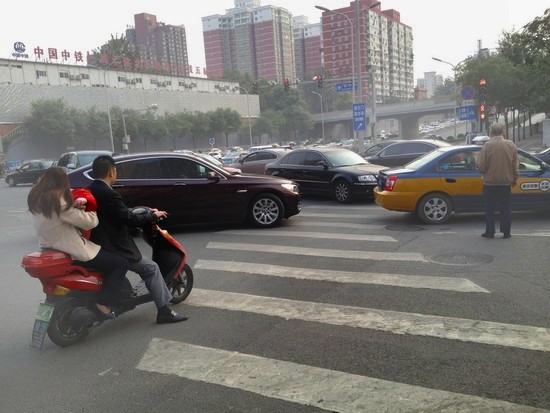} &
			\includegraphics[width=0.095\linewidth]{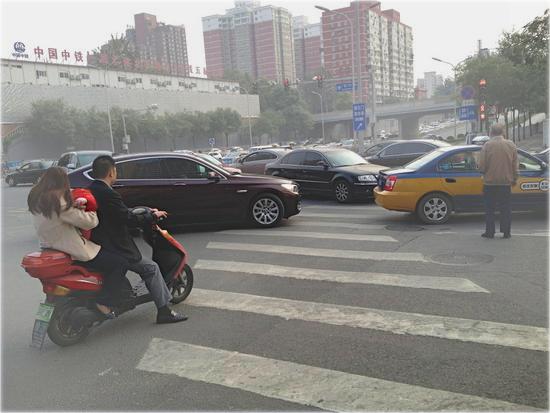} &
			\includegraphics[width=0.095\linewidth]{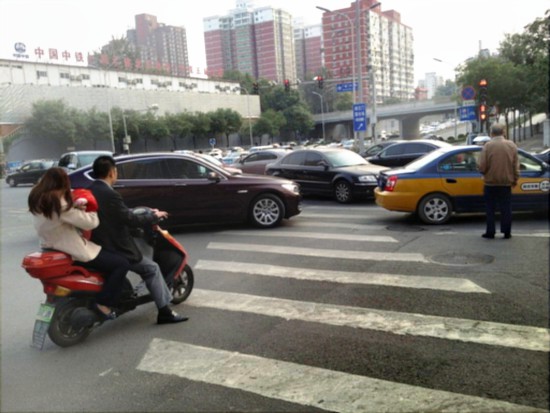} &
			\includegraphics[width=0.095\linewidth]{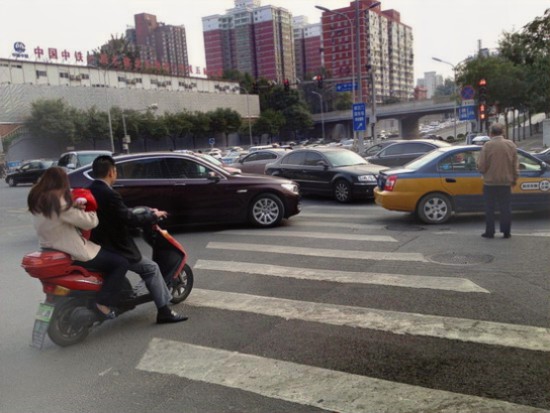} &
			\includegraphics[width=0.095\linewidth]{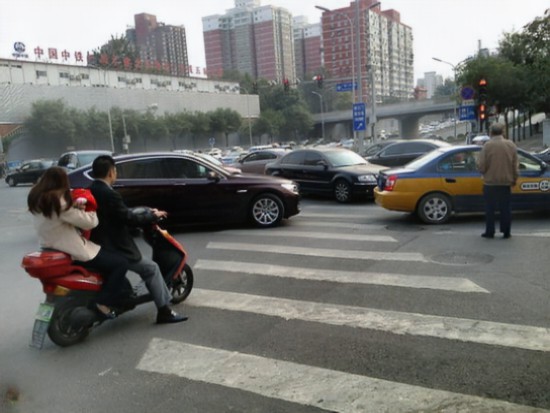} &
			\includegraphics[width=0.095\linewidth]{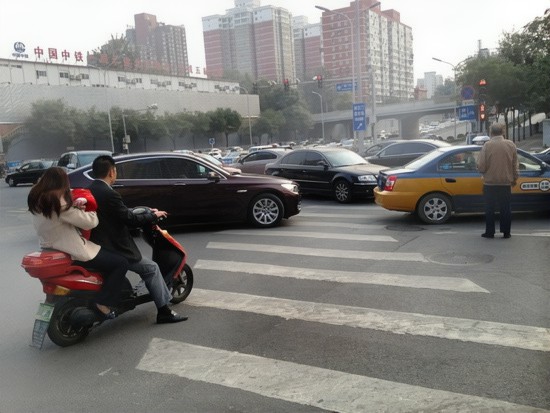} &
			\includegraphics[width=0.095\linewidth]{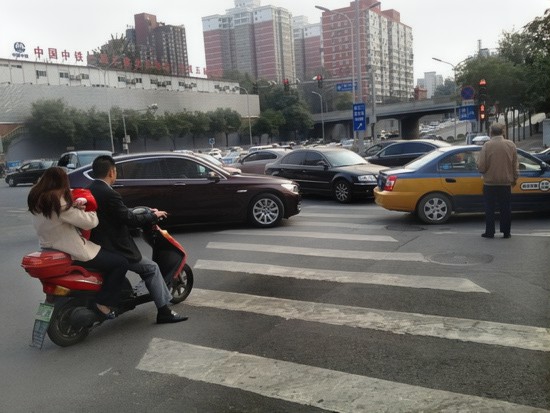} &
			\includegraphics[width=0.095\linewidth]{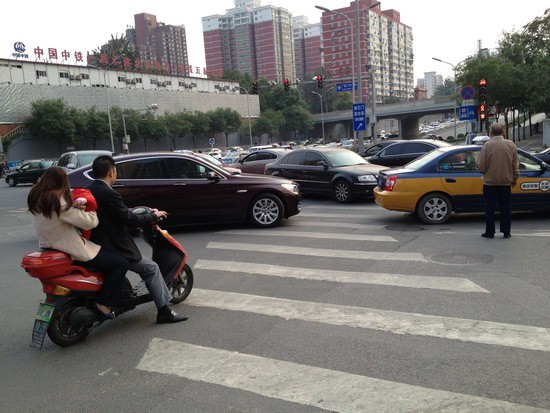} \\
			
			\includegraphics[width=0.095\linewidth]{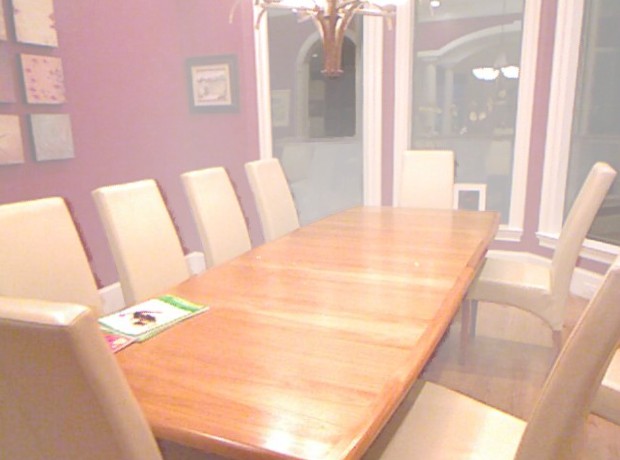} &
			\includegraphics[width=0.095\linewidth]{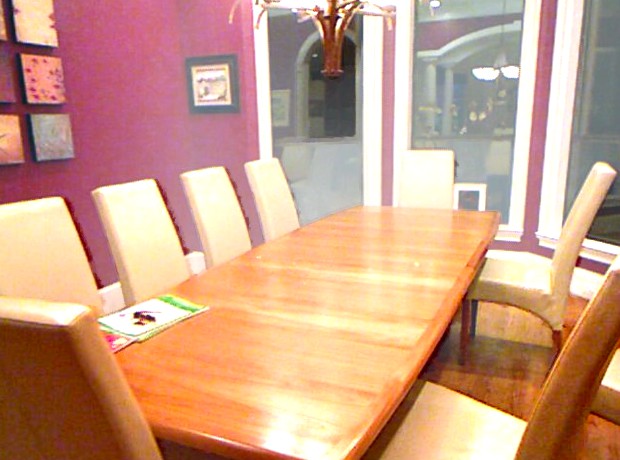} &
			\includegraphics[width=0.095\linewidth]{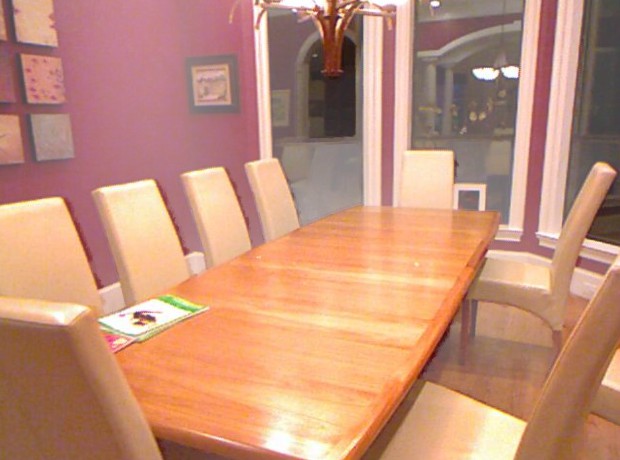} &
			\includegraphics[width=0.095\linewidth]{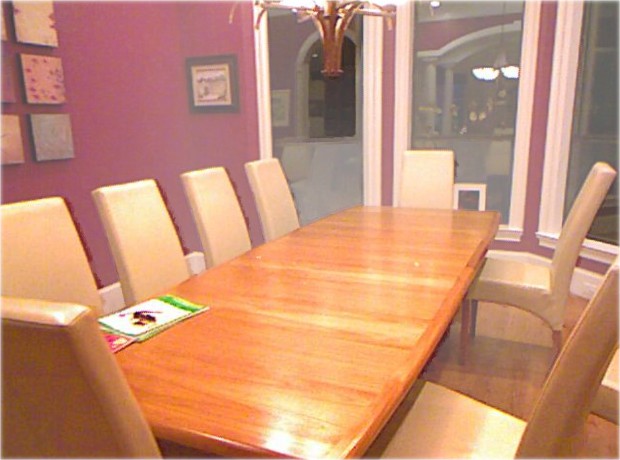} &
			\includegraphics[width=0.095\linewidth]{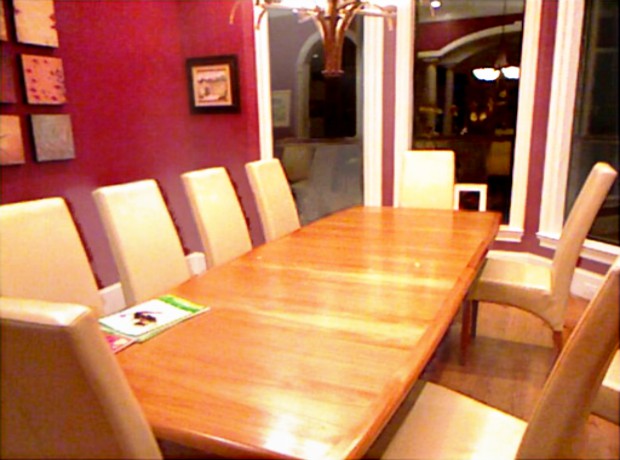} &
			\includegraphics[width=0.095\linewidth]{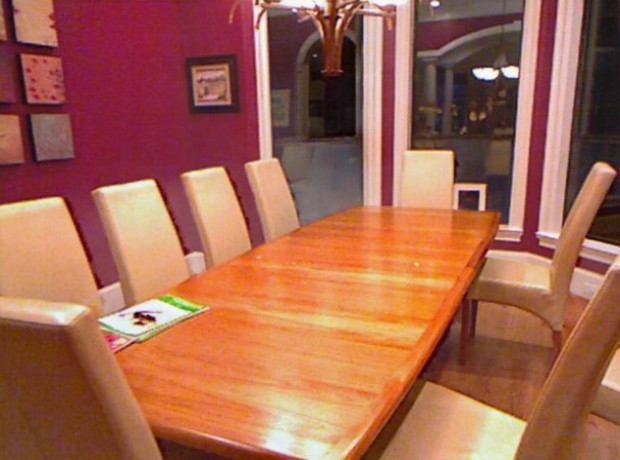} &
			\includegraphics[width=0.095\linewidth]{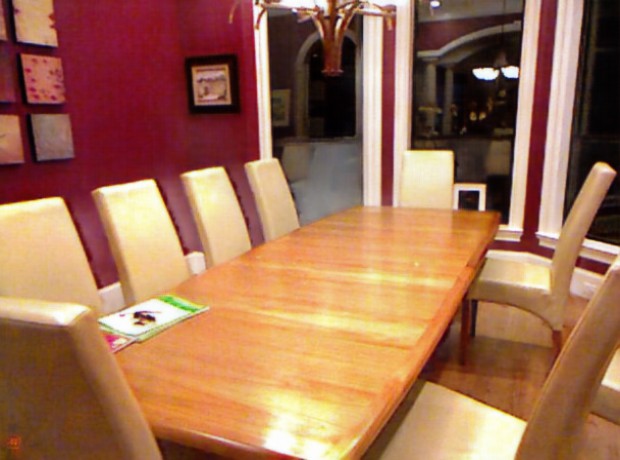} &
			\includegraphics[width=0.095\linewidth]{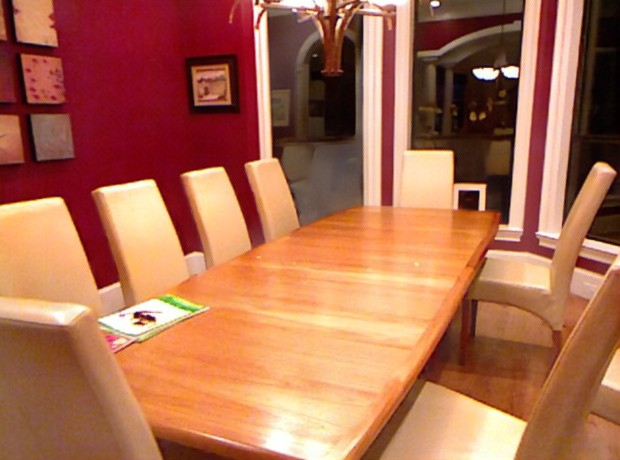} &
			\includegraphics[width=0.095\linewidth]{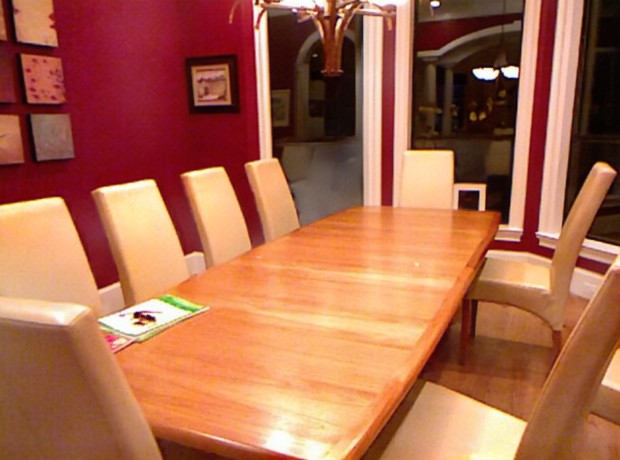} &
			\includegraphics[width=0.095\linewidth]{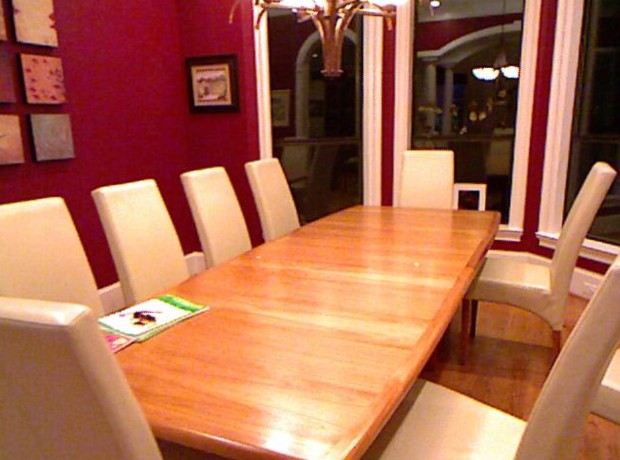} \\
			
			\includegraphics[width=0.095\linewidth]{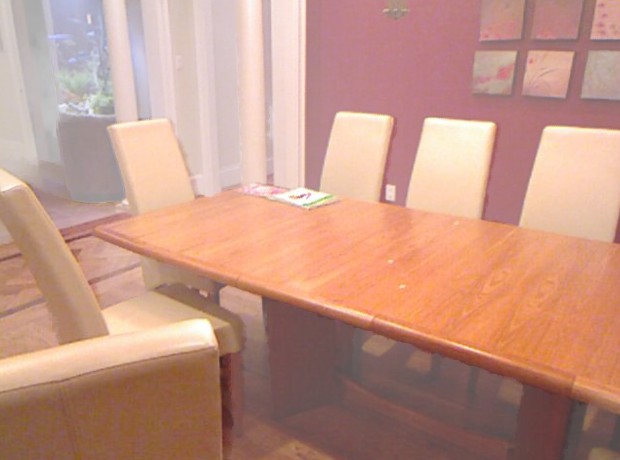} &
			\includegraphics[width=0.095\linewidth]{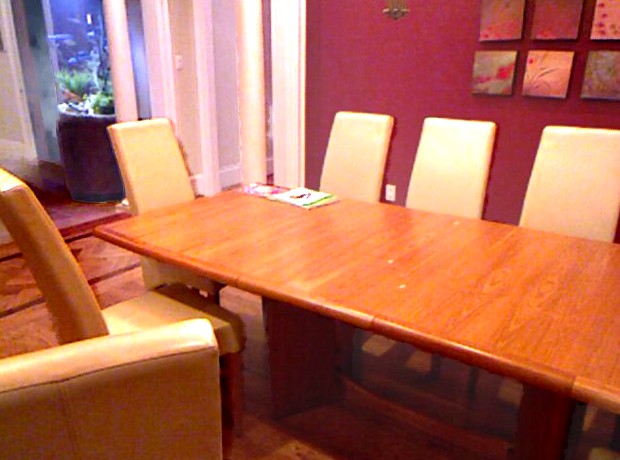} &
			\includegraphics[width=0.095\linewidth]{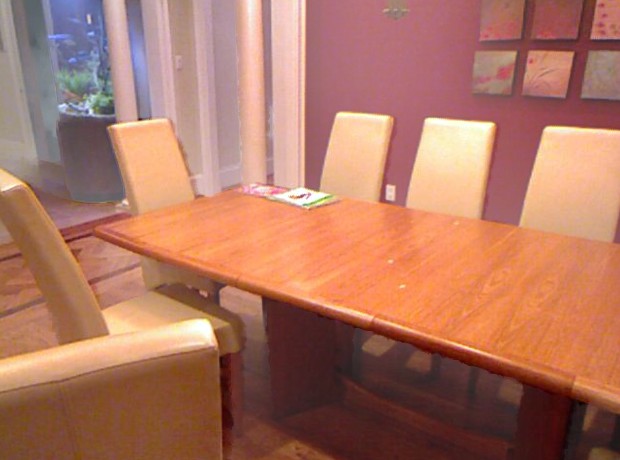} &
			\includegraphics[width=0.095\linewidth]{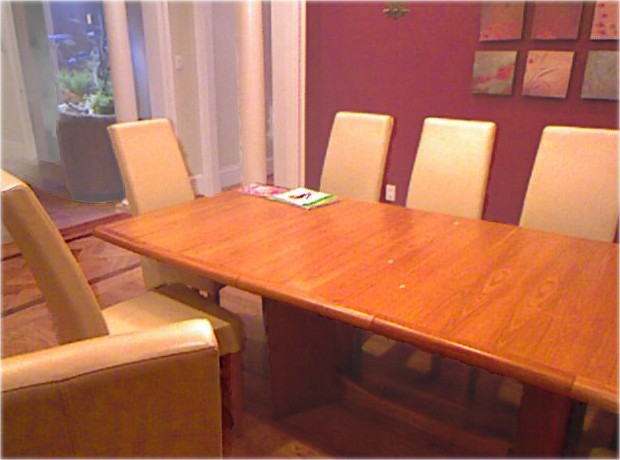} &
			\includegraphics[width=0.095\linewidth]{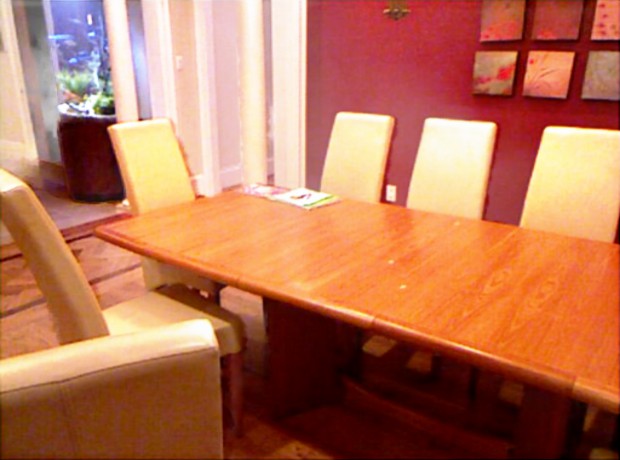} &
			\includegraphics[width=0.095\linewidth]{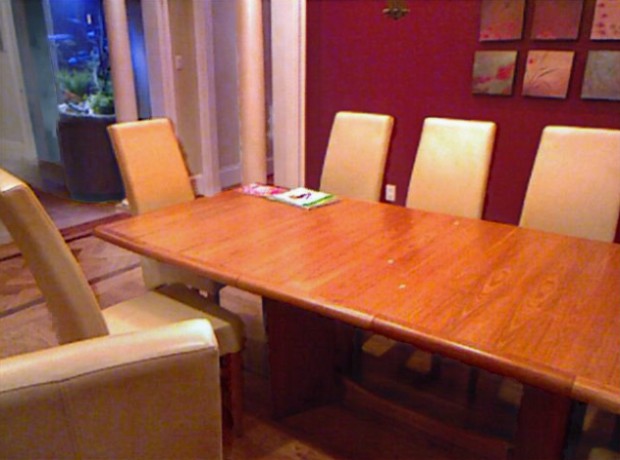} &
			\includegraphics[width=0.095\linewidth]{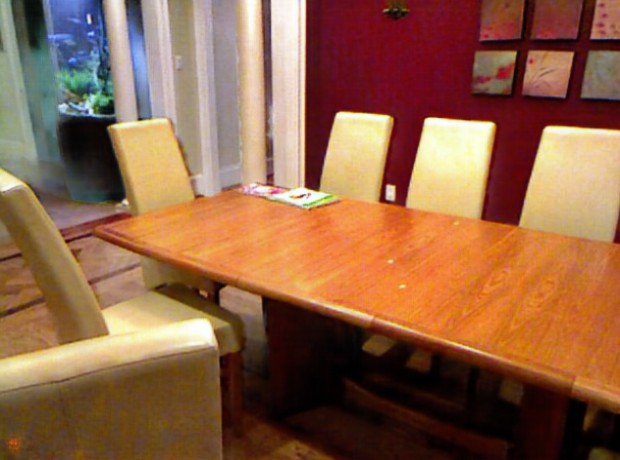} &
			\includegraphics[width=0.095\linewidth]{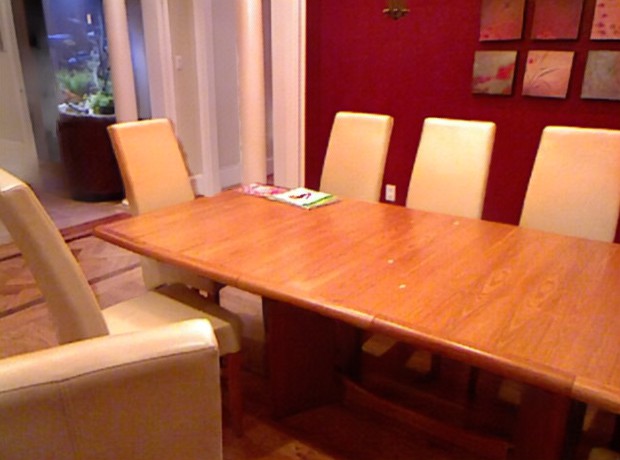} &
			\includegraphics[width=0.095\linewidth]{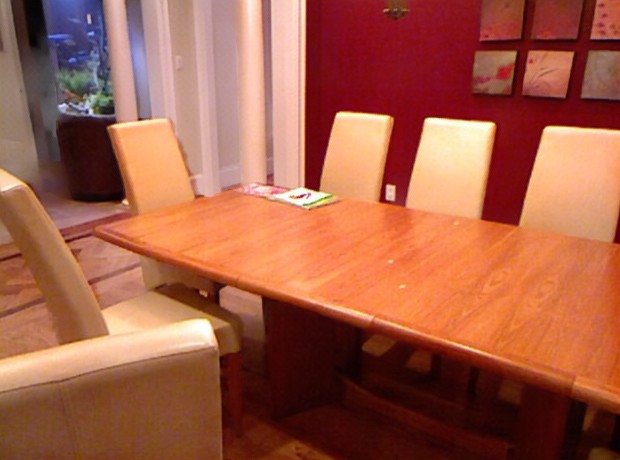} &
			\includegraphics[width=0.095\linewidth]{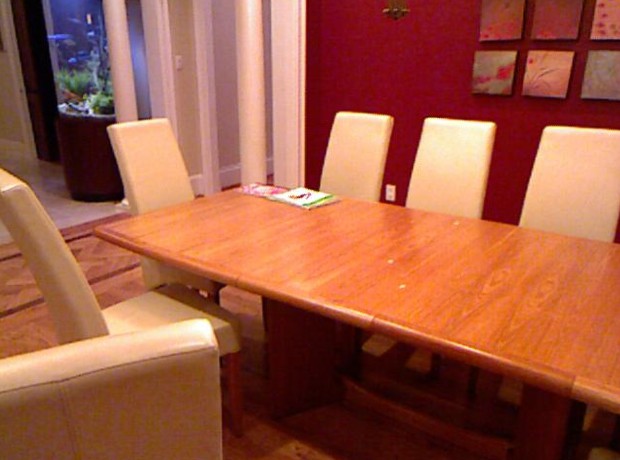} \\
			
			\includegraphics[width=0.095\linewidth]{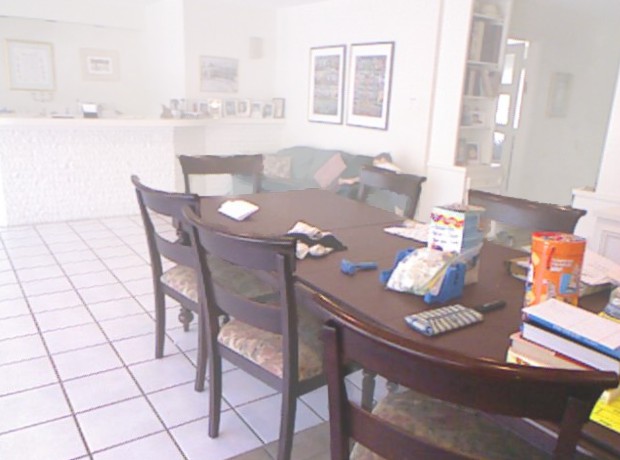} &
			\includegraphics[width=0.095\linewidth]{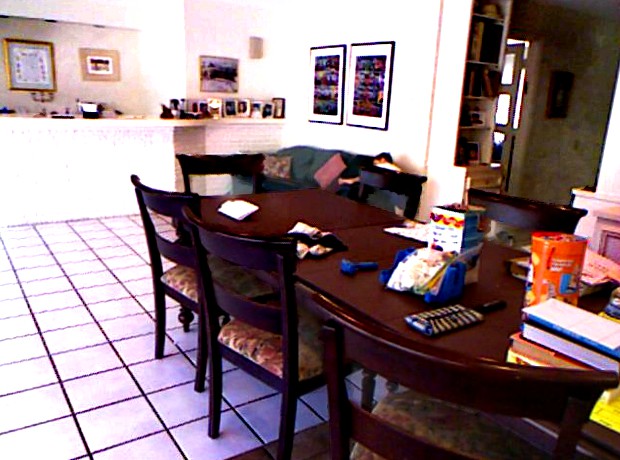} &
			\includegraphics[width=0.095\linewidth]{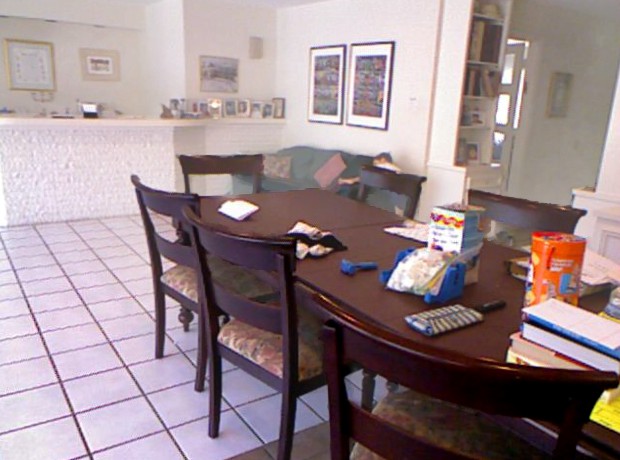} &
			\includegraphics[width=0.095\linewidth]{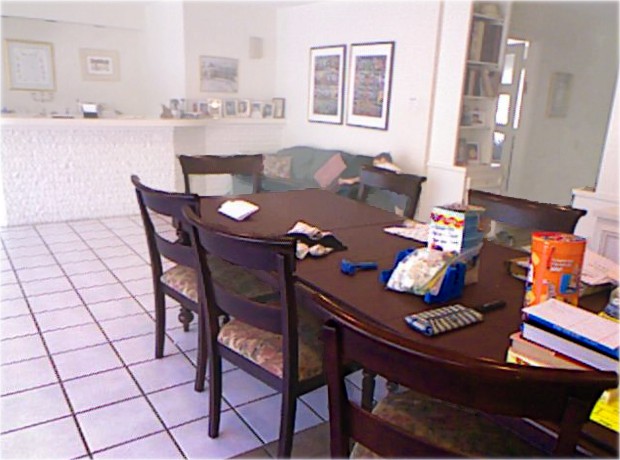} &
			\includegraphics[width=0.095\linewidth]{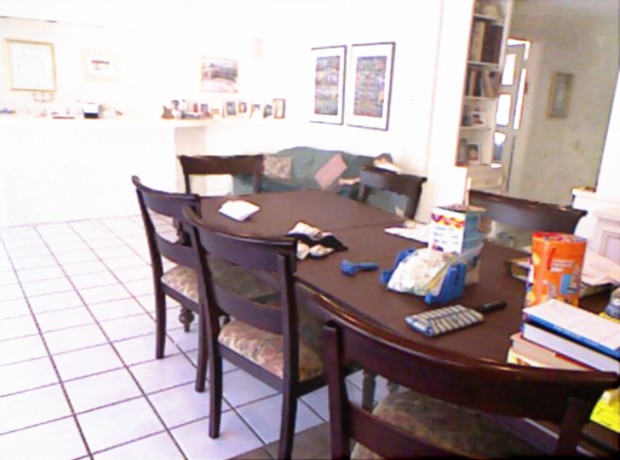} &
			\includegraphics[width=0.095\linewidth]{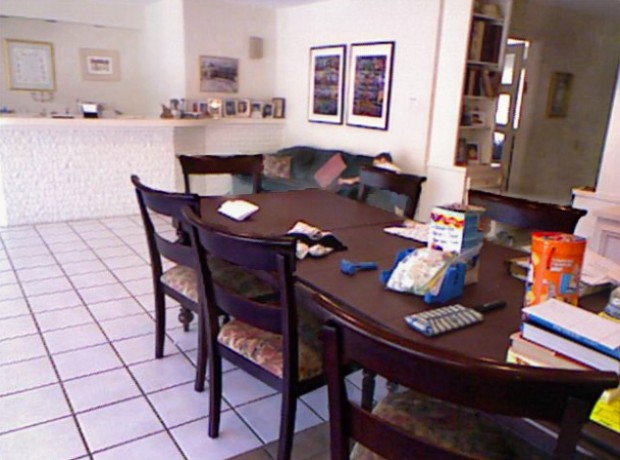} &
			\includegraphics[width=0.095\linewidth]{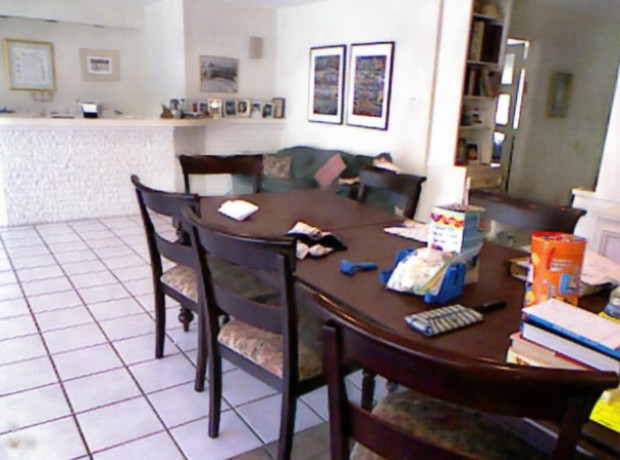} &
			\includegraphics[width=0.095\linewidth]{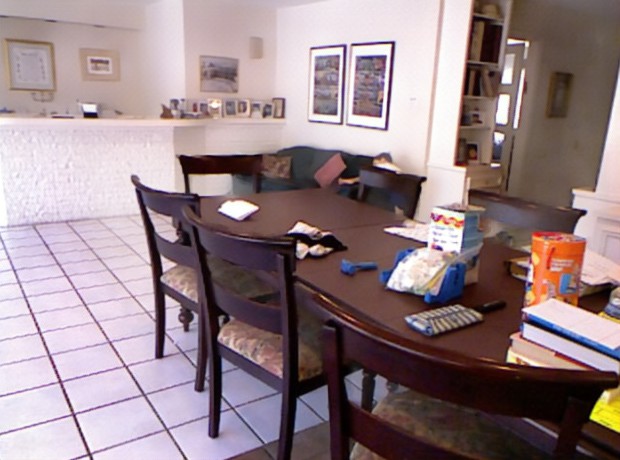} &
			\includegraphics[width=0.095\linewidth]{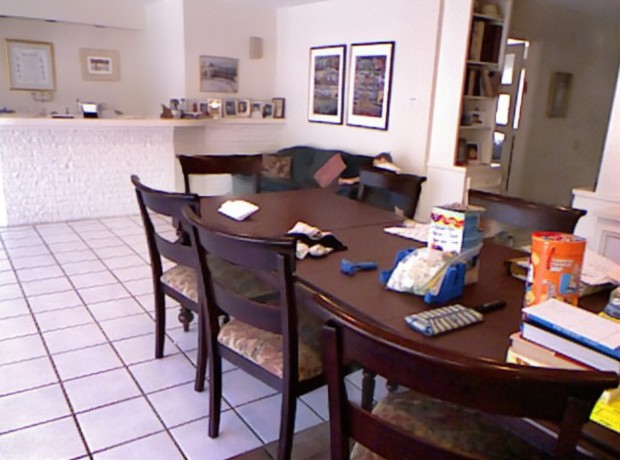} &
			\includegraphics[width=0.095\linewidth]{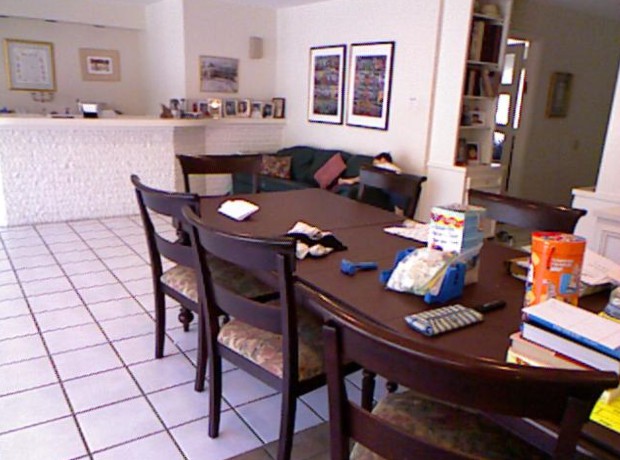} \\
			
			\includegraphics[width=0.095\linewidth]{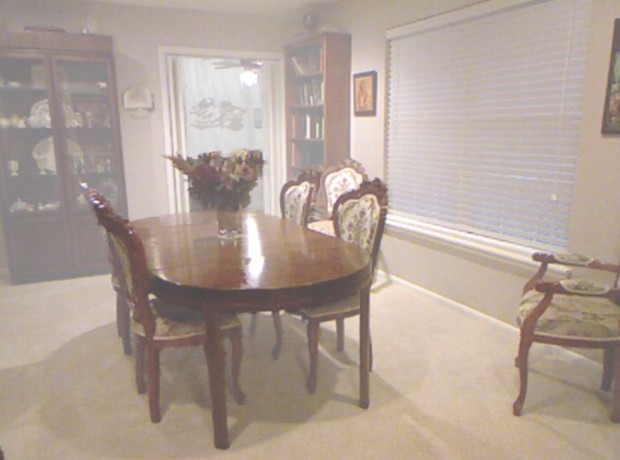} &
			\includegraphics[width=0.095\linewidth]{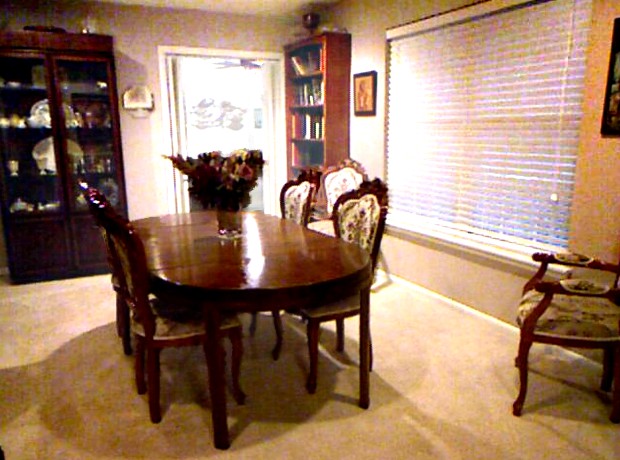} &
			\includegraphics[width=0.095\linewidth]{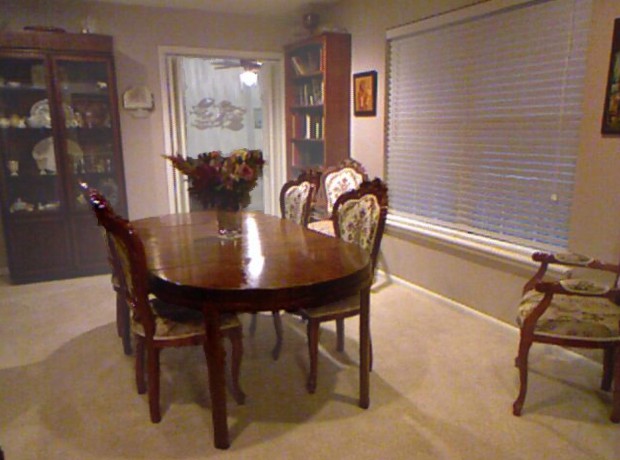} &
			\includegraphics[width=0.095\linewidth]{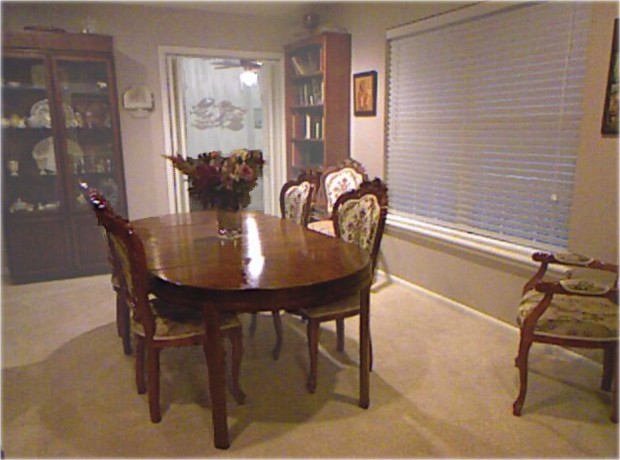} &
			\includegraphics[width=0.095\linewidth]{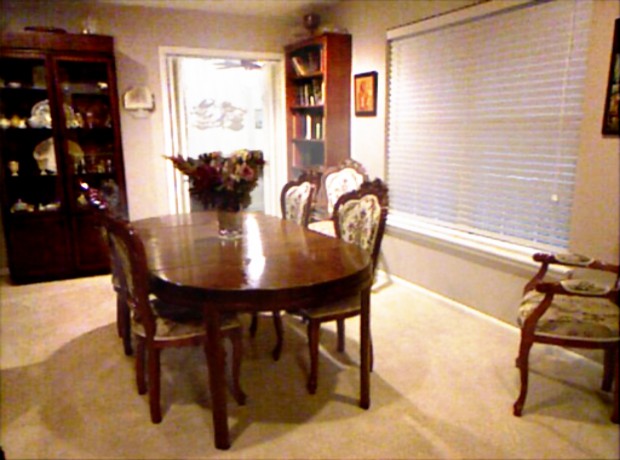} &
			\includegraphics[width=0.095\linewidth]{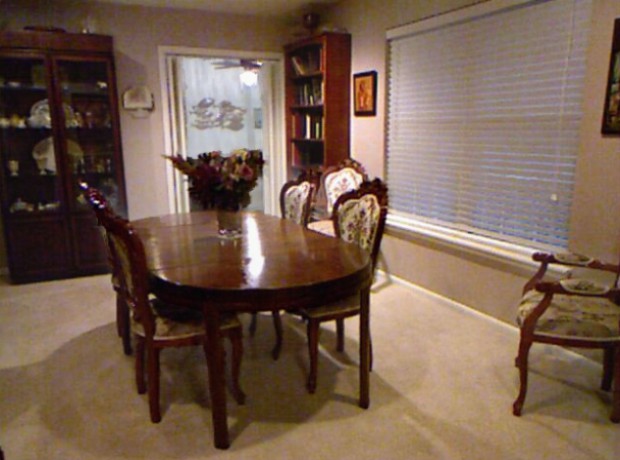} &
			\includegraphics[width=0.095\linewidth]{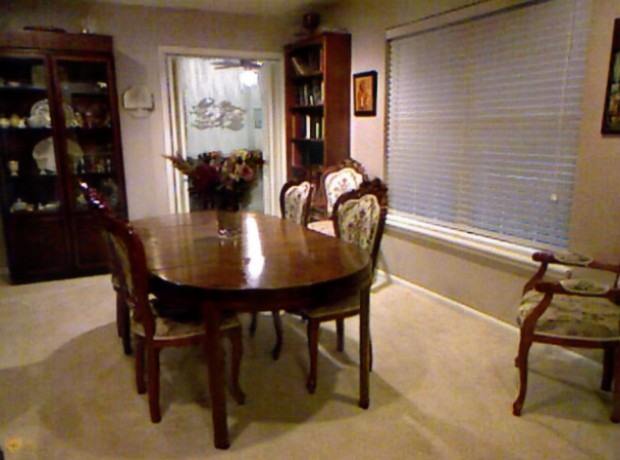} &
			\includegraphics[width=0.095\linewidth]{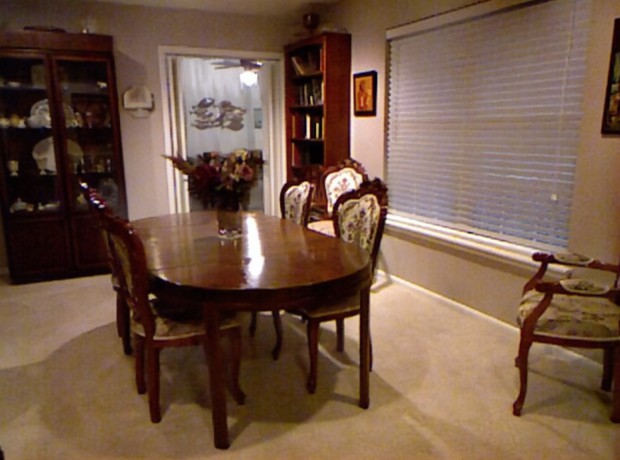} &
			\includegraphics[width=0.095\linewidth]{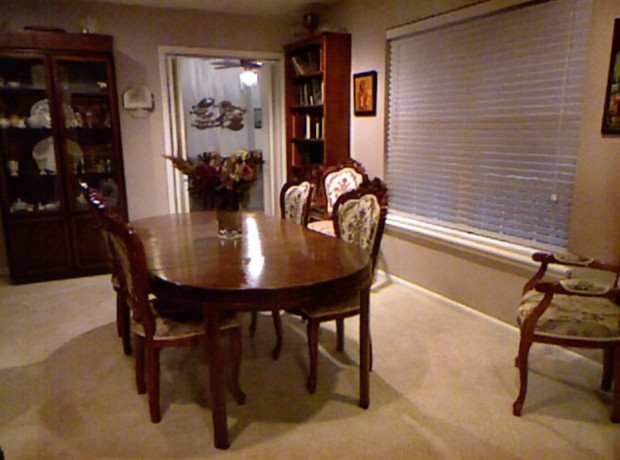} &
			\includegraphics[width=0.095\linewidth]{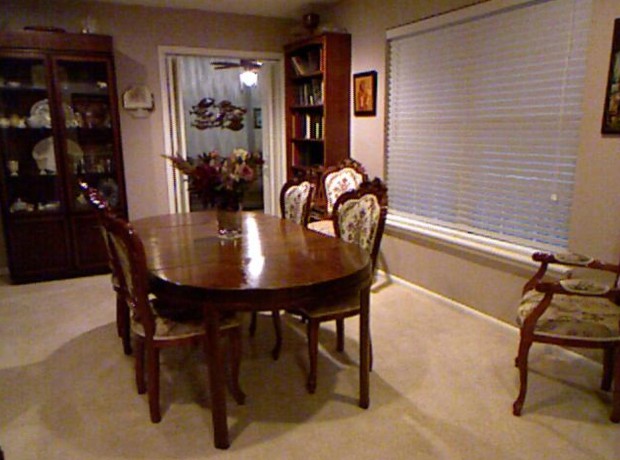} \\

			(a) Hazy image &
			(b) NLD &
			(c) DehazeNet &
			(d) AOD-Net &
			(e) DCPDN &
			(f) EPDN &
			(g) DA\_dehaze &
			(h) T-Net  &
			(i) Stack T-Net &
			(j) GT \\
		\end{tabular}
	\end{center}
	\caption{Visual comparisons on the SOTS dataset. We show the visual results of eight methods on the synthetic data including a kind of prior-based methods and seven kinds of deep-learning-based methods. The images of the first and the last columns represent the hazy images and their corresponding ground truths, respectively. The results of our proposed T-Net and Stack T-Net are separately shown in the eighth and ninth columns.}
	\label{fig: SOTS}
\end{figure*}

\textbf{The second ablation study} is conducted to verify the effectiveness of each module by comparing the performance of several variants of the T-Net. We set the trunk road without the dual attention module (where we use a RDB block to replace the dual attention module) as the basic model. Starting from this basic model, we create other variants by gradually injecting our modifications which include usual skip connections realized by 1x1 convolutions, usual skip connections realized by RDB blocks, our skip connections, as well as our skip connections and dual attention module (T-Net). Table~\ref{tab: tabel2} shows the performance of each variant of T-Net on the SOTS dataset.

\begin{table}[tbp]
\centering
\caption{Comparison on the SOTS dataset of several variants of T-Net.}
\begin{tabular}{Ip{5.1cm}|c|cI}
\Xhline{0.8pt}
\multicolumn{1}{Ic|}{\multirow{1}[4]{*}{Variant}} & \multicolumn{2}{cI}{SOTS } \\
\cline{2-3}         & \multicolumn{1}{c|}{PSNR} & \multicolumn{1}{cI}{SSIM} \\
\Xhline{0.8pt}
the basic model                          &24.19 &0.8228\\ 
\hline
with usual skip connections (1x1 conv)   &27.42 &0.9504\\ 
\hline
with usual skip connections (RDB block)  &27.61 &0.9520\\ 
\hline
with our skip connections                &27.82 &0.9530\\ 
\hline
T-Net                                    &\textbf{28.55} &\textbf{0.9543}\\
\Xhline{0.8pt}
\end{tabular}%
\label{tab: tabel2}
\end{table}%

It is clear that the performance is further improved every time a new component is added to the basic model, which justifies the overall design. As shown in Table~\ref{tab: tabel2}, the comparison among these four variants show the effectiveness of our skip connections with a new fusion strategy. It is demonstrated that RDB blocks used in the lateral connection instead of 1x1 convolution can extract more complex semantic information for dehazing, and that different scale features have different importances in feature fusion. Furthermore, the addition of the dual attention module has the greatest effect on the dehazing performance improvement, which effectively enhances the robustness and generalization ability of the network. This ablation study demonstrates the contribution of each component in our T-Net.

\textbf{The third ablation study} is conducted to evaluate our proposed Stack T-Net with different recursive stage number $K$ and demonstrates the effectiveness of our recursive strategy. Limited by physical memory, we just set $K = 1, 2, 3$. Table~\ref{tab: tabel3} shows the performance on the SOTS dataset of the Stack T-Net with different recursive stage numbers.

\begin{table}[tbp]
\centering
\caption{Comparison on the SOTS dataset of the Stack T-Net with different $K$ stages.}
\begin{tabular}{Ic|c|cI}
\Xhline{0.8pt}
\multirow{1}[4]{*}{Recursive stage number} & \multicolumn{2}{cI}{SOTS}\\
\cline{2-3}          & \multicolumn{1}{C{1.5cm}|}{PSNR} & \multicolumn{1}{C{1.5cm}I}{SSIM}\\
\Xhline{0.8pt}
$K=1$   &28.55 &0.9543 \\ 
\hline
$K=2$   &28.71 &0.9556 \\ 
\hline
$K=3$   &\textbf{28.83} &\textbf{0.9551} \\ 
\Xhline{0.8pt}
\end{tabular}%
\label{tab: tabel3}
\end{table}%

\begin{figure*}[tbp]
	\scriptsize
	\centering
	\renewcommand{\tabcolsep}{1pt} 
	\renewcommand{\arraystretch}{1} 
	\begin{center}
		\begin{tabular}{ccccccccc}
		
			\includegraphics[width=0.107\linewidth]{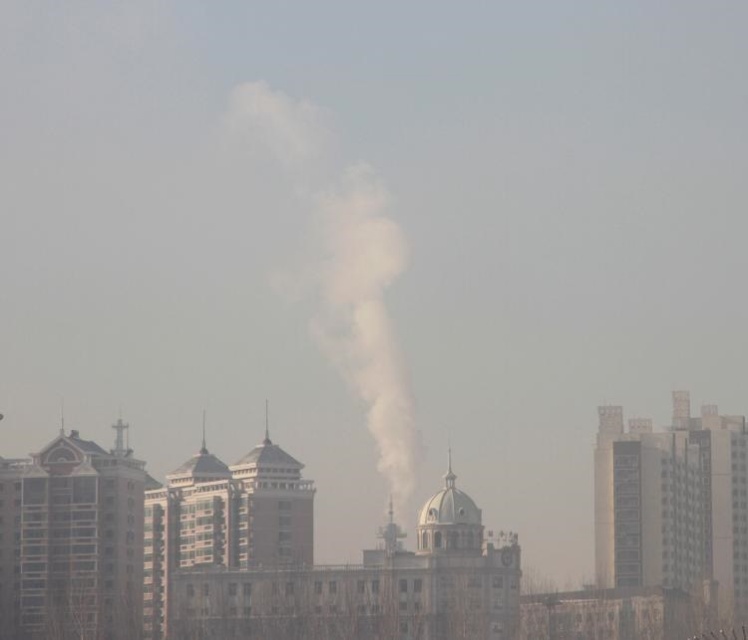} &
			\includegraphics[width=0.107\linewidth]{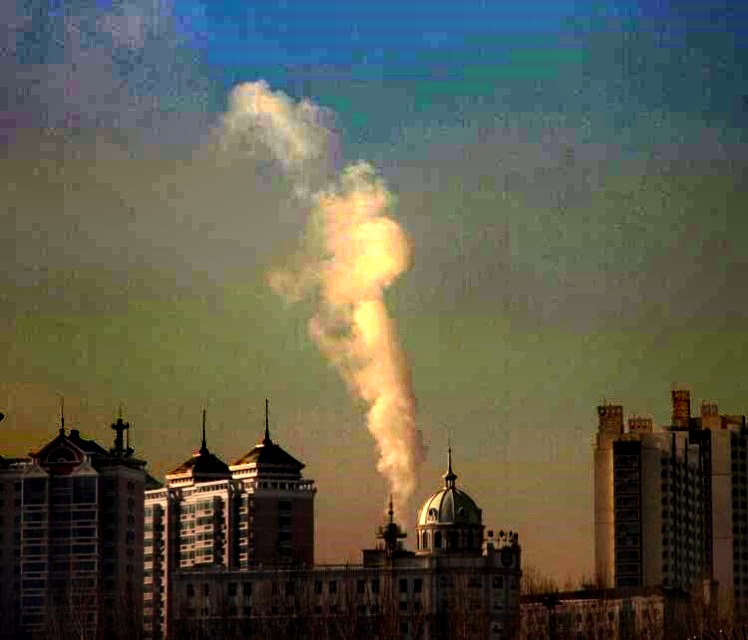} &
			\includegraphics[width=0.107\linewidth]{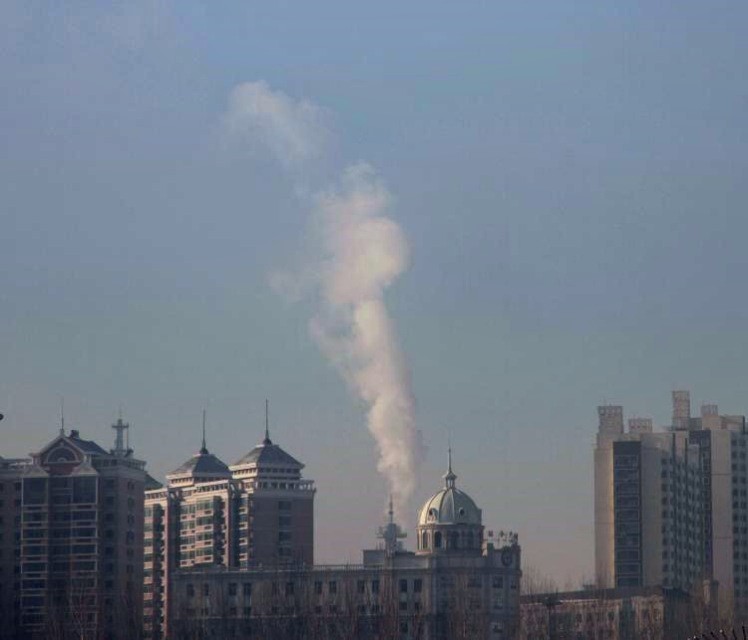} &
			\includegraphics[width=0.107\linewidth]{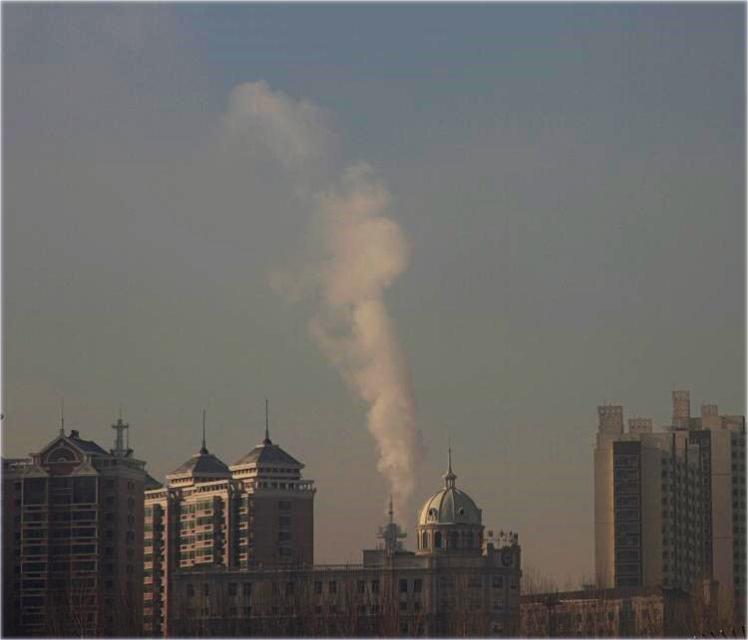} &
			\includegraphics[width=0.107\linewidth]{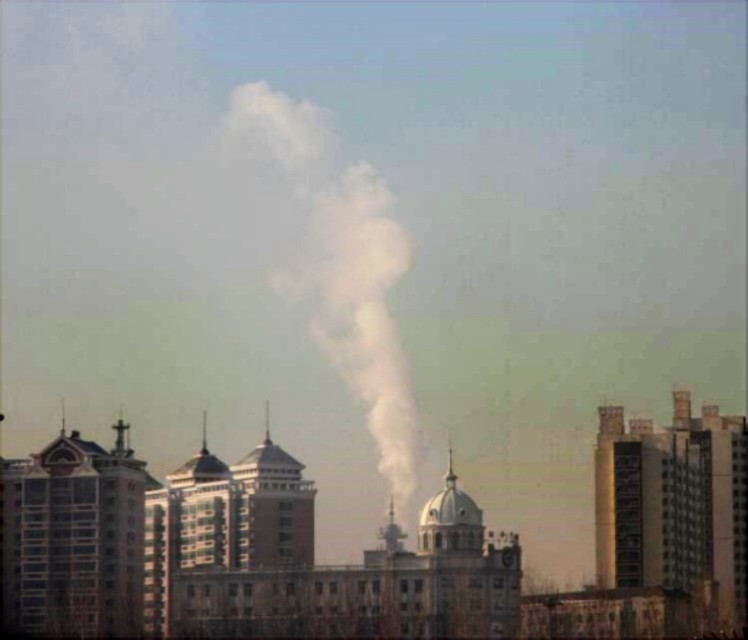} &
			\includegraphics[width=0.107\linewidth]{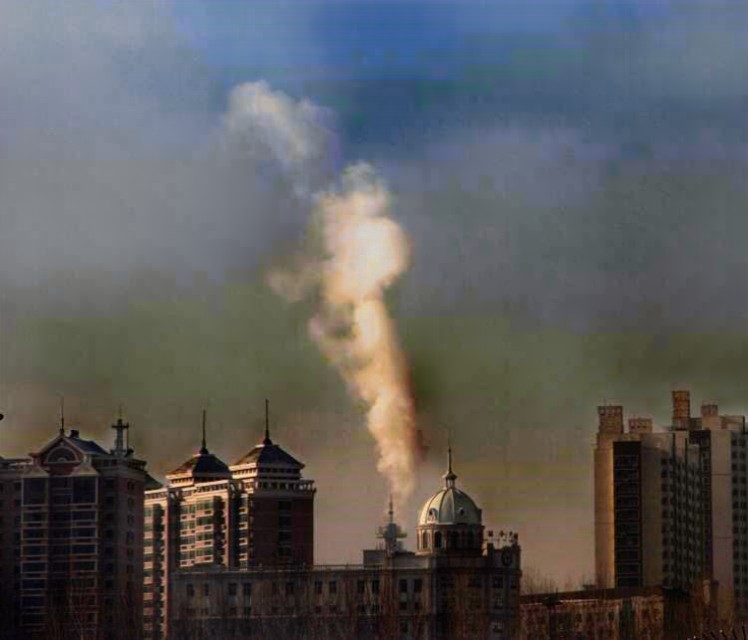} &
			\includegraphics[width=0.107\linewidth]{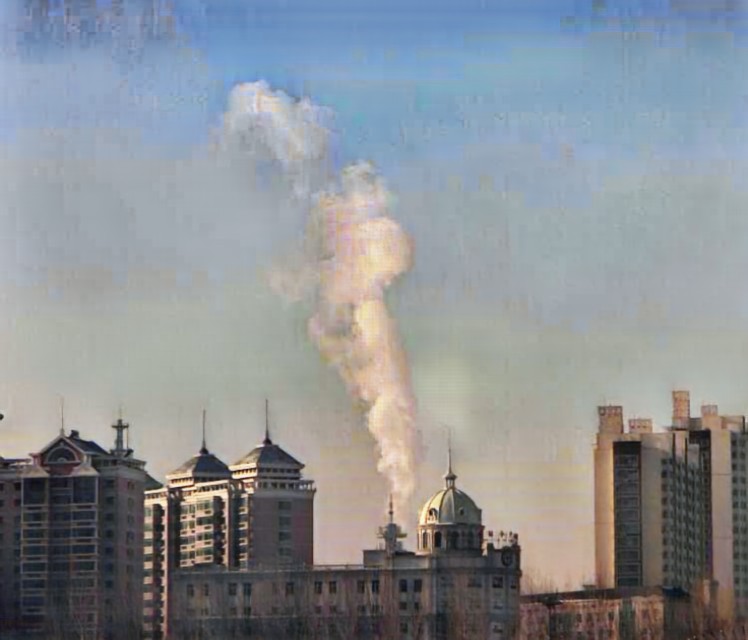} &
			\includegraphics[width=0.107\linewidth]{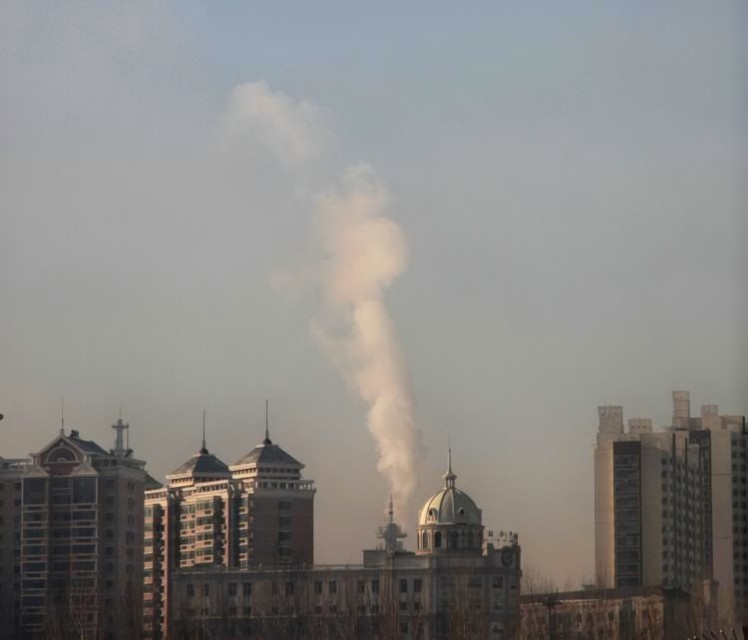} &
			\includegraphics[width=0.107\linewidth]{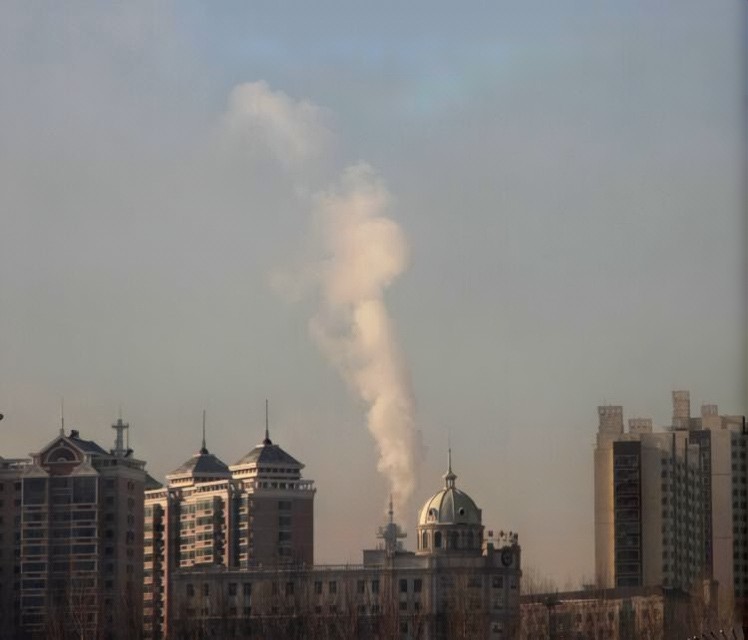} \\		
			
			\includegraphics[width=0.107\linewidth]{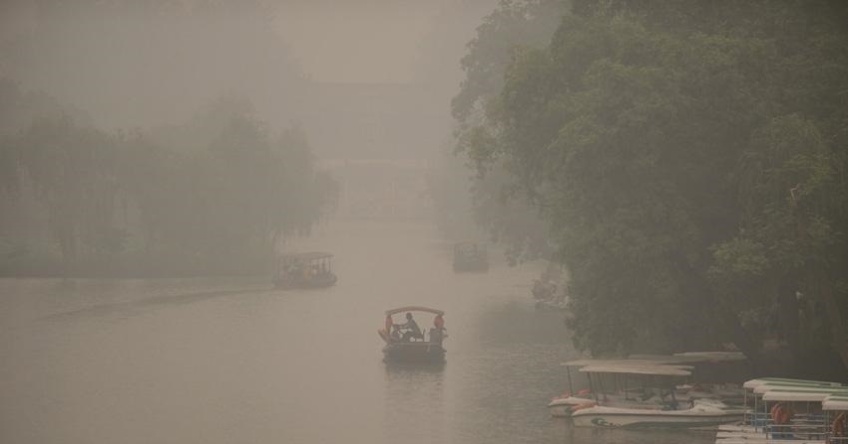} &
			\includegraphics[width=0.107\linewidth]{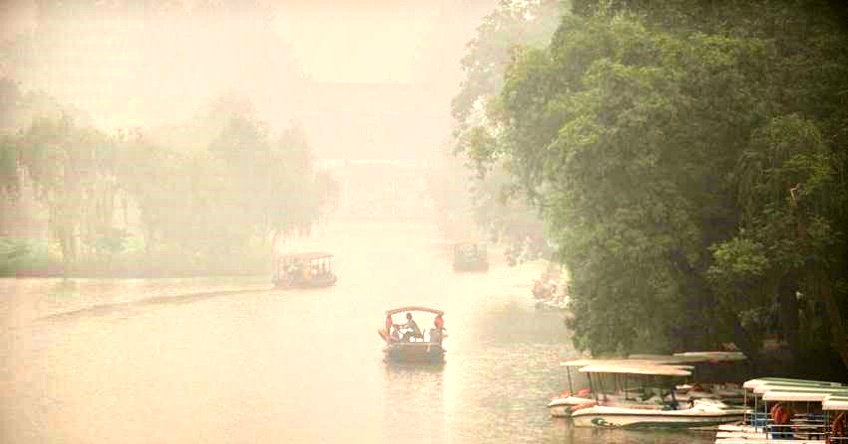} &
			\includegraphics[width=0.107\linewidth]{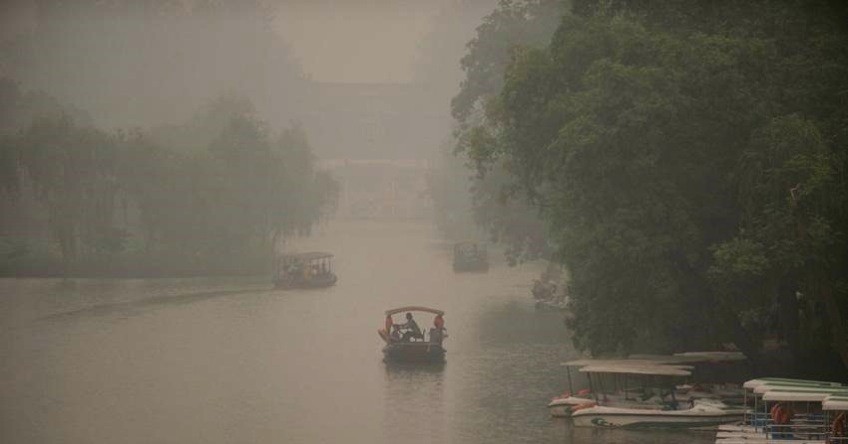} &
			\includegraphics[width=0.107\linewidth]{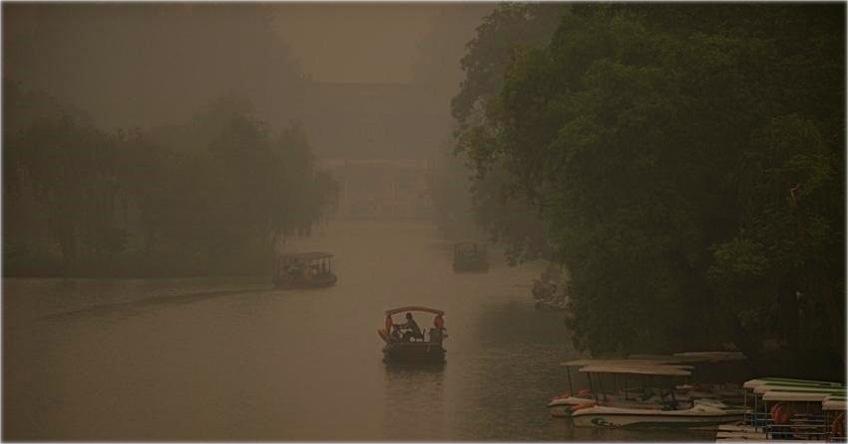} &
			\includegraphics[width=0.107\linewidth]{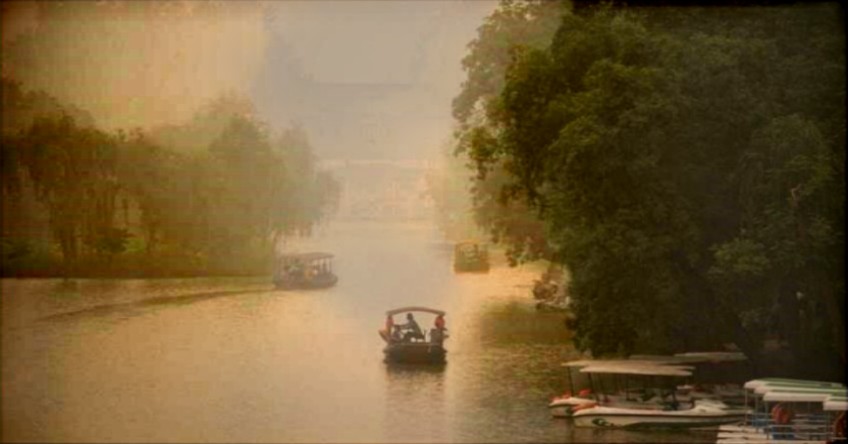} &
			\includegraphics[width=0.107\linewidth]{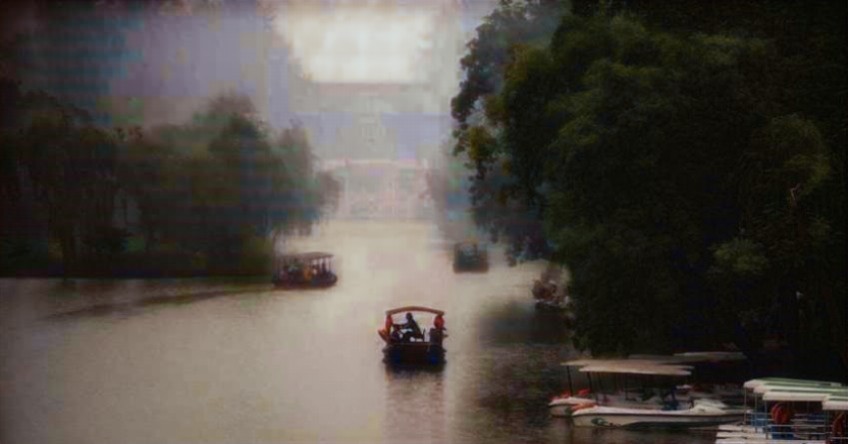} &
			\includegraphics[width=0.107\linewidth]{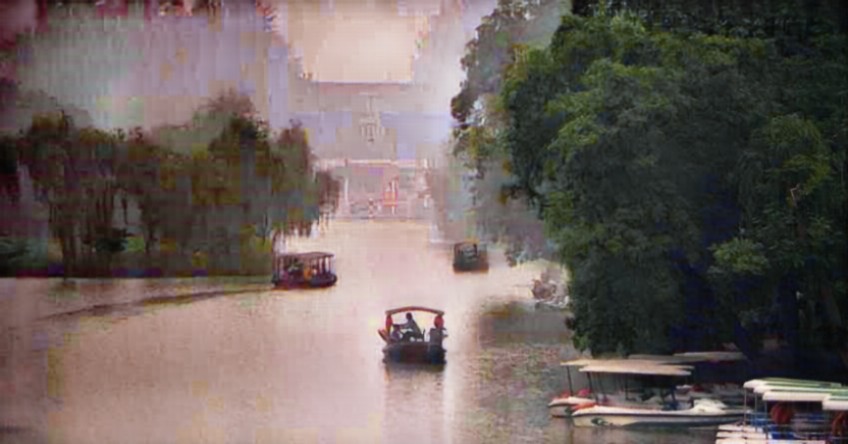} &
			\includegraphics[width=0.107\linewidth]{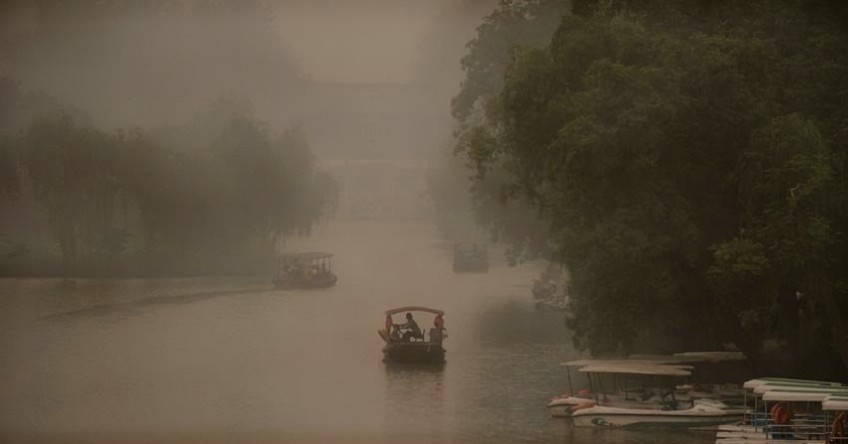} &
			\includegraphics[width=0.107\linewidth]{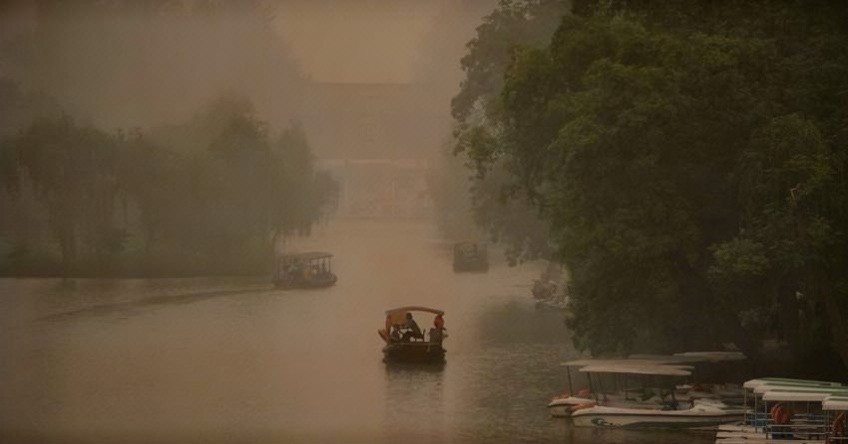} \\
			
			\includegraphics[width=0.107\linewidth]{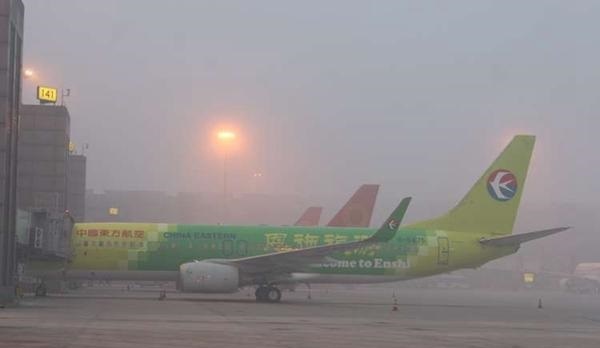} &
			\includegraphics[width=0.107\linewidth]{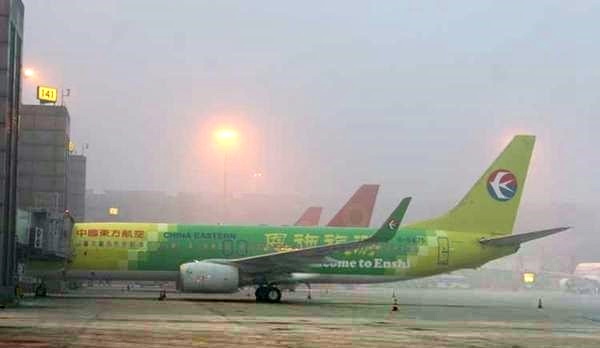} &
			\includegraphics[width=0.107\linewidth]{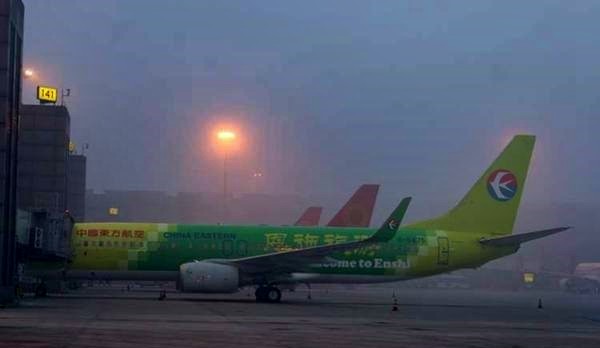} &
		    \includegraphics[width=0.107\linewidth]{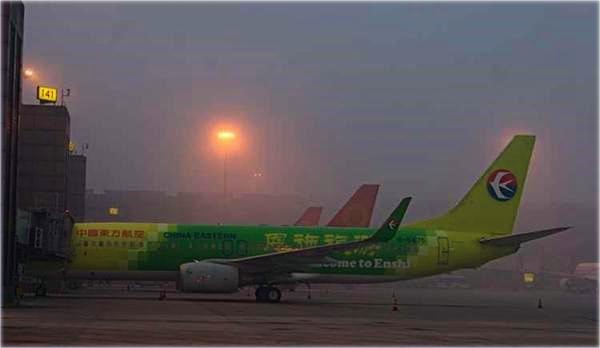} &
			\includegraphics[width=0.107\linewidth]{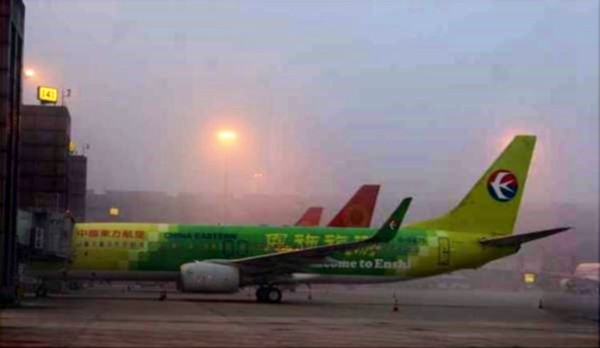} &
			\includegraphics[width=0.107\linewidth]{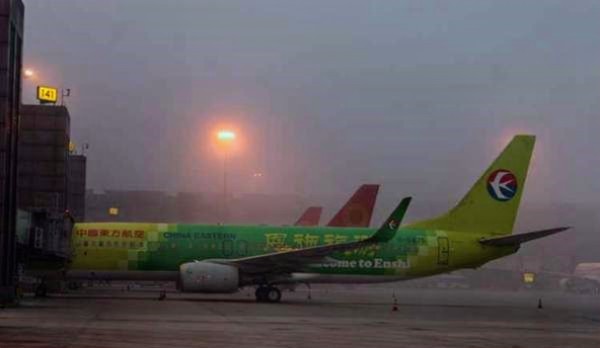} &
			\includegraphics[width=0.107\linewidth]{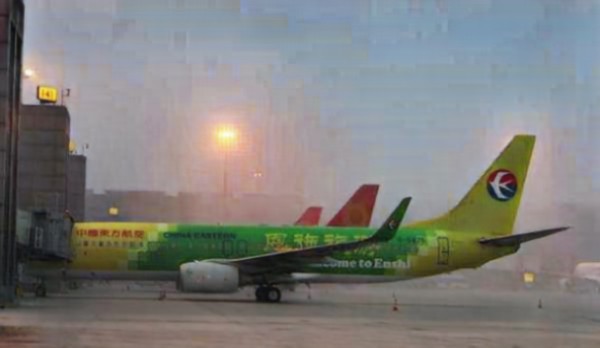} &
			\includegraphics[width=0.107\linewidth]{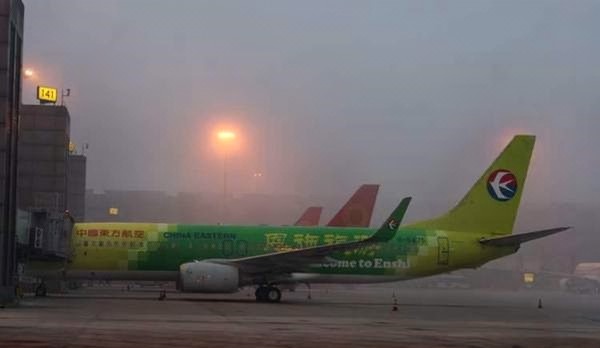} &
			\includegraphics[width=0.107\linewidth]{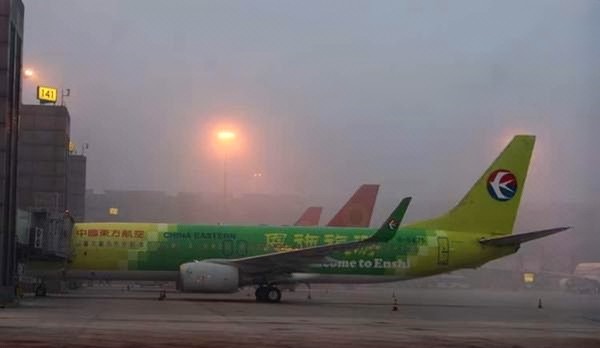} \\
			
			\includegraphics[width=0.107\linewidth]{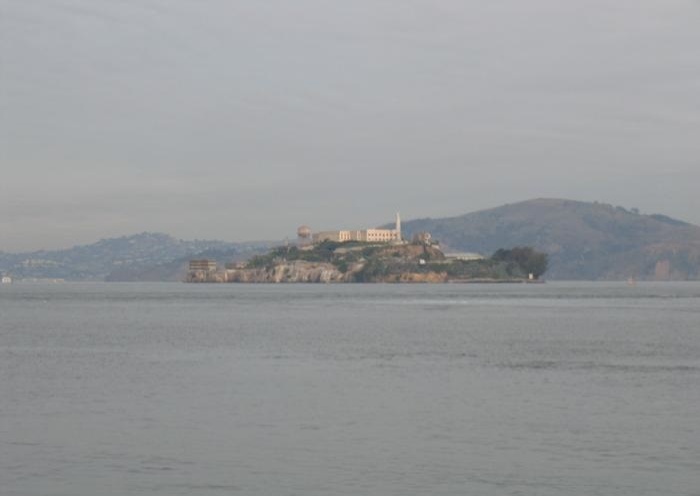} &
			\includegraphics[width=0.107\linewidth]{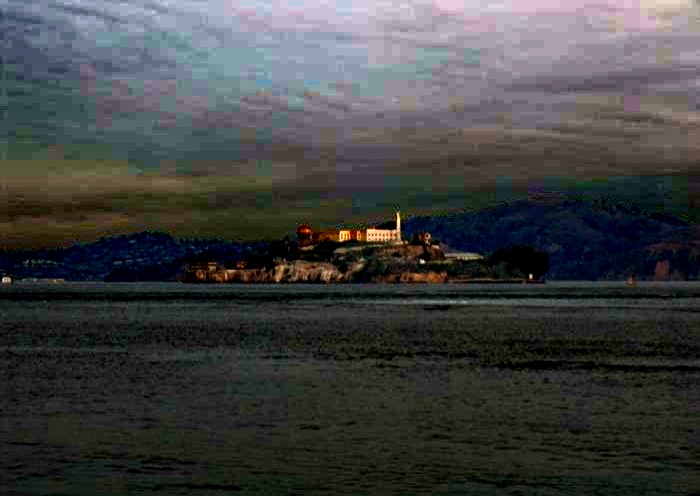} &
			\includegraphics[width=0.107\linewidth]{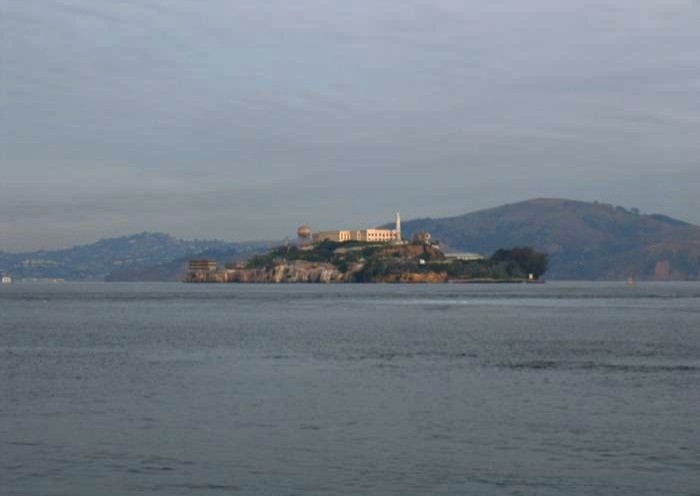} &
			\includegraphics[width=0.107\linewidth]{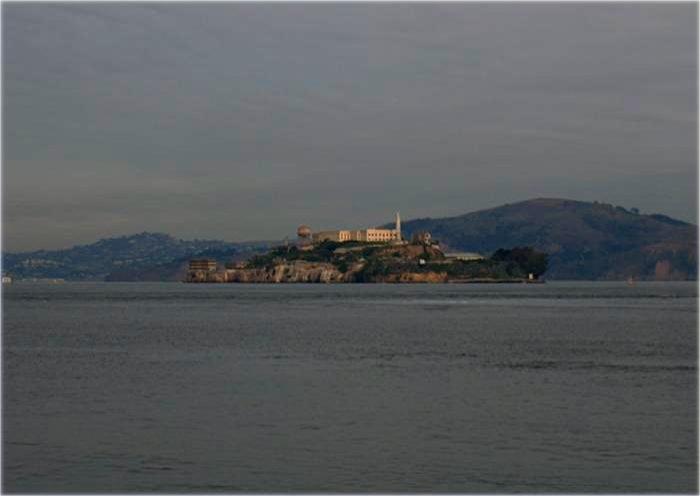} &
			\includegraphics[width=0.107\linewidth]{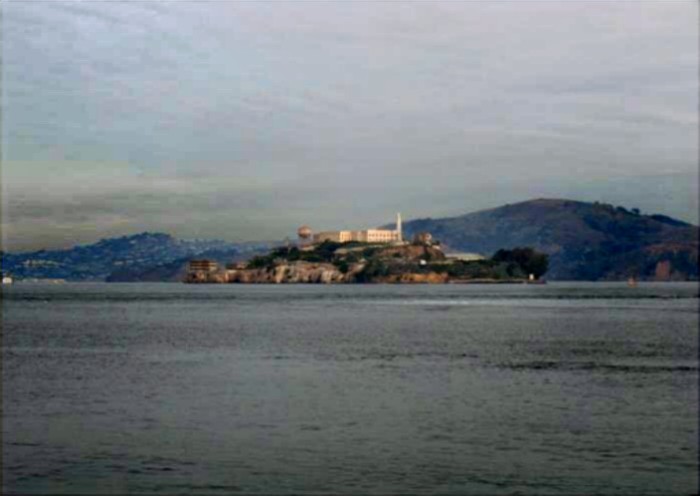} &
			\includegraphics[width=0.107\linewidth]{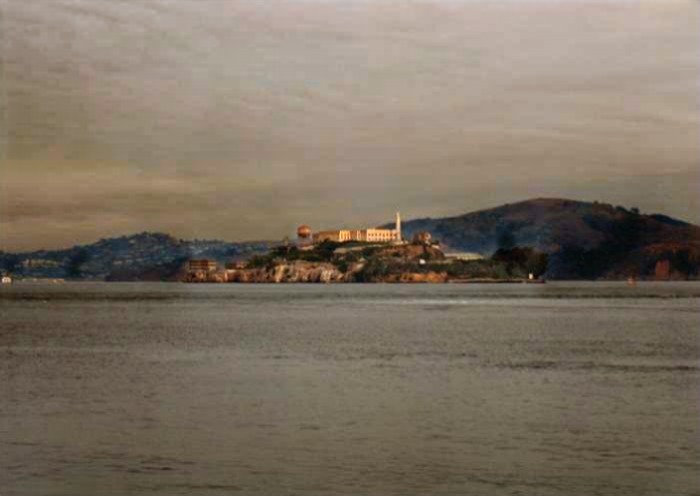} &
			\includegraphics[width=0.107\linewidth]{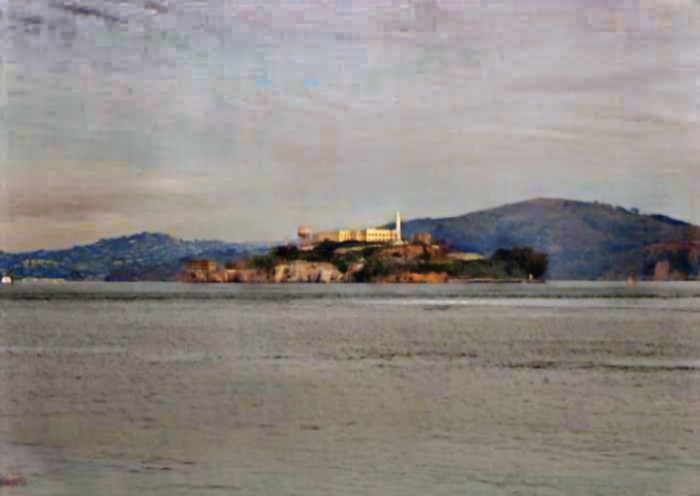} &
			\includegraphics[width=0.107\linewidth]{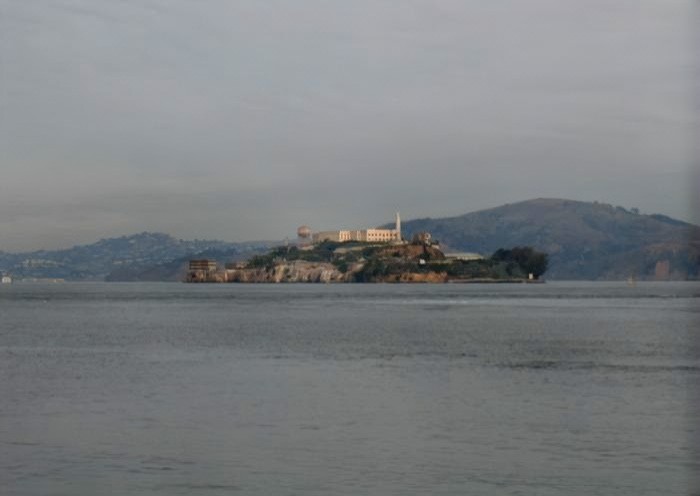} &
			\includegraphics[width=0.107\linewidth]{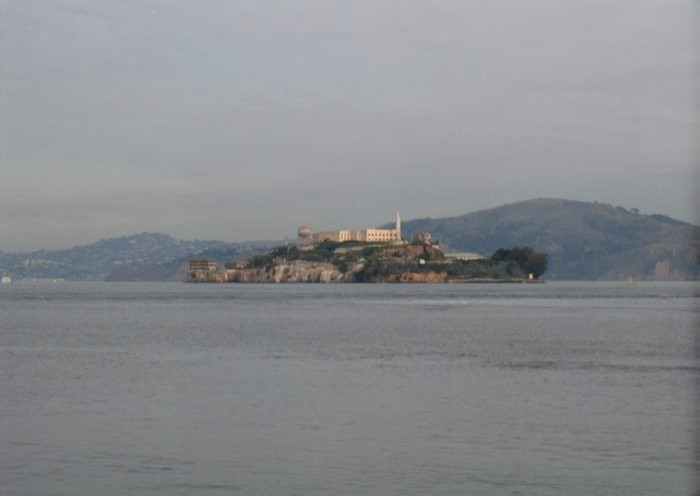} \\
			
			\includegraphics[width=0.107\linewidth]{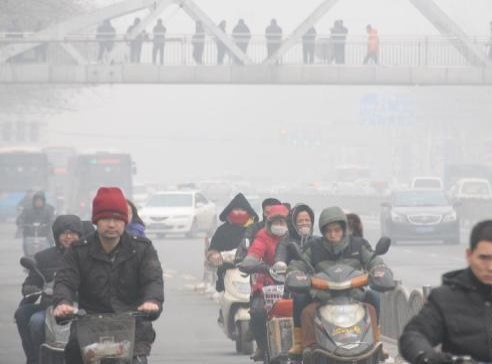} &
			\includegraphics[width=0.107\linewidth]{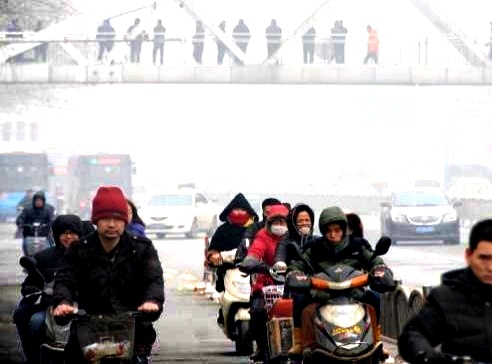} &
			\includegraphics[width=0.107\linewidth]{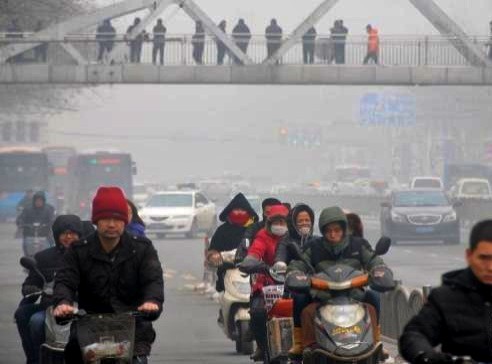} &
			\includegraphics[width=0.107\linewidth]{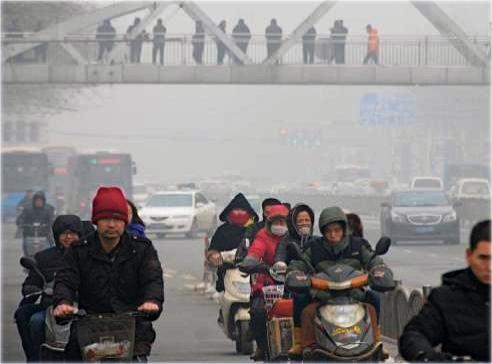} &
			\includegraphics[width=0.107\linewidth]{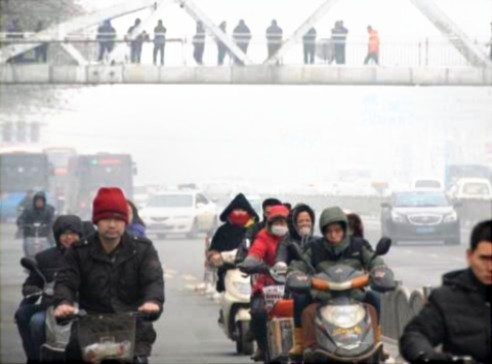} &
			\includegraphics[width=0.107\linewidth]{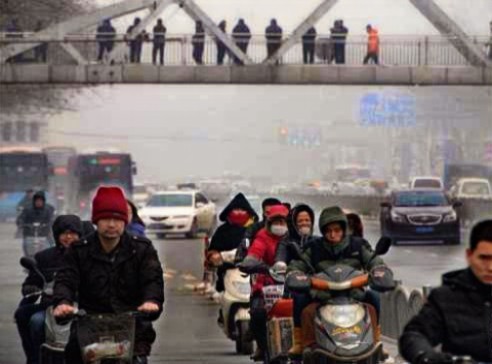} &
			\includegraphics[width=0.107\linewidth]{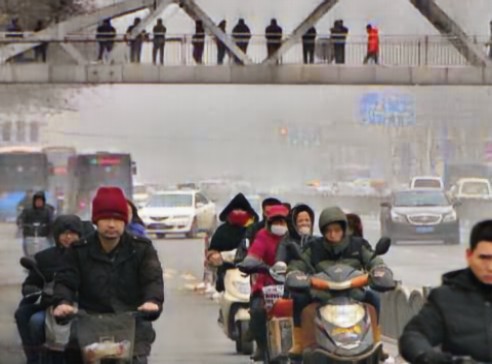} &
			\includegraphics[width=0.107\linewidth]{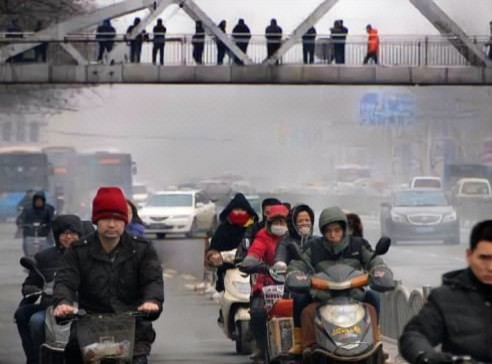} &
			\includegraphics[width=0.107\linewidth]{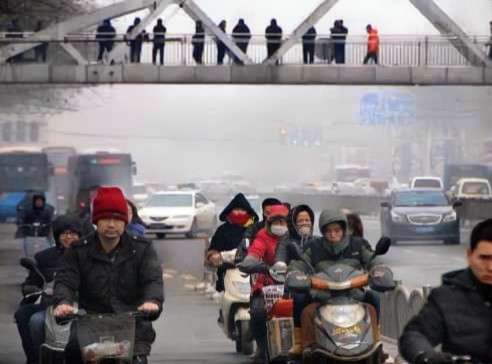} \\
			
			\includegraphics[width=0.107\linewidth]{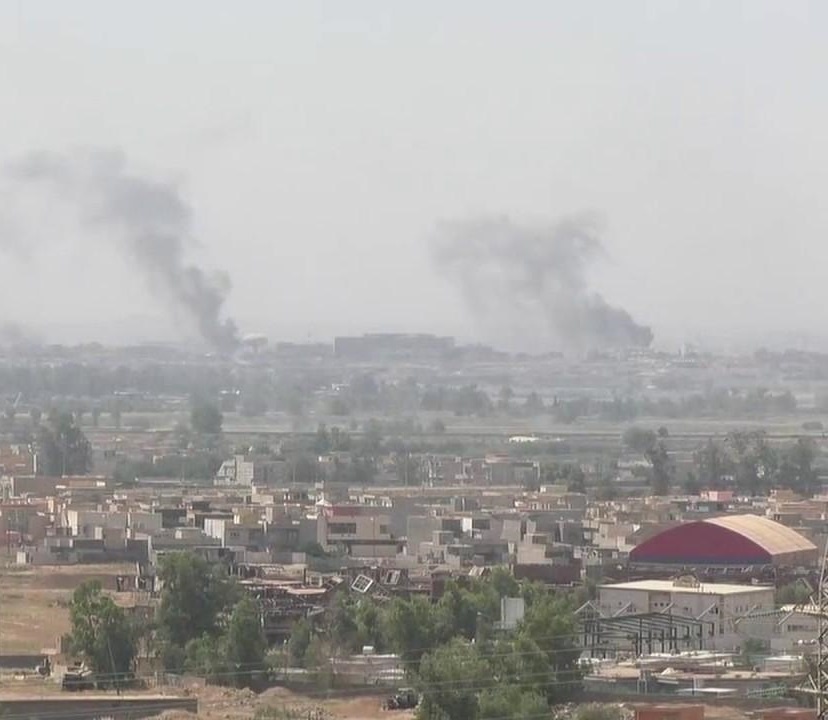} &
			\includegraphics[width=0.107\linewidth]{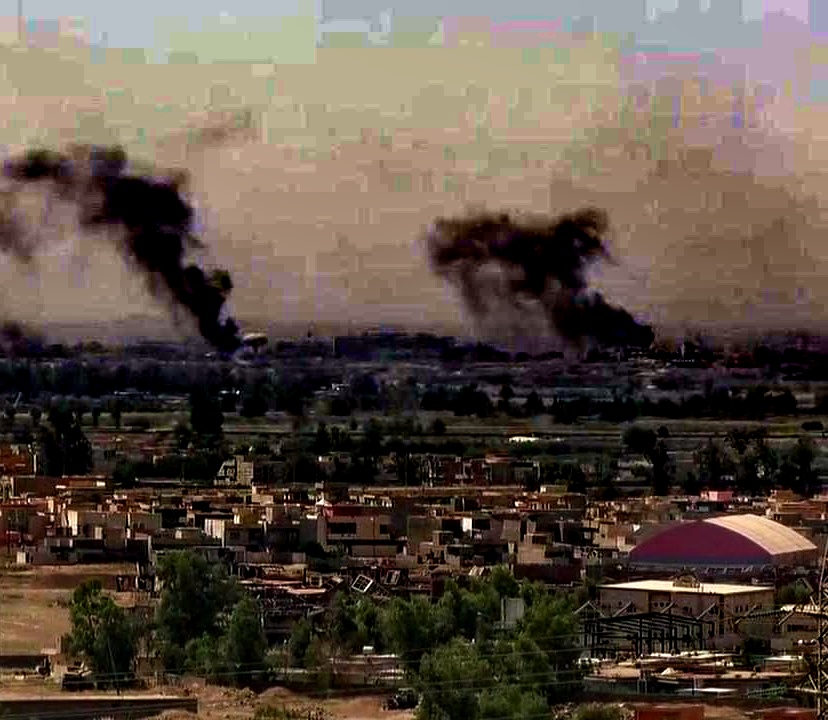} &
			\includegraphics[width=0.107\linewidth]{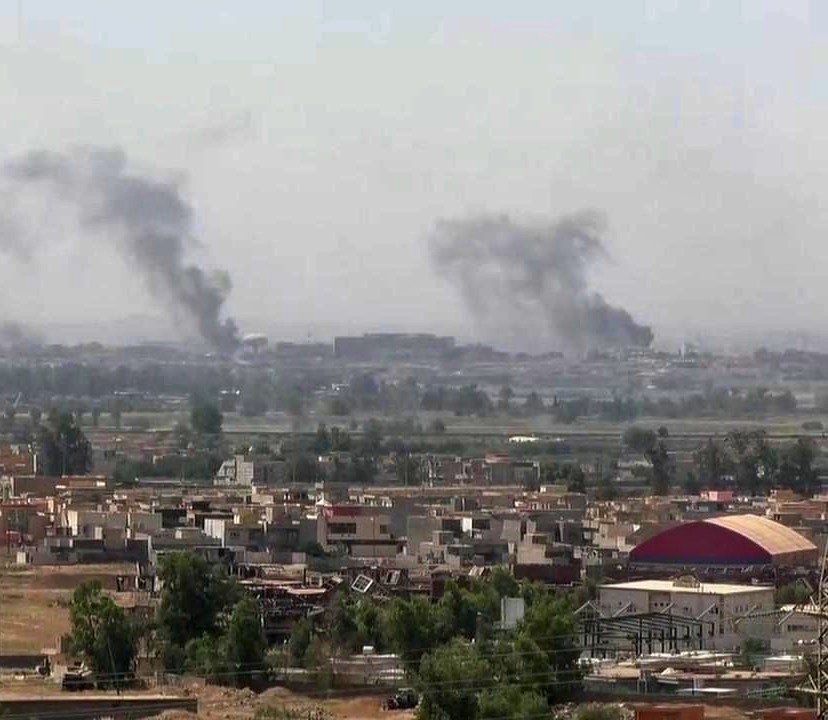} &
			\includegraphics[width=0.107\linewidth]{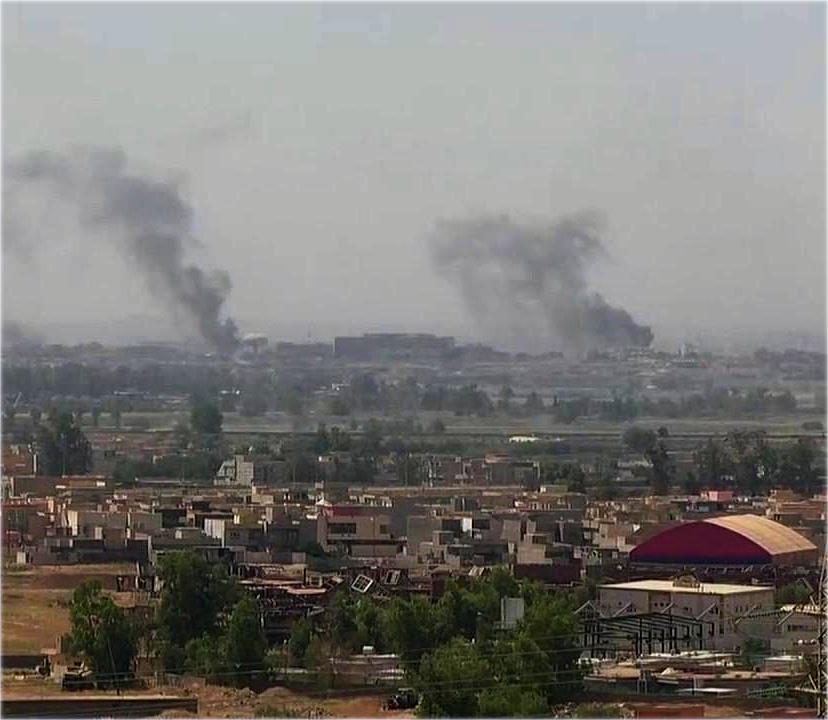} &
			\includegraphics[width=0.107\linewidth]{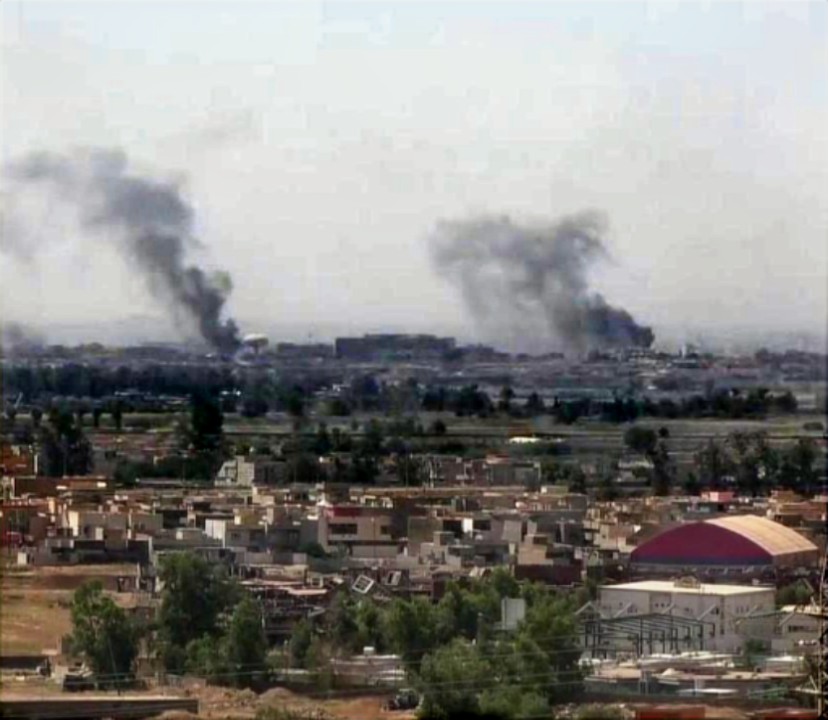} &
			\includegraphics[width=0.107\linewidth]{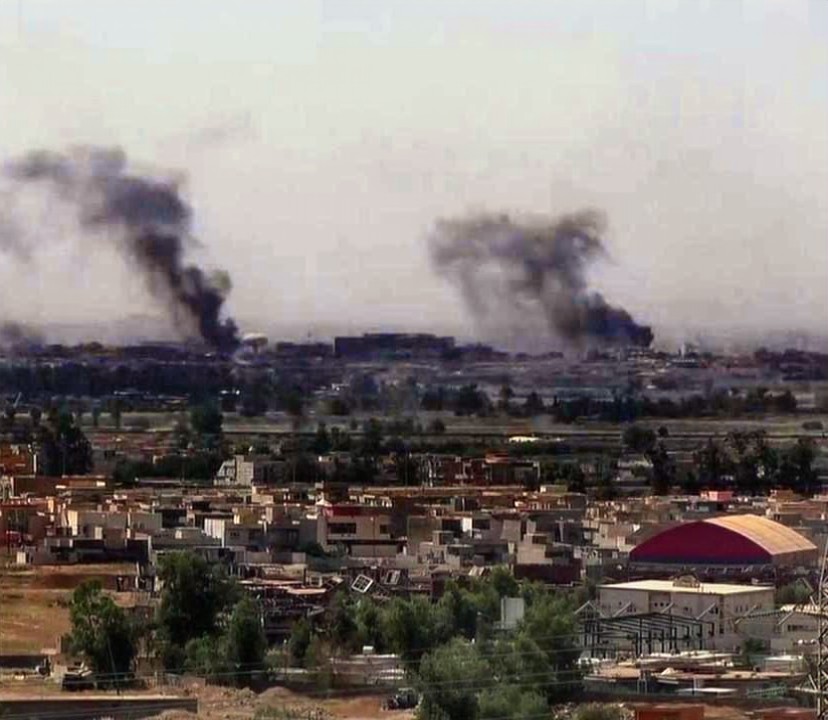} &
			\includegraphics[width=0.107\linewidth]{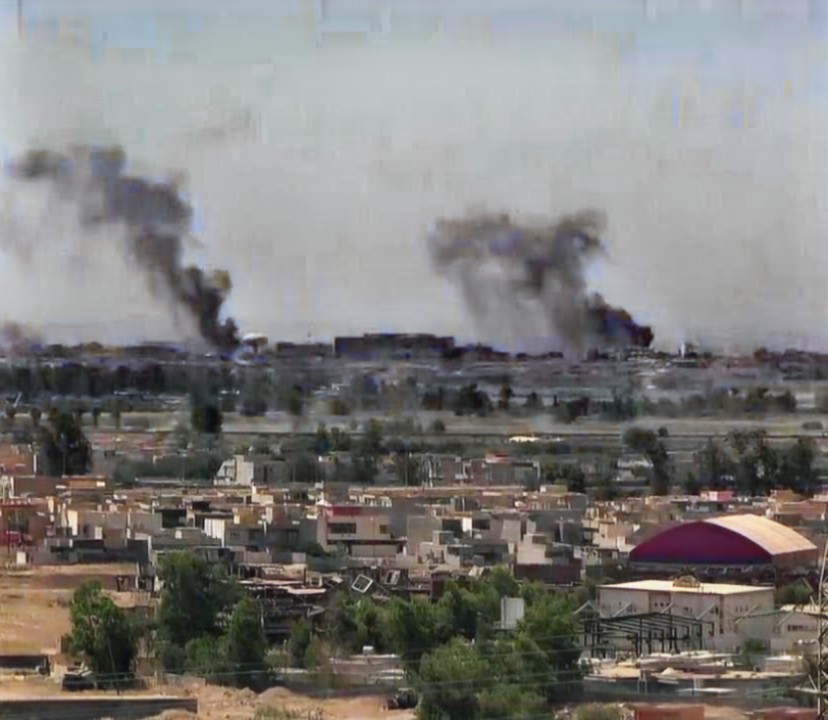} &
			\includegraphics[width=0.107\linewidth]{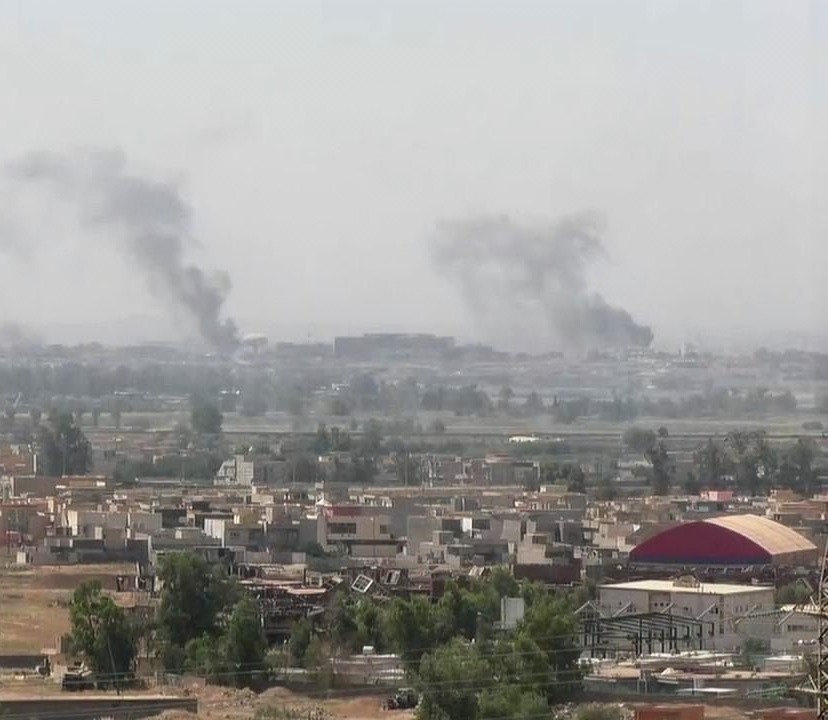} &
			\includegraphics[width=0.107\linewidth]{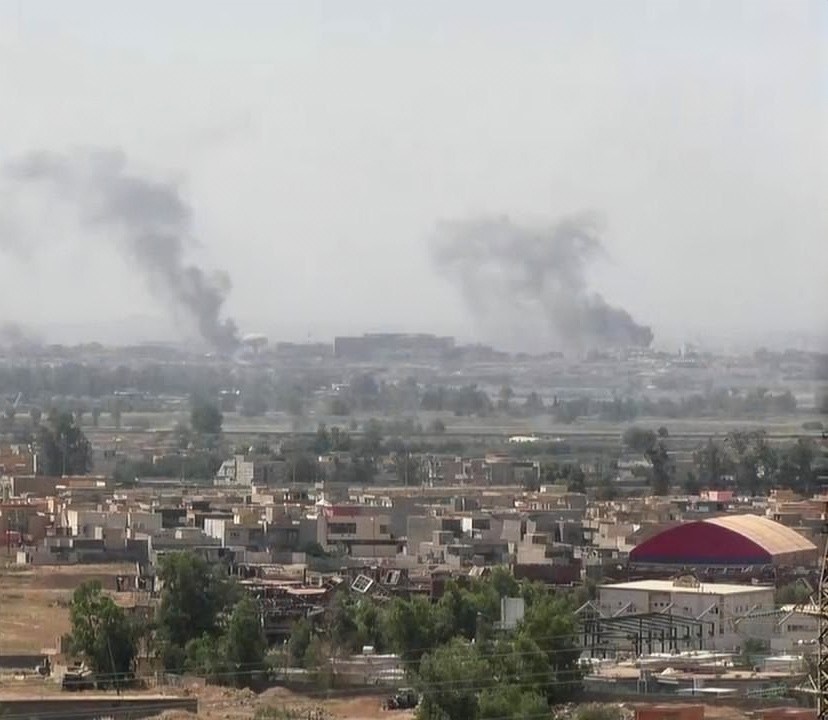} \\	
			
			\includegraphics[width=0.107\linewidth]{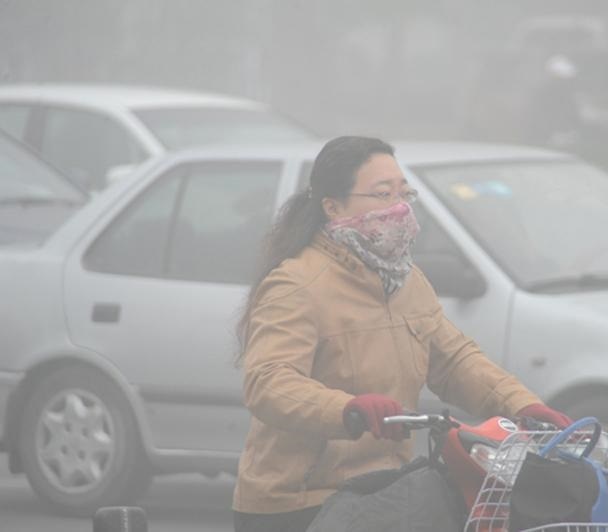} &
			\includegraphics[width=0.107\linewidth]{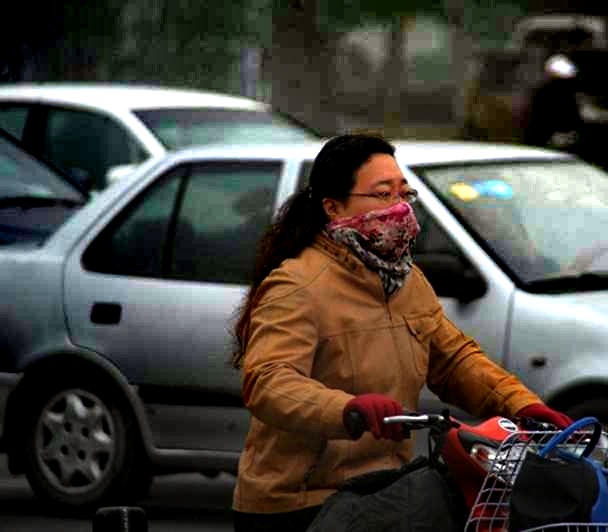} &
			\includegraphics[width=0.107\linewidth]{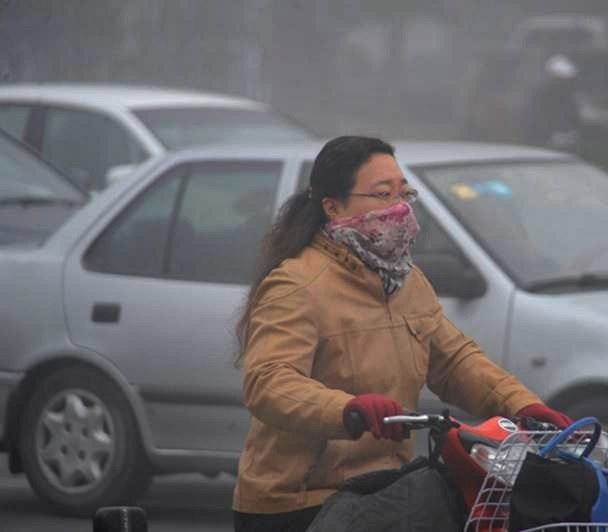} &
			\includegraphics[width=0.107\linewidth]{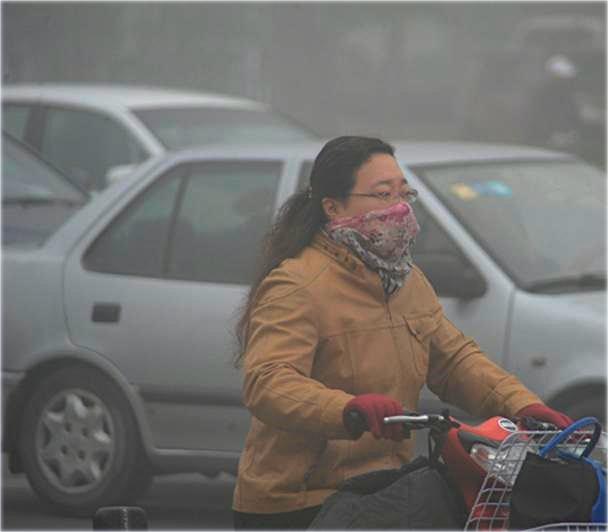} &
			\includegraphics[width=0.107\linewidth]{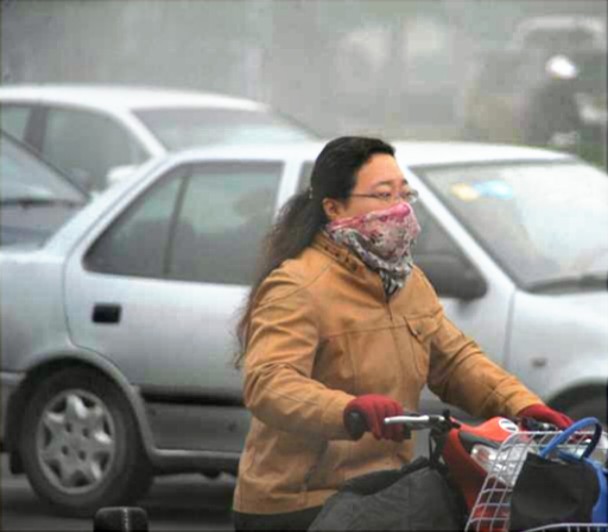} &
			\includegraphics[width=0.107\linewidth]{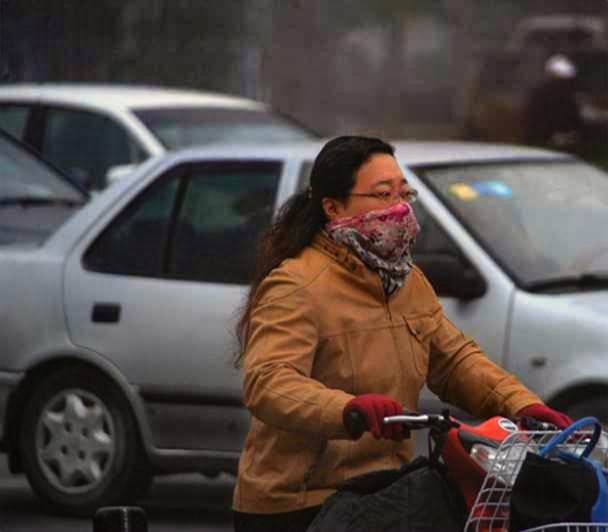} &
			\includegraphics[width=0.107\linewidth]{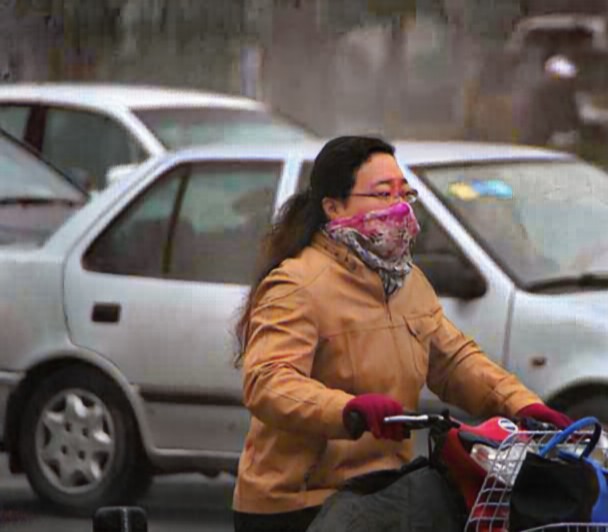} &
			\includegraphics[width=0.107\linewidth]{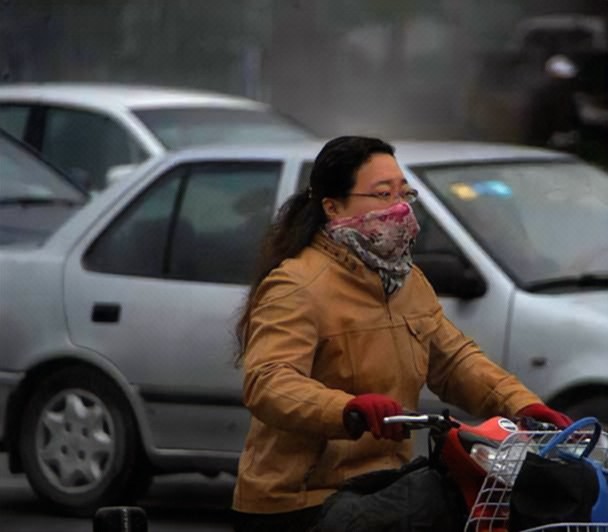} &
			\includegraphics[width=0.107\linewidth]{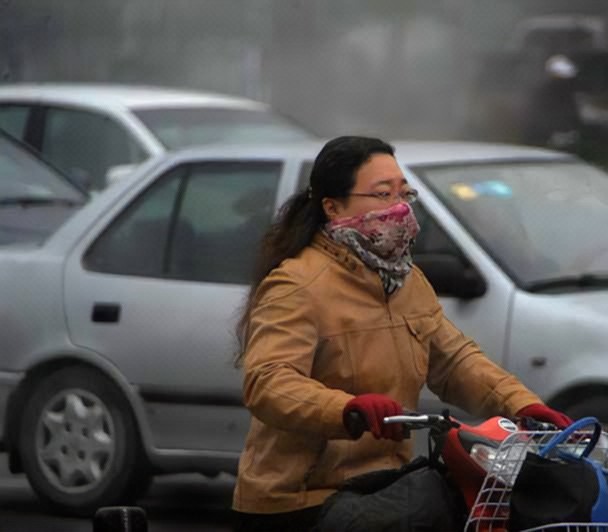} \\

			(a) Hazy image &
			(b) NLD &
			(c) DehazeNet &
			(d) AOD-Net &
			(e) DCPDN &
			(f) EPDN &
			(g) DA\_dehaze &
			(h) T-Net  &
			(i) Stack T-Net \\
		\end{tabular}
	\end{center}
	\caption{Visual comparisons on the URHI dataset. We show the visual results of different methods on the real-world data. The first column shows the hazy images, and other columns represent the dehazed results of different methods. The results of our proposed T-Net and Stack T-Net are separately shown in the last two columns.}
	\label{fig: URHI}
\end{figure*}

As shown in Table~\ref{tab: tabel3}, Stack T-Net usually achieves higher average PSNR and SSIM values as the recursive stage number increases. However, the SSIM value of Stack T-Net with three stages is a bit lower than the value of Stack T-Net with two stages, which is because the best model is saved according to the highest PSNR value, instead of the SSIM value. Actually, the highest SSIM value of Stack T-Net with three stages is higher than that of Stack T-Net with two stages in the experiment. This ablation study demonstrates that our recursive strategy is effective.

\subsection{Performance comparison on synthetic dataset}
The proposed method is tested on the same synthetic dataset SOTS, to qualitatively and quantitatively compare with the state-of-the-art methods which include DCP~\cite{5567108}, NLD~\cite{Berman_2016_CVPR}, MSCNN~\cite{2016Single}, DehazeNet~\cite{2016DehazeNet},  AOD-Net~\cite{8237773}, GFN~\cite{Ren_2018_CVPR}, DCPDN~\cite{8578435}, EPDN~\cite{8953692}, and DA\_dehaze~\cite{Shao_2020_CVPR}. Apart from the DCP and NLD which are prior based methods, the others are deep learning based methods. We test two models of our algorithm, namely T-Net and Stack T-Net with three stages in the comparison experiment. 

Table~\ref{tab: tabel4} presents the quantitative comparison results of different dehazed methods on the SOTS dataset in terms of the average PSNR and SSIM values. Moreover, the data in this table except the last two rows is all quoted from the work of Shao~\cite{Shao_2020_CVPR}. It is clear that our work outperforms the state-of-the-art methods by a wide margin. Not only Stack T-Net but also our plain T-Net outperform the state-of-the-art DA\_haze~\cite{Shao_2020_CVPR} in terms of both the average PSNR and SSIM values on the SOTS dataset.

\begin{table}[tbp]
\centering
\caption{Quantitative comparison on the SOTS dataset between the state-of-the-art dehazing methods.}
\begin{tabular}{IC{3.0cm}|C{1.5cm}|C{1.5cm}I}
\Xhline{0.8pt}
\multirow{1}[4]{*}{Methods} & \multicolumn{2}{cI}{SOTS} \\
\cline{2-3}          & PSNR  & SSIM\\
\Xhline{0.8pt}
DCP~\cite{5567108}   & 15.49  & 0.64 \\
\hline
NLD~\cite{Berman_2016_CVPR}  & 17.27 & 0.75 \\
\hline
MSCNN~\cite{2016Single} & 17.57 & 0.81 \\
\hline
DehazeNet~\cite{2016DehazeNet} & 21.14 & 0.85 \\
\hline
AOD-Net~\cite{8237773} & 19.06 & 0.85 \\
\hline
DCPDN~\cite{8578435} & 19.39 & 0.65 \\
\hline
GFN~\cite{Ren_2018_CVPR}  & 22.30 & 0.88  \\
\hline
EPDN~\cite{8953692} & 23.82 & 0.89 \\
\hline
DA\_dehaze~\cite{Shao_2020_CVPR} & 27.76 & 0.93 \\
\hline
T-Net & 28.55 & 0.95  \\
\hline
Stack T-Net &\textbf{28.83} &\textbf{0.96}  \\
\Xhline{0.8pt}
\end{tabular}%
\label{tab: tabel4}
\end{table}%

Fig.~\ref{fig: SOTS} shows the qualitative comparison results among different dehazing methods, where the synthetic images are chosen from the SOTS datasets. The haze density and scene of the shown images in Fig.~\ref{fig: SOTS} are different, where we choose four images from the indoor and outdoor subset of SOTS, respectively. We note that the first three methods~\cite{Berman_2016_CVPR,2016DehazeNet,8237773} perform poorly on most images and are easy to cause under-dehazing, color oversaturation and other distortion problems. The dehazing results of DCPDN~\cite{8578435} and EPDN~\cite{8953692} seem better, but DCPDN~\cite{8578435} tends to make the brighter areas in images overexposed and EPDN~\cite{8953692} tends to reduce the brightness of the darker areas in images, which both cause the loss of many details in the images (see e.g., the first and third rows in Fig.~\ref{fig: SOTS} (e) (f)). DA\_dehaze~\cite{Shao_2020_CVPR} performs well on most images, but it tends to cause color distortions (see e.g., the first, second, fifth and sixth rows in Fig.~\ref{fig: SOTS} (g) of which the color is a little different from the ground truth images) and cause halo artifacts sometimes (see e.g., the sky area of the third row in Fig.~\ref{fig: SOTS} (g)), where other methods except our methods have the same problem. 

Compared with the state-of-the-art methods, our T-Net and Stack T-Net have the best performance in terms of haze removal and are effective in suppressing halo artifacts and color distortions. The dehazed images are visually most similar to their ground truth ones (see e.g., Fig.~\ref{fig: SOTS} (h) (i)). In addition, we note that Stack T-Net can further eliminate hazy areas on the base of T-Net (see e.g. the first row in Fig.~\ref{fig: SOTS} (h) in which the haze in the middle is eliminated in the same row of (i)), demonstrating that the recursive strategy is helpful for image dehazing.

\subsection{Performance comparison on real-world hazy images}
We further compare our method with the state-of-the-art approaches on real-world images to evaluate the generalization ability of our method in the real-world scenery, where the real-world images used in the experiment are chosen from the URHI dataset. There is not quantitative comparison since the clean ground truth images of the real-world hazy images are not available, and the qualitative comparison results are shown in Fig.~\ref{fig: URHI}.

We note that the results on real-world data are primarily the same as those on the synthetic data. The dehazed images of NLD suffer from severe color distortions and halo artifacts (see e.g., Fig.~\ref{fig: URHI} (b)), as well as other problems such as overexposure (see e.g., the second, fifth rows in Fig.~\ref{fig: URHI} (b)). DehazeNet and AOD-Net tend to under-dehaze images (see e.g., the second, third, fifth and last rows in Fig.~\ref{fig: URHI} (c) (d)). DCPDN, EPDN and DA\_dehaze have better dehazing performance but tend to cause halo artifact in the sky area and the object edges with large color differences (see e.g., the sky area of the first, second, third, fourth and sixth rows and the edge of the airplane of the third row in Fig.~\ref{fig: URHI} (e) (f) (g)). Compared with the state-of-the-art ones, our methods have the best visual effect. As shown in Fig.~\ref{fig: URHI}, in addition to effective haze removal, T-Net and Stack T-Net can also suppress color distortions and halo artifacts. There are almost no artifacts or distortions in our dehazed images of which the sky area and the object edges are clean and smooth (see e.g., the first four rows in Fig.~\ref{fig: URHI} (h) (i)). Moreover, our method can restore the color of the image better than other methods (see e.g., the first row in Fig.~\ref{fig: URHI} of which the color is most similar to the hazy image). Although T-Net and Stack T-Net both have good performance on real world images, Stack T-Net can further improve the quality of the images (see e.g., the fifth row in Fig.~\ref{fig: URHI} (h) (i), where the image of (i) has higher brightness and looks cleaner than the same image of (g)). Specially, even for images with severe haze and deep depth (see e.g., the second row in Fig.~\ref{fig: URHI}), our method can still remove a certain amount of haze while maintaining the authenticity of the images, without causing image distortion due to halo artifacts like other methods. Overall, our methods achieve a good balance between dehazing and maintaining image authenticity.

\section{Conclusion}
\label{sec: concl}
In this work, we have proposed a new end-to-end dehazing network named T-Net, which is built based on the U-Net architecture and contains a backbone module and a dual attention module. Inspired by the recursive strategy, we have further proposed Stack T-Net by repeatedly unfolding the plain T-Net. Through the ablation studies, we verify that the overall design of T-Net is effective and the recursive strategy is helpful for dehazing. Experimental results on both synthetic and real-world images demonstrate that our T-Net and Stack T-Net perform favorably against the state-of-the-art dehazing algorithms, and that our Stack T-Net could further improve the dehazing performance.

The real-world images under severe haze lose amount of texture details and color information, which causes the performance degradation of the state-of-the-art dehazing methods or makes the color and details of the dehazing results deviate from the original images. Our future work will focus on this problem, and further study the inherent latent features of invariance.

\section*{Acknowledgment}
This work was partially supported by National Natural Science Foundation of China (Nos. 61771319, 62076165, 61871154), Natural Science Foundation of Guangdong Province (No. 2019A1515011307), Shenzhen Science and Technology Project (No. JCYJ20180507182259896) and the other project (Nos. 2020KCXTD004, WDZC20195500201).

\bibliographystyle{IEEEtran}
\bibliography{egbib}

\end{document}